%% file: main.tex
\documentclass[letterpaper,11pt]{article}
\usepackage{parskip}
\usepackage[english]{babel}
\usepackage[nottoc]{tocbibind}

\usepackage[square]{natbib}

\usepackage{diagbox}
\usepackage{makecell}
\usepackage{booktabs}
\usepackage{tikz}
\usepackage{amsmath}
\usepackage{amsfonts}
\usepackage{amssymb}
\usepackage{amsthm}
\usepackage{url,ifthen}
\usepackage{enumitem}
\usepackage{srcltx}
\usepackage{multirow}
\usepackage{boxedminipage}
\usepackage[margin=1.1in]{geometry}
\usepackage{nicefrac}
\usepackage{xspace}
\usepackage{graphicx}
\usepackage{color}
\usepackage{colortbl}
\usepackage{setspace}
\usepackage{soul}

\usepackage{hyperref}
\usepackage{cleveref}
\usepackage{mathptmx}

\definecolor{DarkGreen}{rgb}{0.1,0.5,0.1}
\definecolor{DarkRed}{rgb}{0.5,0.1,0.1}
\definecolor{DarkBlue}{rgb}{0.1,0.1,0.5}
\definecolor{Gray}{rgb}{0.2,0.2,0.2}

\usepackage{listings}
\lstdefinestyle{mystyle}{
    commentstyle=\color{DarkBlue},
    keywordstyle=\color{DarkRed},
    numberstyle=\tiny\color{Gray},
    stringstyle=\color{DarkGreen},
    basicstyle=\footnotesize,
    breakatwhitespace=false,         
    breaklines=true,                 
    captionpos=b,                    
    keepspaces=true,                 
    numbers=left,                    
    numbersep=5pt,                  
    showspaces=false,                
    showstringspaces=false,
    showtabs=false,                  
    tabsize=2
}
\usepackage[small]{caption}
\usepackage{hyperref}
\hypersetup{
    unicode=false,          
    pdftoolbar=true,        
    pdfmenubar=true,        
    pdffitwindow=false,      
    pdfnewwindow=true,      
    colorlinks=true,       
    linkcolor=DarkGreen,          
    citecolor=DarkGreen,        
    filecolor=DarkRed,      
    urlcolor=DarkBlue,          
    pdftitle={},
    pdfauthor={},
}

\def\draft{1}

\def\submit{0}

\ifnum\draft=1 
    \def\ShowAuthNotes{1}
\else
    \def\ShowAuthNotes{0}
\fi

\ifnum\submit=1
\newcommand{\forsubmit}[1]{#1}
\newcommand{\forreals}[1]{}
\else
\newcommand{\forreals}[1]{#1}
\newcommand{\forsubmit}[1]{}
\fi

\ifnum\ShowAuthNotes=1
\newcommand{\authnote}[2]{{ \footnotesize \bf{\color{DarkRed}[#1's Note:
{\color{DarkBlue}#2}]}}}
\else
\newcommand{\authnote}[2]{}
\fi

\newtheorem{theorem}{Theorem}[section]
\newtheorem{remark}[theorem]{Remark}
\newtheorem{lemma}[theorem]{Lemma}
\newtheorem{corollary}[theorem]{Corollary}
\newtheorem{proposition}[theorem]{Proposition}

\theoremstyle{definition}

\newtheorem{assumption}[theorem]{Assumption}

\newtheoremstyle{example_contd}
{\topsep} {\topsep}%
{}
{}
{\bfseries}
{.}
{1em}
{\thmname{#1} \thmnumber{ #2}\thmnote{#3} (continued)}

\theoremstyle{example_contd}

\newcommand{\chapterref}[1]{\hyperref[ch:#1]{Chapter~\ref{ch:#1}}}

\newcommand{\claimref}[1]{\hyperref[claim:#1]{Claim~\ref{claim:#1}}}

\newcommand{\corollaryref}[1]{\hyperref[cor:#1]{Corollary~\ref{cor:#1}}}

\newcommand{\definitionref}[1]{\hyperref[def:#1]{Definition~\ref{def:#1}}}

\newcommand{\equationref}[1]{\hyperref[eq:#1]{Equation~\ref{eq:#1}}}

\newcommand{\factref}[1]{\hyperref[fact:#1]{Fact~\ref{fact:#1}}}

\newcommand{\figureref}[1]{\hyperref[fig:#1]{Figure~\ref{fig:#1}}}

\newcommand{\tableref}[1]{\hyperref[tab:#1]{Table~\ref{tab:#1}}}

\newcommand{\itemref}[1]{\hyperref[item:#1]{Item~(\ref{item:#1})}}

\newcommand{\lemmaref}[1]{\hyperref[lem:#1]{Lemma~\ref{lem:#1}}}

\newcommand{\propref}[1]{\hyperref[prop:#1]{Proposition~\ref{prop:#1}}}

\newcommand{\propositionref}[1]{\hyperref[prop:#1]{Proposition~\ref{prop:#1}}}

\newcommand{\remarkref}[1]{\hyperref[rem:#1]{Remark~\ref{rem:#1}}}

\newcommand{\sectionref}[1]{\hyperref[sec:#1]{Section~\ref{sec:#1}}}

\newcommand{\theoremref}[1]{\hyperref[thm:#1]{Theorem~\ref{thm:#1}}}

\usepackage[utf8]{inputenc}
\usepackage[T1]{fontenc}
\usepackage{kpfonts}
\usepackage{microtype}

\newcommand{\Psymb}{\mathbb{P}}

\DeclareMathOperator*{\ProbOp}{\Psymb r}
\renewcommand{\Pr}{\ProbOp}

\renewcommand{\hat}{\widehat}

\newcommand{\cA}{{\cal A}}

\newcommand{\cD}{{\cal D}}

\newcommand{\cX}{{\cal X}}
\newcommand{\cY}{{\cal Y}}
\newcommand{\cZ}{{\cal Z}}

\renewcommand{\leq}{\leqslant}
\renewcommand{\le}{\leqslant}
\renewcommand{\geq}{\geqslant}
\renewcommand{\ge}{\geqslant}

\usepackage{bm}

\newcommand{\ignore}[1]{}

\DeclareMathOperator*{\argmin}{arg\,min}

\renewcommand{\epsilon}{\varepsilon}

\newcommand{\remove}[1]{}

\usepackage{color-edits}
\addauthor{hui}{cyan}

\newcommand{\bE}{\mathbb{E}}
\newcommand{\bG}{\mathbb{G}}

\newcommand{\bN}{\mathbb{N}}

\newcommand{\bR}{{\mathbb R}}

\usepackage{bm}
\usepackage{amsfonts}
\usepackage[utf8]{inputenc} 
\usepackage[T1]{fontenc}    
\usepackage{hyperref}       
\usepackage{url}            
\usepackage{booktabs}       
\usepackage{amsfonts}       
\usepackage{nicefrac}       
\usepackage{microtype}      
\usepackage{xcolor}         
\usepackage{comment}

\usepackage{algorithm}
\usepackage{algpseudocode}
\usepackage{diagbox}
\usepackage{makecell}
\usepackage{booktabs}
\usepackage{tikz}
\usepackage{amsmath}
\allowdisplaybreaks
\usepackage{amssymb}
\usepackage{amsthm}
\usepackage{url,ifthen}
\usepackage{enumitem}
\usepackage{srcltx}
\usepackage{multirow}
\usepackage{boxedminipage}
\usepackage{nicefrac}
\usepackage{xspace}
\usepackage{graphicx}
\usepackage{array,booktabs,longtable}
\usepackage{color}
\usepackage{colortbl}
\usepackage{setspace}
\usepackage{soul}
\usepackage{graphicx}
\usepackage{wrapfig}
\usepackage{subcaption}
\usepackage{enumitem}
\usepackage{cleveref}
\definecolor{DarkGreen}{rgb}{0.1,0.5,0.1}
\definecolor{DarkRed}{rgb}{0.5,0.1,0.1}
\definecolor{DarkBlue}{rgb}{0.1,0.1,0.5}
\definecolor{Gray}{rgb}{0.2,0.2,0.2}

\title{Statistical Inference under Performativity: From Policy Feedback to Self-Evolving Systems}

\usepackage{authblk}

\author[1]{Tianwei Gao}
\author[2]{Garrett G. Wen}
\author[3]{Huiying Zhong}
\author[4]{Lihua Lei}
\author[5]{Weijie Su}
\author[1]{Zhun Deng}
\affil[1]{University of North Carolina at Chapel Hill}
\affil[2]{Yale University}
\affil[3]{Massachusetts Institute of Technology}
\affil[4]{Stanford University}
\affil[5]{University of Pennsylvania}

\date{}

\begin{document}

\maketitle

\begin{abstract}
Performativity arises when prediction-informed decisions change the data-generating process, creating feedback between deployment and learning. This phenomenon occurs in policy-making and in self-evolving AI systems that shape their own future training data. Statistical inference is challenging because estimation errors in earlier predictors affect subsequent data distributions and propagate into later updates. We develop an end-to-end framework for statistical inference under performativity. For repeated risk minimization and repeated gradient descent, we establish central limit theorems at fixed deployment rounds and derive covariance recursions that capture both current-round sampling variation and uncertainty inherited from earlier deployments. We develop consistent covariance estimators using policy perturbations and score matching. These results can yield asymptotically valid confidence regions for population iterates and the performative stable point. We further extend prediction-powered inference to this dynamic setting, combining labeled observations with predictions on additional unlabeled data to improve statistical efficiency. We evaluate our framework in synthetic, semi-synthetic educational, and self-evolving large language model routing settings. Our experiments demonstrate that ignoring propagated uncertainty can lead to substantial undercoverage even after repeated retraining. Together, these results provide statistical tools for learning systems that help determine the distributions from which they learn. Our work establish a complete statistical inference under performativity, introducing new challenges and inference settings that we believe will provide substantial value to policy-making, statistics, and machine learning. 
\end{abstract}

\input{Sections/intro}

\input{Sections/background}


\input{Sections/method}
\input{Sections/method3}

\input{Sections/experiment}

\input{Sections/conclusion}

\clearpage
\bibliography{refs}
\bibliographystyle{plainnat}

\clearpage
\newpage

\appendix



\input{Sections/App_A}

\input{Sections/app_exp_new}

\end{document}

%% file: Sections/intro.tex
\section{Introduction}

Prediction-informed decisions can shape the very outcomes they seek to predict. This phenomenon, known as \textit{performativity}, arises when deploying a predictive model changes the underlying data-generating process. For example, lending policies based on predicted default risk can alter borrowers' consumption and repayment behavior, thereby influencing the outcomes used to train future predictors.

The same feedback arises in self-evolving large language model (LLM) systems, where the current model guides the generation and selection of data for subsequent learning \citep{zelikman2022star,singh2023beyond,wang2025temporal,zweiger2026self}. By influencing which prompts and responses are generated and retained, these systems participate in producing their own supervision. Updating the model can therefore change not only its predictions but also the distribution of its future training data.

In both settings, deployment and learning form a feedback loop: the deployed predictor shapes the data collected for retraining, and the updated predictor can induce a further distributional shift. This raises a statistical question that goes beyond prediction accuracy and convergence: \textit{how can we quantify uncertainty in predictors along this evolving sequence and in the equilibrium they may approach?}

The framework of \textit{performative prediction} formalizes this dependence between predictive models and the distributions they induce \citep{perdomo2020performative,lin2023plug,Perdomo2025RevisitingTP}. For a model parameter $\theta$ in a parameter space $\Theta$, the \textit{performative risk} is defined as
\begin{equation}\label{eq:pp}
    \operatorname{PR}(\theta)
    :=
    \mathbb{E}_{z\sim\mathcal{D}(\theta)}
    \bigl[\ell(z;\theta)\bigr],
\end{equation}
where $z=(x,y)\in\mathcal{X}\times\mathcal{Y}$ is an input-output pair, $\mathcal{D}(\theta)$ is the distribution induced by deploying $\theta$, and $\ell(z;\theta)$ is the loss incurred on observation $z$. Finding a \textit{performatively optimal} point $\theta_{\text{PO}}\in\arg\min_{\theta\in\Theta}\operatorname{PR}(\theta)$ is challenging: the distribution map $\mathcal{D}(\cdot)$ is generally unknown, and the performative risk can be nonconvex even when the loss is convex in $\theta$ \citep{Miller2021OutsideTE}. One approach is to estimate this map within a specified parametric family and optimize the resulting plug-in performative risk \citep{lin2023plug}. This approach requires an explicit model of the deployment response, and its optimality guarantees depend on how well the chosen family captures that response.

A natural alternative is a \textit{performative stable} point \citep{perdomo2020performative}, which satisfies
\[
    \theta_{\text{PS}}
    \in
    \arg\min_{\theta\in\Theta}
    \mathbb{E}_{z\sim\mathcal{D}(\theta_{\text{PS}})}
    \bigl[\ell(z;\theta)\bigr].
\]
At this equilibrium, the predictor minimizes risk under the distribution induced by its own deployment, so further population-level retraining leaves it unchanged. Under suitable regularity conditions, the stable point is unique and can be approached through repeated retraining without explicitly modeling the distribution map. When deployment-induced shifts are sufficiently mild, it is also close to a performatively optimal point, making it a useful proxy for the Stackelberg equilibrium in strategic classification. These properties make performative stability a meaningful target for both policy-making and self-evolving systems whose deployment shapes subsequent supervision.

Convergence to a stable point, however, does not by itself quantify uncertainty. In policy-making, $\theta$ often represents an actionable quantity, such as a tax rate or a credit-score cutoff. Confidence regions and hypothesis tests are therefore important complements to point estimates and convergence guarantees. More generally, statistical inference is needed both for the population predictors at intermediate deployment rounds and for the equilibrium they approach.

The central challenge is that uncertainty itself propagates through the feedback loop. Estimation errors in earlier predictors affect subsequent data distributions and, consequently, the variability of later updates. Accounting only for the sampling uncertainty within the current round can therefore miss an important source of variation. Our analysis characterizes this propagation through the distributional score, which measures the sensitivity of the log density of $\mathcal{D}(\theta)$ to the deployed parameter $\theta$. Estimating this sensitivity from data is a key step toward constructing valid confidence regions under performative feedback.

Previous work on performative prediction has largely emphasized prediction performance and convergence analysis. A closely related contribution by \citet{cutler2024stochastic} establishes asymptotic normality for averaged stochastic gradient iterates as the number of updates tends to infinity. This regime differs from applications in which only a limited number of costly deployments may be feasible, such as policy-making. Moreover, their results do not provide a data-driven procedure for estimating the covariance required for inference. Our focus is instead on inference at fixed deployment rounds as the sample size within each round grows, together with covariance estimation that makes the resulting procedures implementable.

Our preliminary NeurIPS paper \citep{li2025statistical} established, to the best of our knowledge, the first end-to-end statistical inference framework for repeated risk minimization (RRM) under performativity and introduced prediction-powered inference (PPI) into this setting. The present work extends this framework to repeated gradient descent (RGD) and develops applications to self-evolving LLM systems.

\paragraph{Our contributions.}
We develop a framework for statistical inference that accounts for the dependence between estimation, deployment, and subsequent data collection. Our contributions are threefold.

\begin{enumerate}
    \item \textbf{Statistical inference for performative learning trajectories and stable points.}
    We study RRM and RGD, two iterative procedures for approaching $\theta_{\text{PS}}$ (Section~\ref{sec:bg}), and establish central limit theorems for their empirical iterates at each fixed deployment round. The limiting covariance recursions account for both current-round sampling variation and uncertainty propagated from earlier deployments. When the distributional score is unknown, we estimate it through policy perturbations and score matching, and establish consistency of the resulting covariance estimator under the stated regularity and score-model conditions (Section~\ref{subsec:RRMcov}). This approach estimates the sensitivity needed for inference directly, without fitting a full parametric distribution map. The resulting estimators yield asymptotically valid confidence regions for the population iterates and, after accounting for the remaining iteration bias, for the performative stable point.

    \item \textbf{Prediction-powered inference under performativity.}
    We extend PPI \citep{angelopoulos2023ppi++} to the dynamic performative setting by combining labeled observations with predictions on additional unlabeled data (Appendix~\ref{sec:ppi}). We derive the corresponding central limit theorem and covariance recursion along the RRM trajectory, and select the PPI weight at each round to minimize the trace of the asymptotic covariance with earlier weights held fixed. This provides a data-driven way to improve statistical efficiency while accounting for deployment-induced feedback. Such gains are particularly relevant when reliable labeled feedback is difficult to collect, as in survey-based applications \citep{huang2015insufficient}.

    \item \textbf{Evaluation across policy feedback and self-evolving LLM systems.}

We evaluate our framework in three settings with distinct feedback mechanisms (Section~\ref{sec:exp}): a synthetic model in which deployment changes the population composition; a semi-synthetic educational setting grounded in a study of growth-mindset interventions \citep{yeager2019national,athey2019estimating}, where predictions guide interventions that affect subsequent student outcomes; and a self-evolving LLM routing task. Importantly, in the third setting, we formalize a connection between performative prediction and self-evolving AI by modeling how the deployed router determines which responses are sampled and thereby shapes the supervision used for its own subsequent updates. This formulation enables statistical inference for the router's population learning trajectory and performative stable point. Across these settings, conventional confidence regions for a one-step empirical risk minimizer can substantially undercover the performative stable point despite achieving near-nominal coverage of their own target. A controlled synthetic study with stronger feedback further demonstrates that repeated retraining alone does not ensure calibrated inference: covariance estimation must also account for uncertainty propagated across deployments.

\end{enumerate}

Together, these results extend statistical inference from models trained on a fixed distribution to learning procedures that help determine the distributions from which they learn. The framework connects uncertainty quantification for policy feedback with inference for self-evolving AI systems, while enabling more efficient use of limited labeled data through prediction-powered inference.

\subsection{Related Work}
\paragraph{Performative prediction.} Performativity describes the phenomenon whereby predictions influence the outcomes they aim to predict. \cite{perdomo2020performative} were the first to formalize performative prediction in the supervised‐learning setting; their work, along with the majority of subsequent papers \citep{MendlerDnner2020StochasticOF, Drusvyatskiy2020StochasticOW, Mofakhami2023PerformativePW, Oesterheld2023IncentivizingHP, Taori2022DataFL, Khorsandi2024TightLB, Perdomo2025RevisitingTP,bracale2024microfoundation}, have been focused on performative stability and proposed algorithms for learning performative stable parameters. On the other hand, performative optimality requires much stricter conditions (e.g. distributional assumptions to ensure the convexity of $\text{PR}(\theta)$) than performative stability, a few papers address the problem of finding performative optimal parameters. \cite{Miller2021OutsideTE} introduce a two‐stage method that learns a distribution map to locate the performative optimal parameter. \cite{Kim2022MakingDU} study performative optimality in outcome‐only performative settings. Finally, \cite{Hardt2023PerformativePP} provide a comprehensive overview of learning algorithms, optimization methods, and applications for performative prediction. Unlike these prior works on performative prediction, which focus on prediction accuracy, our work is dedicated to constructing powerful and statistically valid inference procedures under the performative framework.

\paragraph{Prediction-powered inference.} \cite{Angelopoulos2023PredictionpoweredI} first introduced the prediction-powered inference (PPI) framework, which leverages black-box machine learning models to construct valid confidence intervals (CIs) for statistical quantities. Since then, PPI has been extended and applied in various settings. Closely related to our strategies, \cite{angelopoulos2023ppi++} propose PPI++, a more computationally efficient procedure that enhances predictability by accommodating a wider range of models on unlabeled data, while guaranteeing performance (e.g.\ CI width) no worse than that of classical inference methods. Other extensions include Stratified PPI \citep{Fisch2024StratifiedPI}, which incorporates simple data stratification strategies into basic PPI estimates; Cross PPI \citep{Zrnic2023CrosspredictionpoweredI}, which obtains confidence intervals with significantly lower variability by including model training; Bayesian PPI \citep{Hofer2024BayesianPI} and FAB-PPI \citep{Cortinovis2025FABPPIFA}, which propose frameworks for PPI based on Bayesian inference. PPI is also connected to topics such as semi-parametric inference and missing-data imputation \citep{Tsiatis2006SemiparametricTA,Robins1995SemiparametricEI,Demirtas2018FlexibleIO,Song2023AGM}. 
Our work is the first one to study PPI under performativity, and we validate the PPI framework in the performative setting both theoretically and empirically.

\paragraph{Inference in performativity.}\cite{bracale2024microfoundation} study identifiability and estimation error under a specific microfoundation model with performativity. Yet it doesn't address confidence interval construction. A closely related work is \citep{cutler2024stochastic}, which also establishes the asymptotic normality and minimax optimality for performative settings. It focused on averaged stochastic gradient iterates under performativity as time $t\to\infty$, and did not provide fixed round central limit theorem. 
Moreover, \citep{cutler2024stochastic} do not provide a data-driven approach for covariance estimation and thus lacks an end-to-end inference framework. In contrast, our work explicitly handles score function estimation and provides an end-to-end inference method for constructing confidence intervals, which is firstly addressed by our preliminary article \citep{li2025statistical}, to the best of our knowledge. 

%% file: Sections/background.tex
\section{Background and Setup}\label{sec:bg}
We consider the setting in which the current parameter changes
the distribution of the next training run. Let \(\Theta=\mathbb R^d\), let
\(\mathcal D(\theta)\) denote the distribution induced by parameter
\(\theta\in\Theta\), and let \(\ell(z;\theta)\) denote the loss of a
candidate parameter \(\theta\) on an observation \(z\), with gradient
\(\nabla\ell(z;\cdot)\) and Hessian \(\nabla^2\ell(z;\cdot)\) in its parameter argument.

Let \(\theta_0\in\Theta\) be fixed and set
\(\widehat\theta_0=\theta_0\). At round \(t\geq0\), conditional on all
observations and estimators constructed before that round, the labeled
observations \(Z_{t,1},\ldots,Z_{t,n_t}\) are i.i.d. from
\(\mathcal D(\widehat\theta_t)\). At round \(t\),
\(\widehat\theta_t\) therefore determines the sampling distribution, and
batch \(t\) is used to construct \(\widehat\theta_{t+1}\).
\subsection{Repeated Risk Minimization}\label{subsec:bgrrm}

Define the population update
 $
G(\widetilde\theta)
:=\arg\min_{\theta\in\Theta}
\mathbb E_{Z\sim\mathcal D(\widetilde\theta)}
[\ell(Z;\theta)].
$
Under Assumption~\ref{as:pp}, the minimizer is unique. Recall that the main objective of interest is the \textit{performative stable} point, which satisfies the following relationship: 
\[
\theta_{\mathrm{PS}}=G(\theta_{\mathrm{PS}})
=\arg\min_{\theta\in\Theta}
\mathbb E_{Z\sim\mathcal D(\theta_{\mathrm{PS}})}
[\ell(Z;\theta)].
\]

Repeated risk minimization (RRM) is a simple algorithm that can efficiently find $\theta_{\text{PS}}$. Specifically, one starts with an arbitrary $\theta_0$ and repeats the procedure
$
\theta_{t+1}=G(\theta_t), t\geq0,
$
for $t\in \bN$. Under some regularity conditions, this update is well-defined and provably converges to a unique $\theta_{\text{PS}}$ at a linear rate. 

\begin{theorem}[Informal, adopted from \citep{perdomo2020performative}]\label{thm:RRMinformal}
If the loss is smooth, strongly convex, and the mapping $\cD(\cdot)$ satisfies certain Lipchitz conditions, then:

(a). For any $\theta^\prime \in \Theta$, $G(\theta^\prime)$ is unique and well-defined, and $G:\Theta\to \Theta$ is a contraction map. 

(b). $\theta_{\text{PS}}$ is uniquely defined as the fixed point of $G$, and repeated risk minimization converges to $\theta_{\text{PS}}$ at a linear rate.
\end{theorem}

\paragraph{Repeated empirical risk minimization (RERM).}
At round \(t\), repeated empirical risk minimization computes
\[
\widehat\theta_{t+1}
=\arg\min_{\theta\in\Theta}\frac1{n_t}
\sum_{i=1}^{n_t}\ell(Z_{t,i};\theta).
\]
Conditional on the preceding observations and estimators, this empirical
minimizer estimates \(G(\widehat\theta_t)\), whereas the corresponding
deterministic population iterate is \(\theta_{t+1}=G(\theta_t)\).
This gives the population trajectory
\(\theta_0\to\theta_1\to\cdots\) and the empirical trajectory
\(\theta_0\to\widehat\theta_1\to\cdots\). Their difference satisfies
$
\widehat\theta_{t+1}-\theta_{t+1}=\bigl(\widehat\theta_{t+1}- G(\widehat\theta_t)\bigr)
+ \bigl(G(\widehat\theta_t)-G(\theta_t)\bigr),
$ where the first term represents sampling error in the current empirical minimization. The second represents the effect of the preceding estimation error on the population update. Our inference results account for both contributions when characterizing the asymptotic distribution of \(\widehat\theta_{t+1}-\theta_{t+1}\).

\subsection{Repeated Gradient Descent}\label{subsec:bgrgd}

Repeated gradient descent (RGD) uses the population gradient under the
distribution induced by the current parameter \citep{perdomo2020performative} :
\[
\theta_{t+1}
=\theta_t-\alpha
\mathbb E_{Z\sim\mathcal D(\theta_t)}
\left[
\nabla\ell(Z;\theta_t)
\right],
\qquad t\geq0,
\]
where \(\alpha>0\) is the learning rate.
\begin{theorem}[Informal, adopted from \citep{perdomo2020performative}]\label{thm:RGDinformal}
If the loss is smooth, strongly convex, and the mapping $\cD(\cdot)$ satisfies certain Lipchitz conditions, then $\theta_{\text{PS}}$ is uniquely defined and repeated gradient descent converges to $\theta_{\text{PS}}$ at a linear rate.
\end{theorem}

\paragraph{Repeated empirical gradient descent (REGD).}
Starting from \(\widehat\theta_0=\theta_0\), repeated empirical gradient
descent updates
\[
\widehat\theta_{t+1}
=\widehat\theta_t-\frac{\alpha}{n_t}
\sum_{i=1}^{n_t}
\nabla\ell(Z_{t,i};\widehat\theta_t).
\]
We use \(\{\theta_t\}_{t=0}^T\) and
\(\{\widehat\theta_t\}_{t=0}^T\) for the population and empirical RGD
trajectories. For both RERM and REGD, we study inference on
\(\widehat\theta_t\) at each fixed \(t\geq1\), and we let \(n_t=n\) for all
rounds.

\subsection{Assumptions and Notation}\label{subsec:bgas}
We introduce the following foundational assumptions for our setup. We assume that $\cZ$ is a subset of Euclidean space with   norm $\|\cdot\|_{2}$.

\begin{assumption}[Loss and distribution map]\label{as:pp}
The following conditions hold.

(a) (Local Boundedness) For every compact \(K\subset\Theta\), there exists a measurable
\(A_K:\mathcal Z\to[0,\infty)\) such that
 $
|\ell(z;\theta)|\leq A_K(z), \theta\in K,
$
and
$
\sup_{\widetilde\theta\in K}
\mathbb E_{Z\sim\mathcal D(\widetilde\theta)}[A_K(Z)^2]<\infty.
$

(b) (Local Lipschitzness) For every compact \(K\subset\Theta\), there exists a measurable
\(L_K:\mathcal Z\to[0,\infty)\) such that
$
|\ell(z;\theta)-\ell(z;\theta')|
\leq L_K(z)\|\theta-\theta'\|_2,
 \theta,\theta'\in K,
$
and
$
\sup_{\widetilde\theta\in K}
\mathbb E_{Z\sim\mathcal D(\widetilde\theta)}[L_K(Z)^3]<\infty.
$

(c) (Joint Smoothness) The loss is twice continuously differentiable in its second argument, and for all
\(z,z'\in\mathcal Z\) and \(\theta,\theta'\in\Theta\),
\[
\|\nabla\ell(z;\theta)
-\nabla\ell(z;\theta')\|_2
\leq\beta\|\theta-\theta'\|_2,
\qquad 
\|\nabla\ell(z;\theta)
-\nabla\ell(z';\theta)\|_2
\leq\beta\|z-z'\|_2.
\]

(d) (Strong Convexity) For every \(z\in\mathcal Z\), the map
\(\theta\mapsto\ell(z;\theta)\) is
\(\gamma\)-strongly convex: for all \(\theta,\theta'\in\Theta\),
\[
\ell(z;\theta)\geq\ell(z;\theta')
+\nabla\ell(z;\theta')^\top
(\theta-\theta')
+\frac\gamma2\|\theta-\theta'\|_2^2.
\]

(e) ($\epsilon$-Sensitivity) The distribution map is \(\varepsilon\)-sensitive:
 $
W_1\{\mathcal D(\theta),\mathcal D(\theta')\}
\leq\varepsilon\|\theta-\theta'\|_2,
 \theta,\theta'\in\Theta.
$
\end{assumption}
 Under Assumption~\ref{as:pp}, \citet[Theorem 3.5]{perdomo2020performative} gives
$
\|G(\theta)-G(\theta')\|_2
\leq\frac{\varepsilon\beta}{\gamma}\|\theta-\theta'\|_2.
$
 Hence \(G\) has a unique fixed
point when \(\varepsilon\beta<\gamma\).

\begin{assumption}[Differentiability of the distribution map]
\label{as:consistent2}
(a) Each \(\mathcal D(\theta)\) has a positive density
\(p(\cdot;\theta)\) with
respect to a common dominating measure. The map
\(\theta\mapsto\log p(z;\theta)\) is differentiable. For every compact
\(K\subset\Theta\), there exists \(L_{s,K}<\infty\) such that
$
\sup_{\theta\in K}
\|\nabla_{\theta}\log p(z;\theta)\|_2\leq L_{s,K},
 z\in\mathcal Z.
$

(b) The map
$
(\widetilde\theta,\theta)\longmapsto
\mathbb E_{Z\sim\mathcal D(\widetilde\theta)}
[\nabla\ell(Z;\theta)]
$
is continuously differentiable.
\end{assumption}
\paragraph{Notation}
For \(K\in\mathbb N_+\), let \([K]=\{1,\ldots,K\}\). We write
\(\xrightarrow{P}\) and \(\xrightarrow{d}\) for convergence in probability
and in distribution. For sets \(A,B\subset\mathbb R^d\), let
\(A+B=\{a+b:a\in A,b\in B\}\). We write
\(\mathcal N(\mu,\Sigma)\) for a Gaussian distribution,
\(I_d\) for the \(d\times d\) identity matrix,
\(\mathcal B(c,r)\) for the Euclidean ball centered at \(c\) with radius
\(r\).
For a differentiable map \(F:\mathbb R^d\to\mathbb R^d\), \(\nabla F\) denotes
its Jacobian matrix; for a matrix \(A\), \(\|A\|_2\) denotes its spectral norm.

%% file: Sections/method.tex

\section{Statistical Inference for RRM}\label{sec:RRM}
This section develops inference for the population iterate \(\theta_t\)
based on the RERM iterate \(\widehat\theta_t\), with \(t\) fixed and
\(n\to\infty\). Because batch \(t\) is drawn from
\(\mathcal D(\widehat\theta_t)\), the error of \(\widehat\theta_{t+1}\)
contains both the sampling error of the current round and the error
inherited from earlier rounds. Section~\ref{subsec:RRMclt} gives the
resulting central limit theorem, Section~\ref{subsec:RRMcov} estimates its
covariance, and Section~\ref{subsec:infRRM} turns the confidence region for
\(\theta_t\) into one for \(\theta_{\mathrm{PS}}\).

\subsection{Central Limit Theorem of RRM} \label{subsec:RRMclt}
We start from the decomposition in Section~\ref{subsec:bgrrm},
$
\widehat\theta_{t+1}-\theta_{t+1}
=\bigl(\widehat\theta_{t+1}-G(\widehat\theta_t)\bigr)
+\bigl(G(\widehat\theta_t)-G(\theta_t)\bigr).
$
The first term is the error of an M-estimator. Conditional on all preceding
observations and estimators, \(\widehat\theta_{t+1}\) minimizes the empirical
risk of \(n\) i.i.d. observations from \(\mathcal D(\widehat\theta_t)\), and
\(G(\widehat\theta_t)\) minimizes the corresponding population risk. For
\(\widetilde\theta,\theta\in\Theta\), define
\[
H_{\widetilde\theta}(\theta)
:=\mathbb E_{Z\sim\mathcal D(\widetilde\theta)}
[\nabla^2\ell(Z;\theta)],
\qquad
\Sigma_{\widetilde\theta}(\theta)
:=H_{\widetilde\theta}(\theta)^{-1}
\operatorname{Cov}_{Z\sim\mathcal D(\widetilde\theta)}
\bigl(\nabla\ell(Z;\theta)\bigr)
H_{\widetilde\theta}(\theta)^{-1}.
\]
By standard results for M-estimators, the conditional distribution of
\(\sqrt n\{\widehat\theta_{t+1}-G(\widehat\theta_t)\}\) is approximately
\(\mathcal N\bigl(0,\Sigma_{\widehat\theta_t}(G(\widehat\theta_t))\bigr)\),
and \(\Sigma_{\widehat\theta_t}(G(\widehat\theta_t))\) tends to
\(\Sigma_{\theta_t}(\theta_{t+1})\) as \(\widehat\theta_t\to\theta_t\).
The second term is approximately
\(\nabla G(\theta_t)(\widehat\theta_t-\theta_t)\), where \(\nabla G\) is
given by the following proposition.
\begin{proposition}\label{prop:differentiability}
Under Assumptions~\ref{as:pp} and~\ref{as:consistent2}, \(G\) is
continuously differentiable, and
\[
\nabla G(\theta)
=-H_{\theta}(G(\theta))^{-1}
\mathbb E_{Z\sim\mathcal D(\theta)}
\bigl[\nabla\ell(Z;G(\theta))\,
\nabla_{\theta}\log p(Z;\theta)^\top\bigr].
\]
\end{proposition}
The expectation measures how the gradient of the risk at \(G(\theta)\)
changes when the distribution moves with \(\theta\).
\begin{theorem}[Central limit theorem for RERM]\label{thm:cltrrm}
Under Assumptions~\ref{as:pp} and~\ref{as:consistent2}, for each fixed
\(t\geq1\),
\[
\sqrt n(\widehat\theta_t-\theta_t)\xrightarrow{d}\mathcal N(0,V_t),
\]
where \(V_0=0\) and
$
V_{t+1}=\nabla G(\theta_t)\,V_t\,\nabla G(\theta_t)^\top
+\Sigma_{\theta_t}(\theta_{t+1}),  t\geq0.
$
\end{theorem}
If \(\mathcal D(\theta)\) does not depend on \(\theta\), then \(\nabla G=0\)
and \(V_t=\Sigma_{\theta_{t-1}}(\theta_t)\), the covariance of a single
M-estimator. Otherwise the inherited term
\(\nabla G(\theta_t)V_t\nabla G(\theta_t)^\top\) is positive semidefinite, so
treating each round as a separate M-estimation problem understates the
uncertainty of \(\widehat\theta_{t+1}\).

\subsection{Estimation of the Asymptotic Covariance}\label{subsec:RRMcov}
The matrices \(H_{\theta_t}(\theta_{t+1})\) and
\(\Sigma_{\theta_t}(\theta_{t+1})\) are defined under \(\mathcal D(\theta_t)\)
and can be estimated from batch \(t\). The derivative \(\nabla G(\theta_t)\)
also involves the score \(\nabla_\theta\log p(z;\theta)\), which describes
how the distribution changes with the parameter. When this score is known,
as in the routing experiment of Section~\ref{sec:llmexp}, we use it directly
in the estimator below. Otherwise, we estimate it by score matching.

Let \(s(z,\theta;\psi)\in\mathbb R^d\), with coordinates \(s_j\), be a model
for \(\nabla_\theta\log p(z;\theta)\) indexed by \(\psi\in\Psi\), where $\Psi\subset \bR^{p}$ is a subset of $p$-dimensional Euclidean space. We choose
\(\psi\) to minimize
\[
\mathbb E_{Z\sim\mathcal D(\theta)}
\bigl[\|\nabla_\theta\log p(Z;\theta)-s(Z,\theta;\psi)\|_2^2\bigr].
\]
After expanding the square, the term
\(\mathbb E_{Z\sim\mathcal D(\theta)}[\|\nabla_\theta\log p(Z;\theta)\|_2^2]\)
does not depend on \(\psi\), but the cross term involves the unknown score.
\citet{hyvarinen2005estimation} removes such a term for the score in \(z\)
by integrating by parts in \(z\). Here the score is taken in \(\theta\), so
we instead move the derivative outside the expectation.

\begin{lemma}\label{lm:K}
Under Assumption~\ref{as:exchange},
\[
\mathbb E_{Z\sim\mathcal D(\theta)}
\bigl[\nabla_\theta\log p(Z;\theta)^\top s(Z,\theta;\psi)\bigr]
=\sum_{j=1}^d\Bigl\{
\frac{\partial}{\partial\theta^{(j)}}
\mathbb E_{Z\sim\mathcal D(\theta)}[s_j(Z,\theta;\psi)]
-\mathbb E_{Z\sim\mathcal D(\theta)}
[\partial_{\theta^{(j)}}s_j(Z,\theta;\psi)]\Bigr\},
\]
where \(\theta^{(j)}\) is the \(j\)th coordinate of \(\theta\).
\end{lemma}
In the first term, the derivative acts on both the distribution and the
argument of \(s_j\); the second term removes the latter. Only the first term
requires observations at parameters other than \(\theta\).

 We estimate this derivative by policy perturbation
\citep{munro2021treatment,viviano2024policy}. At round \(t\), in addition to
batch \(t\), we use the parameters \(\widehat\theta_t+\eta e_j\),
\(j\in[d]\), where \(\eta>0\) and \(e_j\) is the \(j\)th standard basis
vector. Conditional on all preceding observations and estimators,
\(Z^{(j)}_{t,1},\ldots,Z^{(j)}_{t,k}\) are i.i.d. from
\(\mathcal D(\widehat\theta_t+\eta e_j)\), independently across \(j\) and of
batch \(t\).  

We replace the expectations in Lemma~\ref{lm:K} by sample averages and the
derivative by a finite difference, and drop the term that does not depend
on \(\psi\). Let \(\widehat\psi_{n,k}(\widehat\theta_t)\) be an approximate
minimizer over \(\psi\in\Psi\) of
\begin{equation}\label{eq:smobjective}
\begin{aligned}
&\frac1n\sum_{i=1}^n\Bigl\{\|s(Z_{t,i},\widehat\theta_t;\psi)\|_2^2
+2\sum_{j=1}^d\partial_{\theta^{(j)}}s_j(Z_{t,i},\widehat\theta_t;\psi)\Bigr\}\\
&\quad-\frac2\eta\sum_{j=1}^d\Bigl\{\frac1k\sum_{m=1}^k
s_j(Z^{(j)}_{t,m},\widehat\theta_t+\eta e_j;\psi)
-\frac1n\sum_{i=1}^n s_j(Z_{t,i},\widehat\theta_t;\psi)\Bigr\},
\end{aligned}
\end{equation}
with optimization error \(a_{n,k}\) as in Assumption~\ref{as:perturb1}(d).

Using batch \(t\), which also gives \(\widehat\theta_{t+1}\), define
\[
\widehat{\nabla G}_t
:=-\widehat H_{\widehat\theta_t}(\widehat\theta_{t+1})^{-1}
\frac1n\sum_{i=1}^n
\nabla\ell(Z_{t,i};\widehat\theta_{t+1})\,
s\bigl(Z_{t,i},\widehat\theta_t;\widehat\psi_{n,k}(\widehat\theta_t)\bigr)^\top,
\]
\[
\widehat\Sigma_{\widehat\theta_t}(\widehat\theta_{t+1})
:=\widehat H_{\widehat\theta_t}(\widehat\theta_{t+1})^{-1}
\Bigl\{\frac1n\sum_{i=1}^n
\nabla\ell(Z_{t,i};\widehat\theta_{t+1})
\nabla\ell(Z_{t,i};\widehat\theta_{t+1})^\top\Bigr\}
\widehat H_{\widehat\theta_t}(\widehat\theta_{t+1})^{-1},
\]
where \(\widehat H_{\widehat\theta_t}(\widehat\theta_{t+1})
:=n^{-1}\sum_{i=1}^n\nabla^2\ell(Z_{t,i};\widehat\theta_{t+1})\).
Since the gradients average to zero at \(\widehat\theta_{t+1}\), the middle
factor of \(\widehat\Sigma\) is their sample covariance. Following
Theorem~\ref{thm:cltrrm}, set \(\widehat V_0=0\) and
\begin{equation}\label{eq:defempvar}
\widehat V_{t+1}
:=\widehat{\nabla G}_t\,\widehat V_t\,\widehat{\nabla G}_t^\top
+\widehat\Sigma_{\widehat\theta_t}(\widehat\theta_{t+1}),
\qquad t\geq0.
\end{equation}

The proof decomposes the error into empirical and population terms. Theorem~\ref{thm:cltrrm} controls the differences between $\theta_k$, $\theta_{k+1}$ and $\widehat\theta_k$, $\widehat\theta_{k+1}$. Theorem~\ref{thm:discuss1} controls the score error uniformly over the perturbed parameters.
Taken together, the above result shows the consistency of $\widehat{\nabla G}_{t}$ obtained by policy perturbation.

The next result requires the score model to contain the true score and the
fitted parameters to stay bounded (Assumption~\ref{as:perturb1}), together
with moment and smoothness conditions on \(H\) and on products of loss
gradients and scores (Assumptions~\ref{as:perturb2} and~\ref{as:perturb3}).

\begin{theorem}\label{thm:discuss2}
Suppose Assumptions~\ref{as:pp}, \ref{as:consistent2}, \ref{as:exchange},
and~\ref{as:perturb1}--\ref{as:perturb3} hold. If \(\eta\to0\) and
\(\eta\sqrt{\min(n,k)}\to\infty\), then for each fixed \(t\geq0\),
\[
\|\widehat{\nabla G}_t-\nabla G(\theta_t)\|_2^2
=O_P\Bigl(\frac1{\sqrt n}+\frac1{\eta\sqrt{\min(n,k)}}+\eta+a_{n,k}\Bigr).
\]
\end{theorem}
The four terms are the sampling error of batch \(t\), the sampling error of
the perturbed batches divided by \(\eta\), the bias of the finite
difference, and the optimization error.

\begin{corollary}\label{col:rrmfeasible}
Under the conditions of Theorem~\ref{thm:discuss2},
\(\widehat V_t\xrightarrow{P}V_t\) for each fixed \(t\geq1\). If, in
addition, \(V_t\) is nonsingular, then the region
\[
\widehat{\mathcal R}_t
:=\bigl\{\theta\in\Theta:
n(\widehat\theta_t-\theta)^\top\widehat V_t^{-1}
(\widehat\theta_t-\theta)\leq\chi^2_{d,1-\delta}\bigr\}
\]
satisfies
\(\lim_{n\to\infty}\mathbb P(\theta_t\in\widehat{\mathcal R}_t)=1-\delta\),
where \(\chi^2_{d,1-\delta}\) is the \((1-\delta)\)-quantile of the
\(\chi^2_d\) distribution.
\end{corollary}

\subsection{Confidence Regions for the Performative Stable Point}\label{subsec:infRRM}
The region \(\widehat{\mathcal R}_t\) covers the population iterate
\(\theta_t\), not \(\theta_{\mathrm{PS}}\). When \(\varepsilon\beta<\gamma\),
the contraction of \(G\) in Section~\ref{subsec:bgas} gives
$
\|\theta_t-\theta_{\mathrm{PS}}\|_2
\leq\Bigl(\frac{\varepsilon\beta}{\gamma}\Bigr)^t
\|\theta_0-\theta_{\mathrm{PS}}\|_2.
$
Enlarging \(\widehat{\mathcal R}_t\) by this deterministic bias gives a
bias-aware confidence region \citep{armstrong2018optimal,armstrong2020bias,
imbens2019optimized,noack2019bias}.

\begin{corollary}[Confidence region for \(\theta_{\mathrm{PS}}\)]\label{col:rrmps}
Suppose the conditions of Corollary~\ref{col:rrmfeasible} hold,
\(\varepsilon\beta<\gamma\), and \(\|\theta_0-\theta_{\mathrm{PS}}\|_2\leq B\).
Then
\[
\liminf_{n\to\infty}\mathbb P\Bigl\{\theta_{\mathrm{PS}}\in
\widehat{\mathcal R}_t+\mathcal B\Bigl(0,
B\Bigl(\frac{\varepsilon\beta}{\gamma}\Bigr)^t\Bigr)\Bigr\}\geq1-\delta.
\]
\end{corollary}
 Since \(\theta_{\mathrm{PS}}\) minimizes
\(\theta\mapsto\mathbb E_{Z\sim\mathcal D(\theta_{\mathrm{PS}})}
[\ell(Z;\theta)]\), strong convexity gives
 $
\|\theta_0-\theta_{\mathrm{PS}}\|_2
\leq\frac1\gamma\sup_{z\in\mathcal Z}\|\nabla\ell(z;\theta_0)\|_2.
$
The radius decreases geometrically in \(t\), whereas
\(\widehat{\mathcal R}_t\) has diameter of order \(n^{-1/2}\). The
enlargement is therefore negligible once \((\varepsilon\beta/\gamma)^t\) is
small relative to \(n^{-1/2}\).

\paragraph{Comparison with \citep{perdomo2020performative}.} \citet{perdomo2020performative} study repeated empirical risk minimization and prove that, for a sufficiently large number of iterations $T$, with probability $1-\delta$,
$$\|\widehat{\theta}_T - \theta_{\text{PS}}\|\leq C (\log(\frac{1}{\delta}))^{\frac{1}{m}} n^{-\frac{1}{m}},$$ 
where $C$ is an absolute constant and $m\geq2$ is the dimension of $z=(x,y)$. Our $1-\delta$ confidence rectangle for $\theta_T$ is
$\widehat{\mathcal R}_T(n,\delta)=[\widehat\theta_{T,j}\pm z_{1-\delta/(2d)}\sqrt{\widehat V_{T,jj}/n}]_{j=1}^d$, where $z_{1-\delta/(2d)}$ is the $1-\delta/(2d)$ quantile of the standard normal distribution. The bias-aware region for $\theta_{\mathrm{PS}}$ is obtained by the enlargement in Corollary~\ref{col:rrmps}. For each $j\in[d]$, the Gaussian component satisfies
$$|\widehat{\theta}_{T,j} -\theta_{T,j} |\leq C (\log(\frac{d}{\delta}))^{\frac{1}{2}} n^{-\frac{1}{2}} \quad \text{for any } j \in [d],$$
where $C$ is an absolute constant.
Compared to the $O(n^{-\frac{1}{m}})$ width by \citep{perdomo2020performative}, our interval achieves a better rate $O(n^{-\frac{1}{2}})$ when $m>2$ (ignoring the exponentially small term in the bias-aware adjustment). Thus, our proposed inference method provides a narrower confidence interval and improved statistical efficiency.

%% file: Sections/method3.tex
\section{Statistical Inference for RGD}\label{sec:RGD}
This section develops inference for the population RGD iterate
\(\theta_t\) based on the REGD iterate \(\widehat\theta_t\), with \(t\) fixed
and \(n\to\infty\). The structure follows Section~\ref{sec:RRM}:
Section~\ref{subsec:RGDclt} gives the central limit theorem,
Section~\ref{subsec:RGDcov} estimates its covariance, and
Section~\ref{subsec:infRGD} gives confidence regions for
\(\theta_{\mathrm{PS}}\). We use \(V_t\), \(\widehat V_t\), and
\(\widehat{\mathcal R}_t\) for the RGD counterparts of the quantities in
Section~\ref{sec:RRM}.

\subsection{Central Limit Theorem of RGD}\label{subsec:RGDclt}
For \(\theta\in\Theta\), define the population gradient
$
S(\theta):=\mathbb E_{Z\sim\mathcal D(\theta)}[\nabla\ell(Z;\theta)],
$
so that the RGD update in Section~\ref{subsec:bgrgd} is
\(\theta_{t+1}=\theta_t-\alpha S(\theta_t)\). Subtracting it from the REGD
update gives
\[
\widehat\theta_{t+1}-\theta_{t+1}
=-\alpha\Bigl\{\frac1n\sum_{i=1}^n\nabla\ell(Z_{t,i};\widehat\theta_t)
-S(\widehat\theta_t)\Bigr\}
+\bigl[\{\widehat\theta_t-\alpha S(\widehat\theta_t)\}
-\{\theta_t-\alpha S(\theta_t)\}\bigr].
\]
The first term is the sampling error of the current round. Conditional on
all preceding observations and estimators, \(\widehat\theta_t\) is fixed and
\(\nabla\ell(Z_{t,1};\widehat\theta_t),\ldots,\nabla\ell(Z_{t,n};\widehat\theta_t)\)
are i.i.d.\ with mean \(S(\widehat\theta_t)\). Unlike the RERM case,  we could directly apply   the central limit theorem to get  the approximate
conditional distribution of \(\sqrt n\) times the first term 
\(\mathcal N\bigl(0,\alpha^2\operatorname{Cov}_{Z\sim\mathcal D(\widehat\theta_t)}
(\nabla\ell(Z;\widehat\theta_t))\bigr)\), and this covariance tends to
\(\alpha^2\operatorname{Cov}_{Z\sim\mathcal D(\theta_t)}(\nabla\ell(Z;\theta_t))\)
as \(\widehat\theta_t\to\theta_t\). The second term is the effect of the
preceding estimation error on the population update. By
Assumption~\ref{as:consistent2}(b), \(S\) is continuously differentiable, so
the second term is approximately
\((I_d-\alpha\nabla S(\theta_t))(\widehat\theta_t-\theta_t)\).

\begin{theorem}[Central limit theorem for REGD]\label{thm:cltrgd}
Under Assumptions~\ref{as:pp} and~\ref{as:consistent2}, for each fixed
\(t\geq1\),
\[
\sqrt n(\widehat\theta_t-\theta_t)\xrightarrow{d}\mathcal N(0,V_t),
\]
where \(V_0=0\) and
$
V_{t+1}=(I_d-\alpha\nabla S(\theta_t))\,V_t\,(I_d-\alpha\nabla S(\theta_t))^\top
+\alpha^2\operatorname{Cov}_{Z\sim\mathcal D(\theta_t)}
\bigl(\nabla\ell(Z;\theta_t)\bigr), t\geq0.
$
\end{theorem}
The recursion has the form of Theorem~\ref{thm:cltrrm}, with
\(\nabla G(\theta_t)\) replaced by the Jacobian
\(I_d-\alpha\nabla S(\theta_t)\) of the RGD update and
\(\Sigma_{\theta_t}(\theta_{t+1})\) replaced by the covariance of one gradient
step.

\subsection{Estimation of the Asymptotic Covariance}\label{subsec:RGDcov}
The covariance
\(\operatorname{Cov}_{Z\sim\mathcal D(\theta_t)}(\nabla\ell(Z;\theta_t))\) can
be estimated from batch \(t\). The Jacobian \(\nabla S(\theta_t)\) also
depends on how the distribution changes with the parameter, since \(\theta\)
enters \(S(\theta)\) both through the loss and through \(\mathcal D(\theta)\).
Under Assumptions~\ref{as:pp} and~\ref{as:consistent2}, differentiating both
gives
\[
\nabla S(\theta)=H_\theta(\theta)
+\mathbb E_{Z\sim\mathcal D(\theta)}
\bigl[\nabla\ell(Z;\theta)\,\nabla_\theta\log p(Z;\theta)^\top\bigr],
\]
where the second term is computed as in
Proposition~\ref{prop:differentiability}. It involves the same score as
\(\nabla G\). As in Section~\ref{subsec:RRMcov}, we use the score directly
when it is known; otherwise we replace it by
\(s(\cdot,\widehat\theta_t;\widehat\psi_{n,k}(\widehat\theta_t))\), where
\(\widehat\psi_{n,k}(\widehat\theta_t)\) is computed from batch \(t\) and the
perturbed batches at \(\widehat\theta_t+\eta e_j\), \(j\in[d]\), as in
\eqref{eq:smobjective}.

Using batch \(t\), define
\[
\widehat{\nabla S}_t
:=\frac1n\sum_{i=1}^n\Bigl\{\nabla^2\ell(Z_{t,i};\widehat\theta_t)
+\nabla\ell(Z_{t,i};\widehat\theta_t)\,
s\bigl(Z_{t,i},\widehat\theta_t;\widehat\psi_{n,k}(\widehat\theta_t)\bigr)^\top
\Bigr\}.
\]
Following Theorem~\ref{thm:cltrgd}, set \(\widehat V_0=0\) and, for
\(t\geq0\),
\begin{equation}\label{eq:rgdempvar}
\begin{aligned}
\widehat V_{t+1}
:={}&(I_d-\alpha\widehat{\nabla S}_t)\,\widehat V_t\,
(I_d-\alpha\widehat{\nabla S}_t)^\top\\
&+\frac{\alpha^2}n\sum_{i=1}^n
\nabla\ell(Z_{t,i};\widehat\theta_t)\nabla\ell(Z_{t,i};\widehat\theta_t)^\top
-(\widehat\theta_{t+1}-\widehat\theta_t)(\widehat\theta_{t+1}-\widehat\theta_t)^\top.
\end{aligned}
\end{equation}
By the REGD update, \(\widehat\theta_t-\widehat\theta_{t+1}\) is \(\alpha\)
times the average of the gradients \(\nabla\ell(Z_{t,i};\widehat\theta_t)\),
so the last two terms equal \(\alpha^2\) times their sample covariance.

\begin{theorem}\label{thm:rgdperturb}
Under the conditions of Theorem~\ref{thm:discuss2}, for each fixed
\(t\geq0\),
\[
\|\widehat{\nabla S}_t-\nabla S(\theta_t)\|_2^2
=O_P\Bigl(\frac1{\sqrt n}+\frac1{\eta\sqrt{\min(n,k)}}+\eta+a_{n,k}\Bigr).
\]
\end{theorem}
The rate and the origins of its four terms are the same as in
Theorem~\ref{thm:discuss2}.

\begin{corollary}\label{col:rgdfeasible}
Under the conditions of Theorem~\ref{thm:rgdperturb},
\(\widehat V_t\xrightarrow{P}V_t\) for each fixed \(t\geq1\). If, in
addition, \(V_t\) is nonsingular, then the region \(\widehat{\mathcal R}_t\)
of Corollary~\ref{col:rrmfeasible}, computed from the REGD iterate
\(\widehat\theta_t\) and \(\widehat V_t\) in \eqref{eq:rgdempvar}, satisfies
\(\lim_{n\to\infty}\mathbb P(\theta_t\in\widehat{\mathcal R}_t)=1-\delta\).
\end{corollary}
 
\subsection{Confidence Regions for the Performative Stable Point}\label{subsec:infRGD}
As in Section~\ref{subsec:infRRM}, we enlarge \(\widehat{\mathcal R}_t\) by a
bound on \(\|\theta_t-\theta_{\mathrm{PS}}\|_2\). Define
\[
\rho_{\mathrm{RGD}}
:=1-\alpha\beta\Bigl\{\frac{\gamma}{\beta+\gamma}
-\varepsilon(1+1.5\alpha\beta)\Bigr\}.
\]
If \(\varepsilon<\gamma/\{(\beta+\gamma)(1+1.5\alpha\beta)\}\) and
\(\alpha\leq2/(\beta+\gamma)\), then \(\varepsilon\beta<\gamma\), so
\(\theta_{\mathrm{PS}}\) is well defined, and
\(\rho_{\mathrm{RGD}}\in(0,1)\). In this case,
\citet[Theorem~3.8]{perdomo2020performative} gives
\(\|\theta_t-\theta_{\mathrm{PS}}\|_2
\leq\rho_{\mathrm{RGD}}^t\|\theta_0-\theta_{\mathrm{PS}}\|_2\).

\begin{corollary}[Confidence region for \(\theta_{\mathrm{PS}}\)]\label{col:rgdps}
Suppose the conditions of Corollary~\ref{col:rgdfeasible} hold,
\(\varepsilon<\gamma/\{(\beta+\gamma)(1+1.5\alpha\beta)\}\),
\(\alpha\leq2/(\beta+\gamma)\), and
\(\|\theta_0-\theta_{\mathrm{PS}}\|_2\leq B\). Then
\[
\liminf_{n\to\infty}\mathbb P\bigl\{\theta_{\mathrm{PS}}\in
\widehat{\mathcal R}_t+\mathcal B(0,B\rho_{\mathrm{RGD}}^t)\bigr\}
\geq1-\delta.
\]
\end{corollary}
The bound on \(B\) after Corollary~\ref{col:rrmps} applies here, since it
does not depend on the algorithm. Compared with Corollary~\ref{col:rrmps},
only the contraction factor changes, from \(\varepsilon\beta/\gamma\) to
\(\rho_{\mathrm{RGD}}\). Since \(\rho_{\mathrm{RGD}}\geq1/2\), whereas \(\varepsilon\beta/\gamma\) can be
close to zero, RGD may need more rounds than RRM before the bias is
negligible.

%% file: Sections/experiment.tex
\section{Experiment}\label{sec:exp}
In this section, we provide empirical support for our theoretical results through three experimental settings. More experimental details, as well as additional experiments  on the benefits of score matching and PPI, are deferred to   Appendix~\ref{app:expjmlr}.


\subsection{Controlled Synthetic Studies}\label{sec:puresyntheticexp}
We first consider a   synthetic setting motivated by a stylized lending market, where prediction-based lending decisions affect borrower participation. The lender sets interest rates based on predicted default risk. If lower-risk borrowers are more sensitive to borrowing costs, higher rates disproportionately discourage their participation, increasing the share of higher-risk borrowers. Through this selection response mechanism, the deployed predictor shapes the population used for subsequent training. Formally, let $\mathcal X=[0,1]$ denote the space of borrower risk score and $\mathcal Y=\{0,1\}$ the outcome space, with $Y=1$ indicating default. The instance space is $\mathcal Z=\mathcal X\times\mathcal Y$. We model the predicted default probability as $h_\theta(x)=\sigma(\theta x)$, where $\theta\in\Theta:=\mathbb R$ is a scalar parameter and $\sigma(u)=(1+e^{-u})^{-1}$ is the sigmoid function.

We capture this mechanism through a two-component mixture. 
Specifically, deployment of \(\theta\) induces risk score distribution
$X\sim(1-a(\theta))\,\mathrm{Unif}(0,1)+a(\theta)\,B(5,2),$ where $a(\theta)=0.8\sigma(\kappa(\theta-b))$ is the deployment-dependent mixture weight of the high-risk borrower component, and we set the conditional distribution of binary label $Y$ to be
$Y\mid X=x\sim\mathrm{Ber} \big(\sigma(-1+2x)\big).$ These distributions jointly define the distribution map $\cD(\theta)$.
Thus, deployment changes the borrower risk score distribution: the beta component has higher risk characteristics on average, and its weight increases with $\theta$, while leaving the conditional outcome distribution fixed. 
The predictor is fitted using regularized logistic loss,
$$\ell((x,y);\theta)
=\log(1+e^{\theta x})-y\theta x+\frac{\tau}{2}\|\theta\|_2^2,$$
where $\tau>0$ is a regularization parameter.  Our inference target is the population iterate $\theta_{t}$ and performative stable point $\theta_{\text{PS}}$.

\subsubsection{Mild Performativity Feedback}
We first consider a setting with mild performative feedback to examine the roles of repeated retraining and uncertainty propagation in inference for the performative stable point.
We initialize both algorithms at $\theta_0=1$, set $\kappa=\tau=1$, $b = 0$, and use step size $\alpha=0.5$ for RGD. Each deployment supplies $n=2000$ fresh labeled observations per round, and we conduct 10,000 independent Monte Carlo repetitions for each method. We build confidence intervals for population trajectory and performative stable point using Corollary~\ref{col:rrmfeasible}, \ref{col:rrmps}, \ref{col:rgdfeasible}, \ref{col:rgdps}, and evaluate coverage and interval width at fixed reporting round, where the population trajectories $\{\theta_t\}_{t=1}^{T}$ and the PS point $\theta_{\mathrm{PS}}$ are numerically computed from the specified distribution and serve as evaluation references. {All experiments include two baselines. First, we consider a one-step ERM baseline that fits the predictor once using observations from $\mathcal D(\theta_0)$ entirely without accounting for performativity. Its estimator coincides with the first RRM update, and its conventional confidence interval targets $\theta_1$ without an adjustment for the discrepancy between $\theta_1$ and $\theta_{\mathrm{PS}}$. Second, we consider a per-round M-estimator baseline that constructs
each interval using only the uncertainty estimate from the current
round's M-estimator. It uses the same RRM point estimates and update
samples, but replaces the recursive covariance estimator with an
empirical estimate of the current-round covariance
$\Sigma_{\theta_t}(\theta_{t+1})$. This is equivalent to setting $\widehat{\nabla G}_t=0$ in the covariance recursion, thereby omitting uncertainty inherited from earlier rounds through their effects on subsequent deployment distributions. For the performative stable point, it retains the same deterministic bias enlargement.} Details of score estimation, contraction constants, and bias bounds are provided in  Appendix~\ref{app:pureexp}.

\textit{Results.} Table~\ref{tab:pure-n2000-basis} reports the results at the reporting rounds. Empirical trajectory coverage is close to the 95\% nominal level for both RRM and RGD, and the corresponding confidence intervals for PS point have coverage slightly higher than nominal level, consistent with the conservative enlargement used to account for the remaining population-iteration bias in Corollary~\ref{col:rrmps} and \ref{col:rgdps}. The one-step ERM baseline achieves near-nominal coverage of its own target $\theta_1$ but shows significant undercoverage for the PS point. This illustrates that valid inference for a one-step ERM target does not by itself provide valid inference for the performative stable point, highlighting the need to account for the feedback induced by subsequent deployments. The per-round M-estimator, which omits uncertainty propagated from earlier rounds, achieves nearly identical coverage of both trajectory and the performative stable point at the RRM reporting round. Thus, repeated retraining remains important for approaching the performative stable point even when the contribution of propagated uncertainty is small. We next consider a controlled experiment with stronger feedback to demonstrate the importance of retaining this contribution in covariance estimation. RRM also exhibits faster empirical convergence than RGD in this setting as we have remarked at the end of Section~\ref{subsec:infRGD}.

\begin{table}[htbp]
\centering
\small
\caption{Empirical coverage and mean widths of nominal $95\%$
confidence intervals for the population iterate $\theta_t$ and
performative stable point $\theta_{\mathrm{PS}}$ in the pure-synthetic
experiment.
}
\label{tab:pure-n2000-basis}
\resizebox{\linewidth}{!}{%
\begin{tabular}{lrrrrr}
\toprule
Method
& \shortstack{Reporting\\round}
& \shortstack{Coverage of\\$\theta_t$ (\%)}
& \shortstack{Coverage of\\$\theta_{\mathrm{PS}}$ (\%)}
& \shortstack{Trajectory\\interval width}
& \shortstack{Stable-point\\interval width} \\
\midrule
One-step ERM          & 1  & 94.65 & 81.98 & 0.026 & 0.026 \\
Per-round M-estimator & 6  & 94.78 & 95.23 & 0.025 & 0.026 \\
RRM                   & 6  & 94.78 & 95.24 & 0.025 & 0.026 \\
RGD                   & 78 & 94.99 & 97.09 & 0.016 & 0.017 \\
\bottomrule
\end{tabular}%
}
\end{table}

\begin{figure}
    \centering
    \includegraphics[width=1\linewidth]{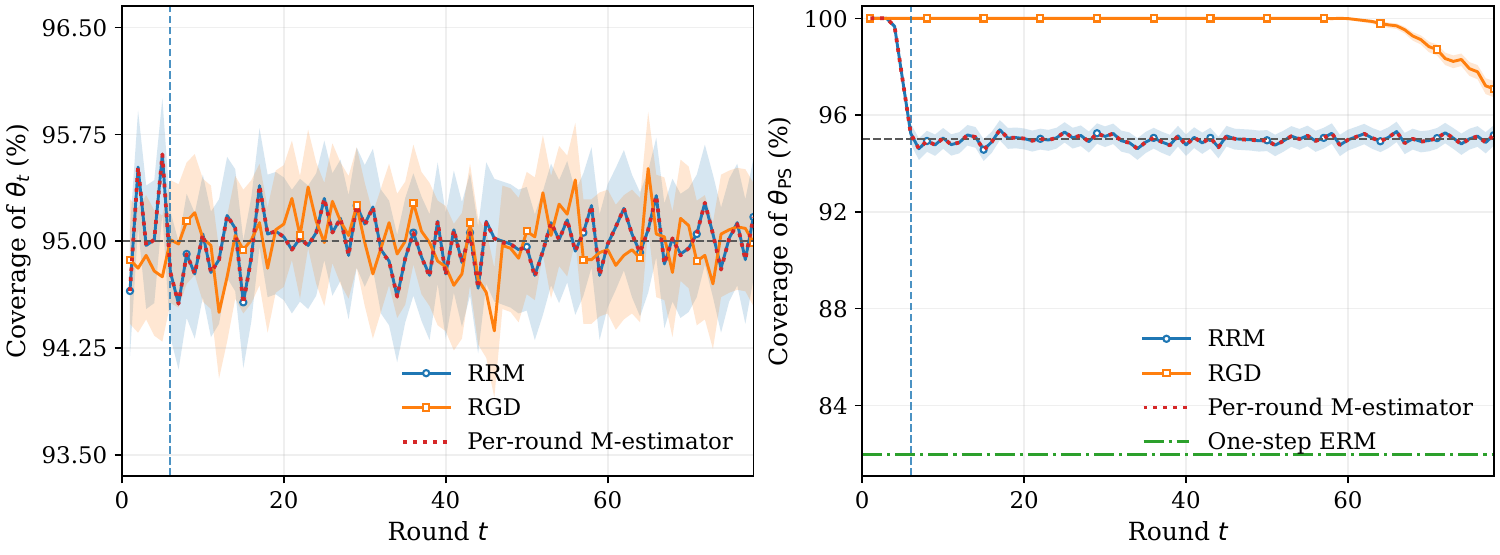}
    \caption{Pure-Synthetic Experiment. Left Panel: Empirical coverage of $\theta_{t}$ across rounds; Right Panel: Empirical coverage of performative stable point $\theta_{\text{PS}}$ across rounds. Vertical dashed line implies the fixed reporting round of RRM selected by $B_t<10^{-3}$.
    }
    \label{fig:pure-main}
\end{figure}

\subsubsection{Uncertainty Propagation under Strong Performative Feedback.}\label{subsubsec:expscore}

We next examine the necessity of accounting for uncertainty propagated through performative feedback. As shown in Section~\ref{sec:RRM}, the covariance of an RRM iterate includes the contribution $\nabla G(\theta_t)V_t\nabla G(\theta_t)^\top$ from earlier rounds, where $\nabla G(\theta_t)$ depends on the distributional score. Omitting this contribution treats each round as a separate M-estimation problem and can understate uncertainty. To study this effect under stronger feedback, we set $\kappa=80$, $b=0.05$, and the activation function becomes $a(\theta)=0.8\sigma\bigl(80(\theta-0.05)\bigr)$. Each deployment supplies $n=10{,}000$ fresh labeled observations,
and we conduct 10,000 independent Monte Carlo repetitions for
each algorithm. We use a correctly specified basis score model, whose construction, fitting procedure and an ablation on candidate score function family in this setup are provided in Appendix~\ref{app:pureexp}.

\textit{Results.}
In this stronger-feedback setting, empirical trajectory coverage is close to the nominal level for both RRM and RGD, while stable-point coverage is slightly higher, consistent with the conservative bias enlargement in
Corollaries~\ref{col:rrmps} and~\ref{col:rgdps}. The one-step ERM baseline achieves near-nominal coverage of
its own target $\theta_{1}$ but exhibits severe undercoverage for the performative stable point, highlighting the importance of repeated retraining under distributional feedback. The per-round M-estimator produces narrower intervals and exhibits substantial undercoverage for both the population trajectory and the performative stable point. Unlike the weaker-feedback setting, the contribution of propagated uncertainty is therefore substantial. These results
illustrate that approaching the performative stable point through repeated retraining does not alone ensure valid inference: the covariance estimator must also account for uncertainty accumulated through deployment feedback.

\begin{figure}
    \centering
    \includegraphics[width=1\linewidth]{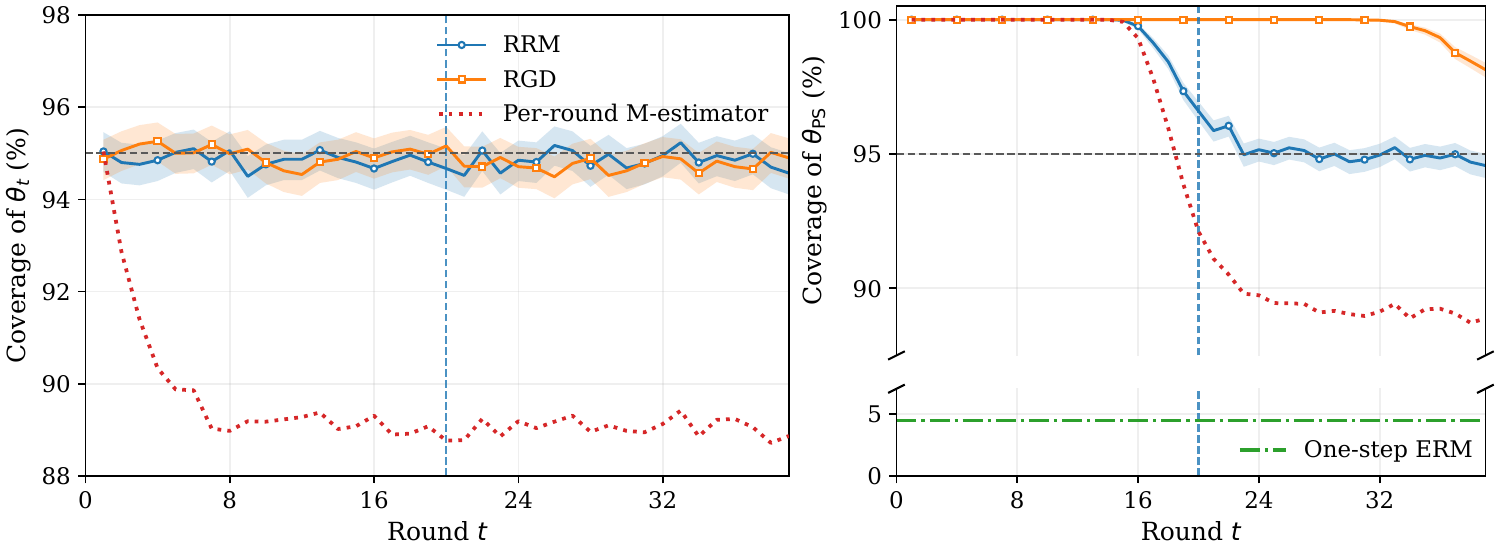}
    \caption{Pure-Synthetic Controlled Experiment with Strong Feedback. Left Panel: Empirical coverage of $\theta_{t}$ across rounds; Right Panel: Empirical coverage of performative stable point $\theta_{\text{PS}}$ across rounds. Vertical dashed line implies the fixed reporting round of RRM selected by $B_t<10^{-3}$. 
    }
    \label{fig:pure-extreme}
\end{figure}

\begin{table}[htbp]
\centering
\small
\caption{Empirical coverage and mean widths of interval in controlled synthetic experiment.}
\label{tab:pure-extreme-main}
\resizebox{\linewidth}{!}{%
\begin{tabular}{lrrrrr}
\toprule
Method
& \shortstack{Reporting\\round}
& \shortstack{Coverage of\\$\theta_t$ (\%)}
& \shortstack{Coverage of\\$\theta_{\mathrm{PS}}$ (\%)}
& \shortstack{Trajectory\\interval width}
& \shortstack{Stable-point\\interval width} \\
\midrule
One-step ERM         & 1  & 95.04 & 4.43  & 0.012 & 0.012 \\
Per-round M-estimator & 20 & 88.77 & 92.11 & 0.011 & 0.013 \\
RRM                  & 20 & 94.67 & 96.60 & 0.014 & 0.015 \\
RGD                  & 39 & 94.90 & 98.14 & 0.010 & 0.012 \\
\bottomrule
\end{tabular}%
}
\end{table}

\subsection{Semi-synthetic Experiment}\label{sec:semiexp}
We next consider a semi-synthetic setup   on the Learning Mindset dataset \citep{athey2019estimating}, an educational dataset modeling how growth-mindset interventions, which encourage students to view intellectual abilities as developable through effort and learning, will affect students’ academic achievement. 
Before intervention assignment, the deployed predictor uses students' characteristics $X$ to estimate the probability that their subsequent academic achievement will meet or exceed a predefined threshold. These predictions guide intervention allocation: students with lower predicted success probabilities are more likely to receive support. Because interventions can affect achievement, deployment changes the distribution of the outcomes the predictor aims to predict and uses for subsequent retraining, creating performative feedback.

Formally, let $\cX$ denote the space of student profiles and $\cY=\{0,1\}$ the outcome space, with $Y=1$ indicating achievement at or above a predefined academic score threshold. Let $W\in\{0,1\}$ indicate intervention assignment, with $W=1$ denoting receipt of intervention. The predictor is $h_\theta(x)=\sigma(\theta^\top\phi(x))$, where $\phi(x)\in\mathbb R^4$ is a feature representation constructed from student profiles using PCA and $\theta\in\Theta=\mathbb R^4$ is the model parameter. The predictions guide intervention assignment: students with lower predicted success probabilities are more likely to receive intervention. Specifically, intervention is assigned with probability
$$
e_\theta(x)
=\delta+(1-2\delta)\sigma\big(-\kappa\theta^\top\phi(x)\big),
$$
where $\kappa>0$ controls the responsiveness of assignment to the predictor, and $0<\delta<1/2$ ensures that both intervention and non-intervention probabilities are bounded below by $\delta$. Given student profile $X=x$ and intervention assignment $W=w$, we generate final achievement $Y\sim\text{Ber}(p(x,w))$, where $p(x,w)$ is the success probability supplied by a neural simulator fitted to the data and held fixed throughout the experiment. Thus, changing the predictor changes intervention assignment and therefore changes the conditional outcome distribution the predictor aims to learn via intervention effect. The predictor is trained using the regularized logistic loss after each deployment.
\[
\ell((x,y);\theta)
=
\log\!\left(1+e^{\theta^\top\phi(x)}\right)
-y\theta^\top\phi(x)
+\frac{\tau}{2}\|\theta\|_2^2,
\qquad \tau=1.
\]
Each deployment supplies $n=2000$ fresh labeled observations. In this experiment, for covariance estimation, we use the analytic distributional score computed from the known intervention rule $e_\theta(X)$ and the recorded intervention assignments $W$. The required population moments are estimated from the observed deployment samples.

\begin{figure}
    \centering
    \includegraphics[width=\linewidth]{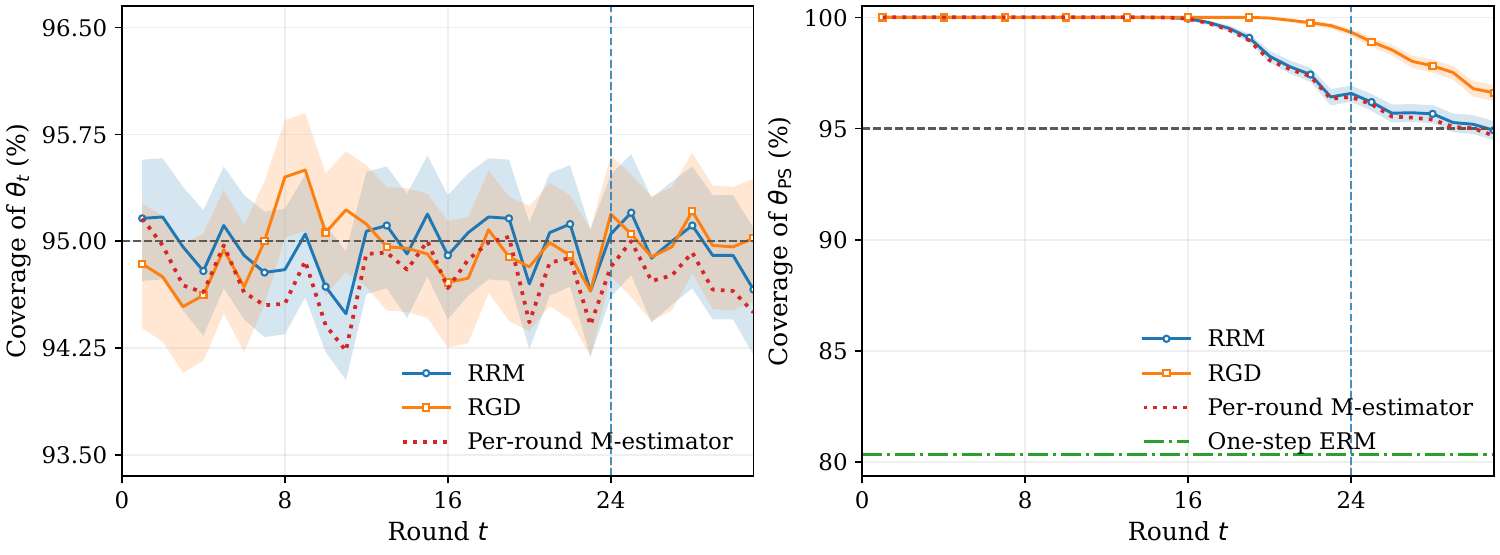}
    \caption{Semi-Synthetic Experiment. Left Panel: Empirical coverage of $\theta_{t}$ across rounds; Right Panel: Empirical coverage of performative stable point $\theta_{\text{PS}}$ across rounds. Vertical dashed line implies the fixed reporting round of RRM selected by $B_t<10^{-3}$. 
    }
    \label{fig:semi-main}
\end{figure}

\textit{Results.} Table~\ref{tab:semi-n2000-exact} reports coverage and mean
Euclidean diameters at fixed reporting rounds: round $24$ for RRM and round $31$ for RGD, applied uniformly across all $10000$ Monte Carlo repetitions. Trajectory coverage is close to the nominal $95\%$ level and the confidence regions for performative stable point attain coverage slightly larger than nominal level due to conservative bias-aware enlargement. The one-step ERM baseline achieves near-nominal coverage of its own target $\theta_1$ but substantially undercovers the performative stable point. The per-round M-estimator yields results close to those of RRM, with slightly smaller confidence regions and lower empirical coverage, consistent with its omission of uncertainty propagated from earlier rounds. Figure~\ref{fig:semi-main} complements these summaries by showing coverage across deployment rounds.

\begin{table}[htbp]
\centering
\small
\caption{Empirical coverage and mean Euclidean diameters of nominal $95\%$
confidence regions for the population iterate $\theta_t$ and
performative stable point $\theta_{\mathrm{PS}}$ in the semi-synthetic
experiment. }
\label{tab:semi-n2000-exact}
\resizebox{\linewidth}{!}{%
\begin{tabular}{lrrrrr}
\toprule
Method
& \shortstack{Reporting\\round}
& \shortstack{Coverage of\\$\theta_t$ (\%)}
& \shortstack{Coverage of\\$\theta_{\mathrm{PS}}$ (\%)}
& \shortstack{Trajectory\\region diameter}
& \shortstack{Stable-point\\region diameter} \\
\midrule
One-step ERM          & 1  & 95.16 & 80.35 & 0.046 & 0.046 \\
Per-round M-estimator & 24 & 94.82 & 96.42 & 0.046 & 0.048 \\
RRM                   & 24 & 95.05 & 96.58 & 0.046 & 0.048 \\
RGD                   & 31 & 95.02 & 96.61 & 0.043 & 0.045 \\
\bottomrule
\end{tabular}%
}
\end{table}

\subsection{Self-improving LLM routing}\label{sec:llmexp}
Finally, we consider a self-evolving model routing task on real datasets consisting of mathematical word problems benchmark GSM8K \citep{cobbe2021training} and commonsense knowledge benchmark CommonsenseQA \citep{talmor-etal-2019-commonsenseqa}. We use a frozen Qwen2.5-1.5B model with two generation modes: fast mode and slow mode. 

We consider a stylized model of self-evolving routing based on Best-of-\(N\) (BoN) selection. Given an input question, the deployed router independently routes the question to a mode for each of $N$ candidates and collects the response. We evaluate the responses using a fixed utility and assign the winning mode as the supervision label. The router is then updated to learn to predict this label from the input. Performativity arises because deployment affects the supervision label distribution, thereby altering the distribution of the winning-mode labels the router aims to predict.

Specifically, fix positive integer $N\ge 2$, let  $\cX$ be the problem space,  $\cA=\{0,1\}$ the  routing choice space, with routing choice $A=1$ denoting routing to slow mode. We adopt a logistic router $h_{\theta}(x)=\sigma(\theta^\top \phi(x))$, where $\phi:\mathcal{X}\to \bR^{5}$ is a five-dimensional feature vector map. The routing probability is defined by $\pi_\theta(1\mid x) =(1-\xi)h_\theta(x)+\frac{\xi}{2},$ where $\xi=0.2$ provides a positive routing probability to each mode bounded away from $0$. For each problem $X$, routing choices $A_{1},\ldots,A_{N}$ are i.i.d. sampled conditionally on $X$ using $\pi_{\theta}(\cdot|X)$ and draw the corresponding mode response $R_{j}$. The highest-scoring candidate supplies the supervision label for the next BoN update:
$J^*=\arg\max_{j\in [N]}u(X,R_j),$
where $u$ is a fixed utility function.
We fit the router to these winning-mode labels using regularized logistic loss,
\[
\ell(Z;\theta)
= w(Z)\big[\log\left(1+e^{\theta^\top\phi(X)}\right)
-A_{{J}^*} \theta^\top\phi(X)\big]
+\frac{\tau}{2}\|\theta\|_2^2,
\]
where \(Z=(X,A_1,R_1,\ldots,A_N,R_N)\) and $w(Z)\in [0,1]$ is a utility advantage weight, with larger weight indicating stronger preference to the winning mode. Changing the deployed parameter alters the candidate distribution and hence the conditional distribution of \(A_{J^*}\), even though the question population and mode-specific response distributions remain fixed. We apply RRM and RGD to this setting and evaluate inference for their population trajectories and performative stable points. For covariance estimation, we use the analytic distributional score computed from the known routing probabilities and the sampled mode
choices $A_1,\ldots,A_N$. The required population moments are estimated from the observed deployment samples.

\begin{figure}[htbp]
\centering
\includegraphics[width=\linewidth]{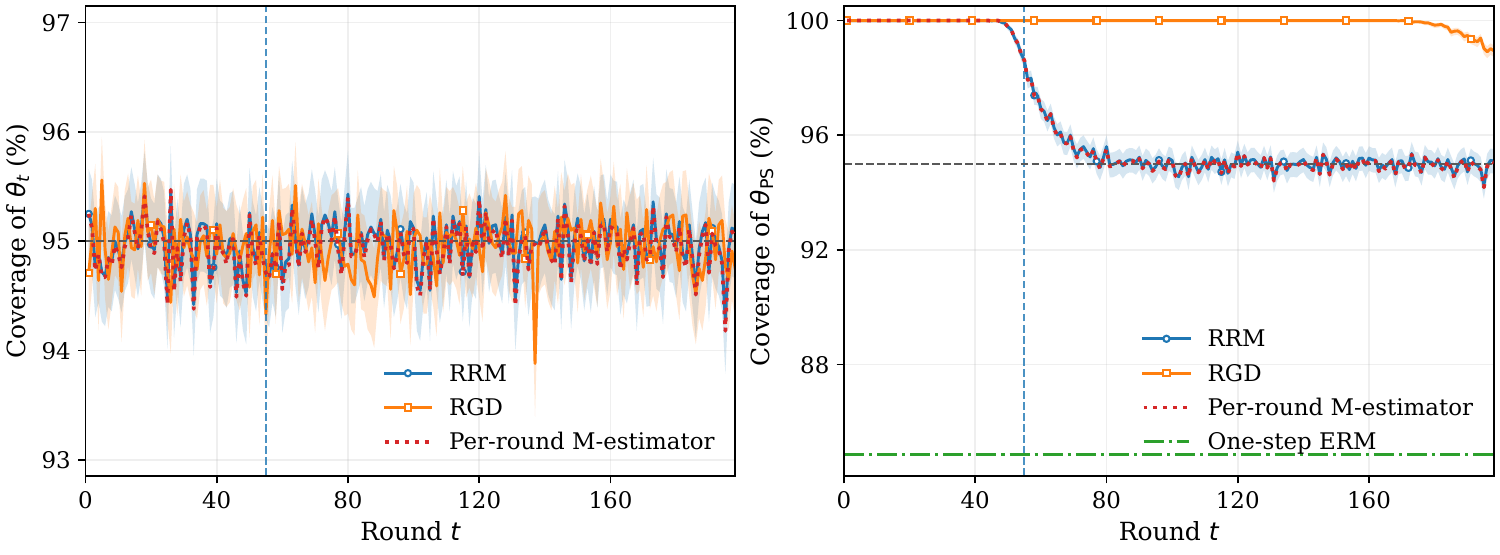}
\caption{LLM routing with Best-of-$5$ selection and $n=10000$. 
Left: empirical coverage of $\theta_t$; right: empirical coverage of
$\theta_{\mathrm{PS}}$. Both methods are shown through round $198$.
The vertical dashed line marks the fixed reporting round for RRM $55$, selected by $B_t<10^{-3}$. Horizontal dashed lines indicate nominal $95\%$
coverage; shaded bands are pointwise $95\%$ exact binomial Monte Carlo confidence intervals.}
\label{fig:llm-routing}
\end{figure}

\textit{Results.} Table~\ref{tab:llm-n2000-exact} reports coverage and mean Euclidean
diameters for best-of-$5$ routing with $n=10,000$ at fixed reporting
rounds: round $55$ for RRM and round $198$ for RGD.
Trajectory coverage is $94.74\%$ for RRM and $94.75\%$ for RGD, close to the nominal $95\%$ coverage level. The confidence regions for the performative stable point attain coverage above the nominal level, consistent with the conservative enlargement used to account for the remaining population-iteration bias. The one-step ERM baseline achieves near-nominal coverage of its own target $\theta_1$,  but significantly undercovers the performative stable point, while the per-round M-estimator yields results very close to those of RRM. Figure~\ref{fig:llm-routing} complements these summaries by
showing coverage across deployment rounds.

Table~\ref{tab:llm-routing} summarizes changes in routing probability
relative to the initial equal-mixture policy and average utility at
the performative stable point on the merged question pool.
The mean total variation distance is $3.00\%$, with routing shifting toward the slow mode, and the overall expected utility
increases from $0.483$ to $0.500$. Table~\ref{tab:llm-utility-advantage} compares the slow-mode
utility advantage for individual response draws and
best-of-$5$ candidate sets. Best-of-$5$ increases this advantage
on GSM8K but reduces it on CommonsenseQA. This pattern is
consistent with the larger routing shift and utility gain
observed on GSM8K. Detailed experimental configurations are provided in
Appendix~\ref{app:llmexp}.

\begin{table}[htbp]
\centering
\small
\caption{Empirical coverage and mean Euclidean diameters of nominal
$95\%$ confidence regions in the LLM routing experiment.}
\label{tab:llm-n2000-exact}
\resizebox{\linewidth}{!}{%
\begin{tabular}{lrrrrr}
\toprule
Method
& \shortstack{Reporting\\round}
& \shortstack{Coverage of\\$\theta_t$ (\%)}
& \shortstack{Coverage of\\$\theta_{\mathrm{PS}}$ (\%)}
& \shortstack{Trajectory\\region diameter}
& \shortstack{Stable-point\\region diameter} \\
\midrule
One-step ERM          & 1   & 95.25 & 84.86 & 0.015 & 0.015 \\
Per-round M-estimator & 55  & 94.71 & 98.60 & 0.015 & 0.016 \\
RRM                   & 55  & 94.74 & 98.61 & 0.015 & 0.016 \\
RGD                   & 198 & 94.75 & 98.95 & 0.014 & 0.016 \\
\bottomrule
\end{tabular}%
}
\end{table}

\begin{table}[htbp]
\centering
\small
\setlength{\tabcolsep}{4pt}
\caption{Routing deviation and utility in the LLM routing experiment,
evaluated at the population performative stable point.
Routing deviation is the mean total variation distance from the
initial equal-mixture policy. Utility is the expected value of
answer correctness minus a token-cost penalty of $0.1$ per
$1{,}000$ generated tokens, evaluated for one routed response-bank entry.}
\label{tab:llm-routing}
\begin{tabular}{@{}lrrrr@{}}
\toprule
Question subset
& \shortstack{Slow-mode\\routing probability (\%)}
& \shortstack{Mean routing\\deviation (\%)}
& \shortstack{Initial\\utility}
& \shortstack{Stable-point\\router utility} \\
\midrule
All questions & 52.99 & 3.00 & 0.483 & 0.500 \\
GSM8K         & 56.46 & 6.46 & 0.413 & 0.452 \\
CommonsenseQA & 50.21 & 0.22 & 0.539 & 0.539 \\
\bottomrule
\end{tabular}
\end{table}

\begin{table}[htbp]
\centering
\small
\setlength{\tabcolsep}{4pt}
\caption{Utility comparison. The single-draw advantage is mean slow utility minus mean fast utility. The Best-of-$5$ advantage is the expected difference between the best slow and fast candidate utilities under the initial equal-mixture policy, conditional on both modes appearing. }
\label{tab:llm-utility-advantage}
\begin{tabular}{@{}lrrrr@{}}
\toprule
Dataset
& \shortstack{Mean fast\\utility}
& \shortstack{Mean slow\\utility}
& \shortstack{Single-draw\\advantage}
& \shortstack{Best-of-$5$\\advantage} \\
\midrule
GSM8K         & 0.1125 & 0.7128 & 0.6003 & 0.6535 \\
CommonsenseQA & 0.5074 & 0.5709 & 0.0635 & 0.0186 \\
\bottomrule
\end{tabular}
\end{table}

\subsection{Feedback Strength and Uncertainty Propagation}
\label{sec:exp-feedback}

The experiments illustrate two consequences of performative feedback
for inference. First, a predictor fitted to the initial deployment
distribution need not coincide with the performative stable point,
as demonstrated by the one-step ERM comparisons. Second, repeated
retraining propagates estimation errors through subsequent deployment
distributions. For RRM, this propagation is governed by $\nabla G$,
which contributes
$\nabla G(\theta_t)V_t\nabla G(\theta_t)^\top$
to the covariance of the next iterate. The per-round M-estimator
omits this positive semidefinite contribution and can therefore
underestimate uncertainty, even after the population trajectory has
approached the performative stable point. For RGD, uncertainty
propagates through the corresponding update Jacobian
$I_d-\alpha\nabla S(\theta_t)$.

The importance of this correction varies across experimental
settings. In the synthetic experiment with $\kappa=1$ and the two
practical settings, the RRM map has relatively small local
derivatives. The proposed method consequently produces confidence
regions close in diameter to those of the per-round M-estimator,
while accounting for uncertainty from earlier rounds. Under the
stronger feedback induced by $\kappa=80$, omitting this contribution
leads to substantial undercoverage, and incorporating the score term
markedly improves calibration. These findings explain why repeated
retraining can remain important for approaching the performative
stable point even when the additional covariance correction is small.

In applications, the strength of this feedback is generally unknown. Even when the score is analytically available, as in our semi-synthetic and LLM evolving experiments, the Jacobian remains unknown because it depends on population moments involving the loss, which we have to estimate from observed data. The estimator developed in Section~\ref{subsec:RRMcov} consistently estimates $\nabla G$ under the stated regularity and score-model richness assumptions, allowing this sensitivity to be assessed from data. Table~\ref{tab:feedback-jacobian} evaluates the estimator used at each RRM reporting round against the population derivative at the same realized deployment. For each reporting round $T$, we report the population Jacobian norm $\|\nabla G(\widehat{\theta}_{T-1})\|_2$ averaged across repetitions, together with the Jacobian estimation RMSE. The norm summarizes local feedback strength near the performative stable point, whereas the RMSE measures estimation accuracy. Here the score specification and Jacobian RMSE refer to the proposed estimator; coverage refers to the per-round M-estimator with the same deterministic bias enlargement. This sensitivity governs the covariance contribution $\nabla G(\theta_{T-1})V_{T-1}\nabla G(\theta_{T-1})^\top$ from uncertainty inherited at round $T$. These diagnostics indicate relatively mild feedback in the $\kappa=1$ synthetic experiment and the two practical settings, and stronger feedback in the $\kappa=80$ experiment. All four configurations substantially improve on the zero-Jacobian baseline in Jacobian estimation RMSE. We defer the experimental details to Appendix~\ref{app:jacobian}.

\begin{table}[htbp]
\centering
\small
\caption{Feedback strength, Jacobian estimation error, and coverage
of the performative stable point by the per-round M-estimator,
using 10,000 Monte Carlo repetitions. }
\label{tab:feedback-jacobian}
\resizebox{\linewidth}{!}{%
\begin{tabular}{llrrrr}
\toprule
Setting
& Score
& Round
& \shortstack{Mean population\\Jacobian norm}
& \shortstack{Jacobian\\RMSE}
& \shortstack{Per-round M-estimator\\coverage of $\theta_{\mathrm{PS}}$ (\%)} \\
\midrule
Synthetic ($\kappa=1$)
& Basis & 6 & 0.0079 & 0.0061 & 95.23 \\
Synthetic ($\kappa=80$)
& Basis & 20 & 0.5719 & 0.0329 & 92.11 \\
Semi-synthetic
& Analytic & 24 & 0.1312 & 0.0441 & 96.42 \\
LLM routing
& Analytic & 55 & 0.0661 & 0.0105 & 98.60 \\
\bottomrule
\end{tabular}%
}
\end{table}
Our estimator of $\nabla G$ therefore provides a policy designer with information about how changes in the deployed predictor affect the next population update. The same estimate enters the covariance recursion, accounting for feedback without requiring its strength to be known in advance. In the settings with mild feedback, this correction entails little additional interval width; when feedback is stronger, it addresses an important source of undercoverage.  These estimates characterize local sensitivity at the observed deployments, while the uniform contraction bounds used for stable-point bias control are established separately.

%% file: Sections/conclusion.tex
\section{Conclusion}
In this paper, we study statistical inference under performativity. We derive asymptotic distributions for the RRM and RGD trajectories and construct confidence regions for the performative stable point. We further extend prediction-powered inference to the dynamic setting under performativity. Our confidence regions for $\theta_{\mathrm{PS}}$ account for the remaining iteration bias; direct inference methods for $\theta_{\mathrm{PS}}$ are left for future work. We evaluate our framework in synthetic, semi-synthetic educational, and self-evolving LLM routing settings, demonstrating the practical need to account for performativity in statistical inference.

\section{Acknowledgements}
We would like to acknowledge the helpful discussion and suggestions from Tijana Zrinc, Juan Perdomo, Anastasios Angelopoulos, and Steven Wu.

%% file: Sections/App_A.tex

\appendix
\tableofcontents
\newpage

\input{Sections/method2}

\section{Theoretical Details}\label{app:tech}
This appendix gives the assumptions and proofs omitted from
Sections~\ref{sec:RRM}--\ref{sec:ppi}. We first prove the RERM limit theory
and the consistency of its covariance estimator, and then give the
corresponding arguments for RGD and PPI.

When \(\widetilde\theta\) is fixed conditional on the preceding
observations, \(\widehat{\mathbb E}_{\widetilde\theta,m}\) denotes the
empirical average of \(m\) conditionally i.i.d. observations from
\(\mathcal D(\widetilde\theta)\).

\input{Sections/appA_vanilla}
\input{Sections/appA_score}
\input{Sections/appA_perturb}
\input{Sections/appA_rgd}

\input{Sections/appA_PPI}

%% file: Sections/method2.tex



\section{Prediction-Powered Inference under Performativity}\label{sec:ppi}

In this section, we extend prediction-powered inference (PPI) to this dynamic performative setting. First, Section~\ref{subsec:bgppi} introduces the setup of PPI under performativity, Section~\ref{subsec:PPIclt} gives the central limit
theorem for population trajectory, Section~\ref{subsec:PPIcov} describes a weight tuning procedure and an estimator of the covariance, and Section~\ref{subsec:infPPI} gives confidence regions for
\(\theta_{\mathrm{PS}}\). We retain the notation in Section~\ref{sec:bg}.

\subsection{Setup of PPI under Performativity}\label{subsec:bgppi}

Prediction-powered inference (PPI) combines labeled observations with additional
unlabeled observations and a prediction rule \(f:\mathcal X\to\mathcal Y\)
\citep{angelopoulos2023ppi++}. We treat \(f\) as fixed throughout and set
\(\widehat\theta_0^{\mathrm{PPI}}=\theta_0\). At round \(t\), conditional
on all preceding observations and estimators, let
\[
\{(X_{t,i},Y_{t,i})\}_{i=1}^{n_t}
\quad\text{and}\quad
\{X^u_{t,j}\}_{j=1}^{N_t}
\]
be independent samples from
\(\mathcal D(\widehat\theta_t^{\mathrm{PPI}})\) and its
\(X\)-marginal, respectively. For \(\lambda\in[0,1]\), define
\[
\widehat\theta_{t+1}^{\mathrm{PPI}}(\lambda)
:=\arg\min_{\theta\in\Theta}
\left[
\frac1{n_t}\sum_{i=1}^{n_t}
\bigl\{\ell(X_{t,i},Y_{t,i};\theta)
-\lambda\ell(X_{t,i},f(X_{t,i});\theta)\bigr\}
+\frac{\lambda}{N_t}\sum_{j=1}^{N_t}
\ell(X^u_{t,j},f(X^u_{t,j});\theta)
\right].
\]

Given a data-dependent \(\widehat\lambda_{t+1}\in[0,1]\), define
$
\widehat\theta_{t+1}^{\mathrm{PPI}}
:=\widehat\theta_{t+1}^{\mathrm{PPI}}
(\widehat\lambda_{t+1}).
$
For fixed \(\lambda\), the conditional population minimizer is
\(G(\widehat\theta_t^{\mathrm{PPI}})\), so the deterministic population trajectory remains
\(\theta_{t+1}=G(\theta_t)\).  Setting \(\widehat\lambda_{t+1}=0\) gives the empirical RRM update.

In the following context, we write \(\widehat\theta_t\) and
\(\widehat\theta_t(\lambda)\) for \(\widehat\theta_t^{\mathrm{PPI}}\) and
\(\widehat\theta_t^{\mathrm{PPI}}(\lambda)\), and use \(V_t\),
\(\widehat V_t\), and \(\widehat{\mathcal R}_t\) for the PPI counterparts of
the quantities in Section~\ref{sec:RRM}.


\subsection{Central Limit Theorem of PPI}\label{subsec:PPIclt}

As noted in Section~\ref{subsec:bgppi}, for a fixed weight \(\lambda\), the
conditional population minimizer of the PPI objective at round \(t\) is
\(G(\widehat\theta_t)\). The decomposition in Section~\ref{subsec:bgrrm}
therefore still applies,
$
\widehat\theta_{t+1}-\theta_{t+1}
=\bigl(\widehat\theta_{t+1}-G(\widehat\theta_t)\bigr)
+\bigl(G(\widehat\theta_t)-G(\theta_t)\bigr),
$
and the second term is the same as for RERM. Only the first term changes.
For \(\widetilde\theta,\theta\in\Theta\) and \(\lambda\in[0,1]\), define
\[
\begin{aligned}
\Sigma_{\widetilde\theta}(\theta;\lambda)
:={}&H_{\widetilde\theta}(\theta)^{-1}
\bigl\{\operatorname{Cov}_{Z\sim\mathcal D(\widetilde\theta)}
\bigl(\nabla\ell(Z;\theta)-\lambda\nabla\ell(X,f(X);\theta)\bigr)\\
&+r\lambda^2\operatorname{Cov}_{Z\sim\mathcal D(\widetilde\theta)}
\bigl(\nabla\ell(X,f(X);\theta)\bigr)\bigr\}
H_{\widetilde\theta}(\theta)^{-1},
\end{aligned}
\]
where \(Z=(X,Y)\). By the M-estimation results of
\citet{angelopoulos2023ppi++}, for a fixed \(\lambda\), the conditional
distribution of \(\sqrt n\{\widehat\theta_{t+1}(\lambda)-G(\widehat\theta_t)\}\)
is approximately
\(\mathcal N\bigl(0,\Sigma_{\widehat\theta_t}(G(\widehat\theta_t);\lambda)\bigr)\).
The two covariances come from the labeled observations and from the
unlabeled covariates, respectively. When \(\lambda=0\), the unlabeled
covariates are not used and
\(\Sigma_{\widetilde\theta}(\theta;0)=\Sigma_{\widetilde\theta}(\theta)\).

\begin{theorem}[Central limit theorem for PPI]\label{thm:ppiclt}
Suppose Assumptions~\ref{as:pp}, \ref{as:consistent2}, and~\ref{as:ppias2}
hold, \(n/N\to r\in[0,\infty)\), and \(\widehat\lambda_t\xrightarrow{P}\lambda_t\)
for each \(t\geq1\), where \(\lambda_t\in[0,1]\) are deterministic. Then for
each fixed \(t\geq1\),
\[
\sqrt n(\widehat\theta_t-\theta_t)\xrightarrow{d}\mathcal N(0,V_t),
\]
where \(V_0=0\) and
$
V_{t+1}=\nabla G(\theta_t)\,V_t\,\nabla G(\theta_t)^\top
+\Sigma_{\theta_t}(\theta_{t+1};\lambda_{t+1}),  t\geq0.
$
\end{theorem}
Assumption~\ref{as:ppias2} makes the PPI objective strongly convex for every
\(\lambda\in[0,1]\) and imposes the moment conditions of
Assumption~\ref{as:pp} on the loss at the predicted labels. The recursion is
that of Theorem~\ref{thm:cltrrm} with \(\Sigma_{\theta_t}(\theta_{t+1})\)
replaced by \(\Sigma_{\theta_t}(\theta_{t+1};\lambda_{t+1})\); setting
\(\widehat\lambda_t=0\) recovers Theorem~\ref{thm:cltrrm}.

\subsection{Choice of the Weights and Covariance Estimation}\label{subsec:PPIcov}
Following \citet{angelopoulos2023ppi++}, we choose the weight in each round
to minimize the trace of the covariance added in that round,
\[
\lambda_{t+1}^*:=\argmin_{\lambda\in[0,1]}
\operatorname{Tr}\Sigma_{\theta_t}(\theta_{t+1};\lambda),\qquad t\geq0.
\]
Since the inherited term in Theorem~\ref{thm:ppiclt} does not depend on
\(\lambda_{t+1}\), \(\lambda_{t+1}^*\) also minimizes
\(\operatorname{Tr}V_{t+1}\) given \(\lambda_1,\ldots,\lambda_t\). Since
\(\lambda=0\) is allowed,
\(\operatorname{Tr}\Sigma_{\theta_t}(\theta_{t+1};\lambda_{t+1}^*)\leq
\operatorname{Tr}\Sigma_{\theta_t}(\theta_{t+1})\): the covariance added in
each round has no larger trace than for RERM.

We estimate \(\lambda_{t+1}^*\) from batch \(t\). Expand the first
covariance in \(\Sigma_{\widetilde\theta}(\theta;\lambda)\) into the
covariance of \(\nabla\ell(Z;\theta)\), its cross-covariances with
\(\nabla\ell(X,f(X);\theta)\), and \(\lambda^2\) times the covariance of
\(\nabla\ell(X,f(X);\theta)\). For \(\theta\in\Theta\) and
\(\lambda\in[0,1]\), let \(\widehat\Sigma_{\widehat\theta_t}(\theta;\lambda)\)
be obtained from \(\Sigma_{\widehat\theta_t}(\theta;\lambda)\) by replacing
\(H_{\widehat\theta_t}(\theta)\) with the average Hessian
\(n^{-1}\sum_{i=1}^n\nabla^2\ell(Z_{t,i};\theta)\), \(r\) with \(n/N\), and
each covariance with its sample version: those involving
\(\nabla\ell(Z;\theta)\) use the \(n\) labeled observations of batch \(t\),
and the covariance of \(\nabla\ell(X,f(X);\theta)\) uses all \(n+N\)
covariates \(X_{t,i}\) and \(X^u_{t,j}\). Choose any
\(\widetilde\lambda_{t+1}\in[0,1]\) and compute the preliminary estimator
\(\widehat\theta_{t+1}(\widetilde\lambda_{t+1})\). We set
\[
\widehat\lambda_{t+1}:=\argmin_{\lambda\in[0,1]}
\operatorname{Tr}\widehat\Sigma_{\widehat\theta_t}
\bigl(\widehat\theta_{t+1}(\widetilde\lambda_{t+1});\lambda\bigr),
\qquad
\widehat\theta_{t+1}:=\widehat\theta_{t+1}(\widehat\lambda_{t+1}).
\]
Both traces are convex quadratic functions of \(\lambda\), so both
minimizers have closed forms (Appendix~\ref{app:resultsforppi}). Following
Theorem~\ref{thm:ppiclt}, set \(\widehat V_0=0\) and
\begin{equation}\label{eq:ppiempvar}
\widehat V_{t+1}
:=\widehat{\nabla G}_t\,\widehat V_t\,\widehat{\nabla G}_t^\top
+\widehat\Sigma_{\widehat\theta_t}(\widehat\theta_{t+1};\widehat\lambda_{t+1}),
\qquad t\geq0,
\end{equation}
where \(\widehat{\nabla G}_t\) is computed as in Section~\ref{subsec:RRMcov}
from the labeled observations of batch \(t\) and the perturbed batches.

\begin{corollary}\label{col:ppiclt}
Suppose the conditions of Theorem~\ref{thm:discuss2} and
Assumption~\ref{as:ppias2} hold, \(n/N\to r\in[0,\infty)\), and
\(\operatorname{Cov}_{Z\sim\mathcal D(\theta_{t-1})}
(\nabla\ell(X,f(X);\theta_t))\neq0\) for each \(t\geq1\). Then for each fixed
\(t\geq1\), \(\widehat\lambda_t\xrightarrow{P}\lambda_t^*\), so
Theorem~\ref{thm:ppiclt} holds with \(\lambda_t=\lambda_t^*\), and
\(\widehat V_t\xrightarrow{P}V_t\). If, in addition, \(V_t\) is nonsingular,
then the region \(\widehat{\mathcal R}_t\) of
Corollary~\ref{col:rrmfeasible}, computed from \(\widehat\theta_t\) and
\(\widehat V_t\) in \eqref{eq:ppiempvar}, satisfies
\(\lim_{n\to\infty}\mathbb P(\theta_t\in\widehat{\mathcal R}_t)=1-\delta\).
\end{corollary}
\subsection{Confidence Regions for the Performative Stable Point}\label{subsec:infPPI}
The population trajectory is the RRM trajectory, so the enlargement in
Section~\ref{subsec:infRRM} applies similarly.  

\begin{corollary}[Confidence region for \(\theta_{\mathrm{PS}}\)]\label{col:ppips}
Suppose the conditions of Corollary~\ref{col:ppiclt} hold,
\(\varepsilon\beta<\gamma\), and
\(\|\theta_0-\theta_{\mathrm{PS}}\|_2\leq B\). Then
\[
\liminf_{n\to\infty}\mathbb P\Bigl\{\theta_{\mathrm{PS}}\in
\widehat{\mathcal R}_t+\mathcal B\Bigl(0,
B\Bigl(\frac{\varepsilon\beta}{\gamma}\Bigr)^t\Bigr)\Bigr\}\geq1-\delta.
\]
\end{corollary}

%% file: Sections/appA_vanilla.tex

\subsection{Details of Section~\ref{sec:RRM}: Theory of Inference under Performativity }\label{subsec:cltrrm}

We provide the omitted details in Section~\ref{sec:RRM}.
\subsubsection{Proof of Proposition~\ref{prop:differentiability}}
\begin{proof}[Proof of Proposition~\ref{prop:differentiability}]
By Assumption~\ref{as:consistent2}(b), the map
\[
(\widetilde\theta,\theta)\longmapsto
\mathbb E_{Z\sim\mathcal D(\widetilde\theta)}[\nabla\ell(Z;\theta)]
\]
is continuously differentiable. Its Jacobian in \(\theta\) is
\(H_{\widetilde\theta}(\theta)\), and
\(H_{\widetilde\theta}(\theta)\succeq\gamma I_d\) by
Assumption~\ref{as:pp}(d). Since \(G(\widetilde\theta)\) is the unique
solution of
\[
\mathbb E_{Z\sim\mathcal D(\widetilde\theta)}
[\nabla\ell(Z;G(\widetilde\theta))]=0,
\]
the implicit function theorem gives
\[
\nabla G(\theta)
=-H_\theta(G(\theta))^{-1}
\nabla_{\widetilde\theta}
\mathbb E_{Z\sim\mathcal D(\widetilde\theta)}
[\nabla\ell(Z;G(\theta))]
\Big|_{\widetilde\theta=\theta}.
\]
It remains to compute the last derivative. For \(\widetilde\theta\) in a
compact convex neighborhood \(K\) of \(\theta\),
\[
\mathbb E_{Z\sim\mathcal D(\widetilde\theta)}[\nabla\ell(Z;G(\theta))]
=\mathbb E_{Z\sim\mathcal D(\theta)}
\left[\nabla\ell(Z;G(\theta))
\frac{p(Z;\widetilde\theta)}{p(Z;\theta)}\right].
\]
The gradient of the ratio in \(\widetilde\theta\) is
\[
\frac{p(Z;\widetilde\theta)}{p(Z;\theta)}
\nabla_\theta\log p(Z;\widetilde\theta).
\]
By Assumption~\ref{as:consistent2}(a) and the mean value theorem, its norm
is at most
\(L_{s,K}\exp\{L_{s,K}\operatorname{diam}(K)\}\), uniformly over
\(\widetilde\theta\in K\). Assumption~\ref{as:pp}(b) bounds
\(\|\nabla\ell(Z;G(\theta))\|_2\) by an integrable function. Dominated
convergence therefore allows differentiation under the expectation. Since
the density ratio equals one at \(\widetilde\theta=\theta\),
\[
\nabla_{\widetilde\theta}
\mathbb E_{Z\sim\mathcal D(\widetilde\theta)}[\nabla\ell(Z;G(\theta))]
\Big|_{\widetilde\theta=\theta}
=\mathbb E_{Z\sim\mathcal D(\theta)}
\bigl[\nabla\ell(Z;G(\theta))
\nabla_\theta\log p(Z;\theta)^\top\bigr].
\]
Substituting this identity into the preceding display proves the result.
\end{proof}

\subsubsection{Consistency of the RERM Trajectory}
Let us denote:
\begin{align*}
    & \mathcal{L}_{\check{\theta}}(\theta):={\operatorname*{\mathbb{E}}}_{\check{\theta}}\ell(z;\theta), \quad \mathcal{L}_{\check{\theta},n}(\theta):=\frac{1}{n}\sum_{i=1}^n\ell(z_i;\theta),
\end{align*}
where the samples $z_i=(x_i,y_i) \sim \mathcal{D}(\check{\theta})$ are i.i.d. drawn from the distribution under $\check{\theta}$.

\begin{lemma}\label{lm:consistency1}
    Under Assumption~\ref{as:pp}, for each $\delta>0$ and compact subsets $K_1, K_2\subset \Theta=\mathbb{R}^d$ , the following convergence holds as $n\to\infty$:
    $$
 \sup_{\check{\theta}\in K_1}P_{\check{\theta}}\left(\sup_{\theta\in K_2}\big|\mathcal{L}_{\check{\theta},n}(\theta)-\mathcal{L}_{\check{\theta}}(\theta)\big|>\delta\right) \to 0,
 $$
 
\end{lemma}
\begin{proof}
    Define $K:=K_1\cup K_2$. For $r>0$, by the compactness of $K_2$, let $\{\theta_1,\ldots,\theta_N\}$ be an $r$-net of $K_2$. 
    Then for each ${\check{\theta}}\in K_1,\theta\in K_{2}$, there exists a $\theta_j$, such that $\|\theta-\theta_j\|_{2}\leq r$, thus recall $\ell(Z;\theta)$ is local $L_K(Z)$-Lipschitz,
    \begin{equation*}
        \big|\mathcal{L}_{\check{\theta},n}(\theta)-\mathcal{L}_{\check{\theta},n}(\theta_j)\big|\leq \widehat{\mathbb{E}}_{\check{\theta},n}L_K(Z)\|\theta_j-\theta\|_{2},\quad \big|\mathcal{L}_{\check{\theta}}(\theta)-\mathcal{L}_{\check{\theta}}(\theta_j)\big|\leq {\mathbb{E}}_{\check{\theta}}L_K(Z)\|\theta_j-\theta\|_{2}, 
    \end{equation*}
    we have 
    \begin{equation}\label{eq:consistency2}
        \sup_{\theta\in K_{2}}\big|\mathcal{L}_{\check{\theta},n}(\theta)-\mathcal{L}_{\check{\theta}}(\theta)\big|
        \leq \sup_{1\leq j\leq N}\big|\mathcal{L}_{\check{\theta},n}(\theta_j)-\mathcal{L}_{\check{\theta}}(\theta_j)\big| + r\big(\widehat{\mathbb{E}}_{\check{\theta},n}L_K(Z) +  \mathbb{E}_{\check{\theta}}L_K(Z)\big).
    \end{equation}
    For the first term on the right hand side of the above formula, note that $K=K_{1}\cup K_{2}$, 
    by Chebyshev's inequality and a union bound,
    \begin{equation*}
        P_{{\check{\theta}}}(\max_{1\leq j\leq N}|\mathcal{L}_{{\check{\theta}},n}(\theta_j) - \mathcal{L}_{{\check{\theta}}}(\theta_j)|>\delta/2)\leq \frac{4N}{n\delta^2}\max_{1\leq j\leq N}\mathbb{E}_{{\check{\theta}}}|\ell(Z;\theta_j)|^2\leq \frac{4N}{n\delta^2}\sup_{{\check{\theta}}\in K}\mathbb{E}_{{\check{\theta}}}|A_{K}(Z)|^2,
    \end{equation*}
    when $n\to\infty$, 
    \begin{equation*}
        \sup_{{\check{\theta}}\in K_1} P_{{\check{\theta}}}(\max_{1\leq j\leq N}|\mathcal{L}_{\check{\theta},n}(\theta_j) - \mathcal{L}_{{\check{\theta}}}(\theta_j)|>\delta)\to 0.
    \end{equation*}
    For the second term on the right hand side of formula Eq.~\eqref{eq:consistency2}, for fixed $r>0$,
    \begin{equation*}
        \sup_{{\check{\theta}}\in K_1}P_{{\check{\theta}}}(|\widehat{\mathbb{E}}_{{\check{\theta}},n}L_K(Z)-\mathbb{E}_{{\check{\theta}}}L_K(Z)|>{\delta})\leq \frac{1}{n\delta^2}\sup_{{\check{\theta}}\in K_1}\mathbb{E}_{{\check{\theta}}}|L_K(Z)|^{2}.
    \end{equation*}

    Then for each fixed $\delta$, select $r>0$ sufficiently small, so that $r\sup_{\check{\theta}\in K}\mathbb{E}_{{\check{\theta}}}|L_K(Z)|<\frac{\delta}{4}$, then
    \begin{equation*}
        \begin{aligned}
        &\sup_{{\check{\theta}}\in K_1}P_{{\check{\theta}}}(r\big(\widehat{\mathbb{E}}_{{\check{\theta}},n}L_K(Z)+\mathbb{E}_{{\check{\theta}}}L_K(Z)\big)>{\delta})
        \\ \leq & \sup_{{\check{\theta}}\in K_1}P_{{\check{\theta}}}(r\big(\widehat{\mathbb{E}}_{{\check{\theta}},n}L_K(Z)-\mathbb{E}_{{\check{\theta}}}L_K(Z)\big)>{\delta}/2)
        \\ \leq & \frac{4r^2N}{n\delta^2}\sup_{{\check{\theta}}\in K_1}\mathbb{E}_{{\check{\theta}}}|L_K(Z)|^{2}.
        \end{aligned}
    \end{equation*}

    By the above inequalities for the terms on the right hand side of formula Eq.~\eqref{eq:consistency2},
    we have 
    \begin{equation*}
 \sup_{{\check{\theta}}\in K_1}P_{\check{\theta}}\left(\sup_{\theta\in K_2}\big|\widehat{\mathbb{E}}_{{\check{\theta}},n}\ell(Z;\theta)-\mathbb{E}_{\check{\theta}}\ell(Z;\theta)\big|>\delta\right) \to 0.
    \end{equation*}

\end{proof}

\begin{lemma}\label{lm:consistency1_1}
    Under Assumption~\ref{as:pp}, for a compact subset $K\subset \Theta$, then for each $n>0$, denote $\widehat{G}({\check{\theta}})=\argmin_{\theta\in \Theta}\mathcal{L}_{{\check{\theta}},n}(\theta)$, when $R\to \infty$, we have
    \begin{equation*}
        \lim_{R\to\infty}\sup_{n\geq 1}\sup_{{\check{\theta}}\in K}P_{{\check{\theta}}}(\|\widehat{G}({\check{\theta}})-G({\check{\theta}})\|_{2}>R)= 0.
    \end{equation*}
\end{lemma}
\begin{proof}[Proof of Lemma~\ref{lm:consistency1_1}]
    By the optimality condition, \begin{equation*}
        \big\langle\nabla_{\theta}{\mathcal{L}}_{{\check{\theta}},n}(\widehat{G}({\check{\theta}})),G({\check{\theta}})-\widehat{G}({\check{\theta}})\big\rangle\leq 0.
    \end{equation*}
    By strong convexity, we have 
    \begin{equation*}
        \big\langle\nabla_{\theta}{\mathcal{L}}_{{\check{\theta}},n}(\widehat{G}({\check{\theta}}))- \nabla_{\theta}{\mathcal{L}}_{{\check{\theta}},n}({G}({\check{\theta}})),\widehat{G}({\check{\theta}})-{G}({\check{\theta}})\big\rangle\geq \gamma \|G({\check{\theta}})-\widehat{G}({\check{\theta}})\|_{2}^2,
    \end{equation*}
    This implies 
    \begin{equation*}
        \big\langle \nabla_{\theta}{\mathcal{L}}_{{\check{\theta}},n}({G}({\check{\theta}})),G({\check{\theta}})-\widehat{G}({\check{\theta}})\big\rangle\geq \gamma \|G({\check{\theta}})-\widehat{G}({\check{\theta}})\|_{2}^2,
    \end{equation*}
    Thus, 
    \begin{equation*}
        \|\nabla_{\theta}\mathcal{L}_{{\check{\theta}},n}(G({\check{\theta}}))\|_{2}\geq \gamma\|G({\check{\theta}})-\widehat{G}({\check{\theta}})\|.
    \end{equation*}

    Therefore, for compact set $K$, there exists $R_1>0$, such that $\|G({\check{\theta}})\|\leq R_{1}$, and denote $K_{1}:=B(0,R_{1})\cup K$. By Assumption~\ref{as:pp}, for each $R>0$,
    \begin{equation*}
        P_{{\check{\theta}}}\bigg(\big\|\nabla_{\theta}\mathcal{L}_{{\check{\theta}},n}(G({\check{\theta}}))\big\|_{2}>R\bigg)\leq \frac{1}{R^2}\mathbb{E}_{{\check{\theta}}}\big[\|\nabla_{\theta}\mathcal{L}_{\check{\theta},n}(G({\check{\theta}}))\|_{2}^2\big]\leq \frac{1}{R^2}\sup_{{\check{\theta}}\in K_{1}}\mathbb{E}_{{\check{\theta}}}\big[L_{K_1}(Z)^2\big],
    \end{equation*}
    let $R\to \infty$, we know that 
    $$\sup_{n\ge 1}\sup_{{\check{\theta}}\in K}P_{{\check{\theta}}}(\|\widehat{G}({\check{\theta}})-G({\check{\theta}})\|_{2}>R)\to 0$$
    
\end{proof}

\begin{proposition}\label{prop:consistency}
Under Assumption~\ref{as:pp}, \(\widehat\theta_t\xrightarrow{P}\theta_t\)
for each fixed \(t\geq1\).
\end{proposition}

\begin{proof}
We use \(\widehat G\) defined in Lemma~\ref{lm:consistency1_1} and prove the result by induction on $t$.
By consistency of $M$-estimator (Theorem 5.9 in \citep{van2000asymptotic}), we have $\widehat{\theta}_{1}\xrightarrow{P} \theta_1$. We assume that as $n\to \infty$, $\widehat{\theta}_{t-1}\xrightarrow{P} \theta_{t-1}$, and set off to prove that $\widehat{\theta}_{t}\xrightarrow{P} \theta_t$.
   \begin{align*}
    \|\theta_t-\widehat{\theta}_t\|&=\|G(\theta_{t-1})-\widehat{G}(\widehat{\theta}_{t-1})\|\\
    &\leq \|G(\widehat{\theta}_{t-1})-\widehat{G}(\widehat{\theta}_{t-1})\| + \|G(\theta_{t-1})-G(\widehat{\theta}_{t-1})\| \\
    &\leq  \|G(\widehat{\theta}_{t-1})-\widehat{G}(\widehat{\theta}_{t-1})\| + \varepsilon\frac\beta\gamma \|\theta_{t-1}-\widehat{\theta}_{t-1}\|,
\end{align*}  
where the last inequality follows from the results derived by \cite{perdomo2020performative}, under Assumption~\ref{as:pp}, we have $\|G(\theta)-G(\theta')\|\le  \frac{\varepsilon \beta}{\gamma}\|\theta-\theta'\|$.

For any $\delta>0$, by inductive assumption, $\widehat{\theta}_{t-1}\xrightarrow{P} \theta_{t-1}$, thus there exists compact subset $K_{1}\subset \Theta$, such that 
\begin{equation*}
    P(\widehat{\theta}_{t-1}\notin K_{1})< \delta/2.
\end{equation*}

By Lemma~\ref{lm:consistency1_1}, there exists compact $K_{2}\subset \Theta$,  such that $G(K_{1})\subset K_{2}$ and that  $P(\widehat{\theta}_{t-1}\in K_{1}, \widehat{G}(\widehat{\theta}_{t-1})\notin K_{2})<\delta$, thus in the following, we only need to prove the situation where $\widehat{\theta}_{t-1}\in K_{1}$ and $\widehat{G}(\widehat{\theta}_{t-1})\in K_{2}$.

Note that $\mathcal{L}_{{\check{\theta}},n}(\widehat{G}({\check{\theta}})) - \mathcal{L}_{{\check{\theta}},n}(G({\check{\theta}})) \leq 0$, we have
\begin{equation*}
    \begin{aligned}
     &\mathcal{L}_{{\check{\theta}}}(\widehat{G}({\check{\theta}})) -\mathcal{L}_{{\check{\theta}}}({G}({\check{\theta}}))
    \\\leq & \big|\mathcal{L}_{{\check{\theta}}}(\widehat{G}({\check{\theta}}))- \mathcal{L}_{{\check{\theta}},n}(\widehat{G}({\check{\theta}}))\big| +  \mathcal{L}_{{\check{\theta}},n}(\widehat{G}({\check{\theta}})) - \mathcal{L}_{{\check{\theta}},n}(G({\check{\theta}})) + \big|\mathcal{L}_{{\check{\theta}},n}({G}({\check{\theta}})) -\mathcal{L}_{{\check{\theta}}}({G}({\check{\theta}}))\big|
    \\ \leq & 2\sup_{\theta\in K_2}\big|\mathcal{L}_{{\check{\theta}},n}(\theta)-\mathcal{L}_{{\check{\theta}}}(\theta)\big|,
    \end{aligned}
\end{equation*}
and $\mathcal{L}_{{\check{\theta}}}(\widehat{G}({\check{\theta}})) -\mathcal{L}_{{\check{\theta}}}({G}({\check{\theta}}))\geq \frac{\gamma}{2}\|\widehat{G}({\check{\theta}})-G({\check{\theta}})\|^{2}_{2}$,
combine these two inequalities, we obtain that  for each ${\check{\theta}}\in K_1$, 
\begin{equation*}
    \|\widehat{G}({\check{\theta}})-G({\check{\theta}})\|_{2}^2\leq \frac{4}{\gamma}\sup_{\theta\in K_{2}}\big|\mathcal{L}_{{\check{\theta}},n}(\theta)-\mathcal{L}_{{\check{\theta}}}(\theta)\big|.
\end{equation*}
We then have for each $\delta>0$,
 $$P_{{\check{\theta}}}(\|\widehat{G}({\check{\theta}})-G({\check{\theta}})\|>\delta)< P_{\check{\theta}}\left(\sup_{\theta\in K_{2}}\big|\mathcal{L}_{{\check{\theta}},n}(\theta)-\mathcal{L}_{{\check{\theta}}}(\theta)\big|>\frac{\gamma\delta^2}{4}\right),$$
by Lemma~\ref{lm:consistency1}, the following uniform convergence formula holds,
 $$
P\bigg(\sup_{\theta\in K_{2}}\big|\mathcal{L}_{{\check{\theta}},n}(\theta)-\mathcal{L}_{{\check{\theta}}}(\theta)\big|>\frac{\gamma\delta^2}{4}\bigg|\widehat{\theta}_{t-1}\bigg)\leq 
\sup_{{\check{\theta}}\in K_{1}}P_{\check{\theta}}\left(\sup_{\theta\in K_{2}}\big|\mathcal{L}_{{\check{\theta}},n}(\theta)-\mathcal{L}_{{\check{\theta}}}(\theta)\big|>\frac{\gamma\delta^2}{4}\right) \to 0,
 $$
it implies $\|\widehat{G}(\widehat{\theta}_{t-1})-{G}(\widehat{\theta}_{t-1})\|_{2}\xrightarrow{P} 0$.
\end{proof}
\subsubsection{Preparation Lemmas}
In this subsection, we provide some technical lemmas that will be used in our proof of Theorem~\ref{thm:cltrrm} and Theorem~\ref{thm:cltrgd}.
\begin{lemma}\label{lm:local}
    For a function sequence $R_{n}:\mathbb{R}^d\to \mathbb{C}$, assume:
    \begin{enumerate}
        \item There exists a constant $C>0$, such that, $\sup_{n}\sup_{x\in \mathbb{R}^d}|R_{n}(x)|<C$;
        \item On any compact subset $K\subset \mathbb{R}^d$, $\sup_{\xi\in K}|R_{n}(\xi)|\to 0$.
    \end{enumerate} 
    Then for any random vector sequence $\{\xi_{n}\}$ taking values in $\mathbb{R}^d$ and $\xi_{n}=O_{P}(1)$,
    \begin{equation*}
        \mathbb{E}|R_{n}(\xi_{n})|\to 0.
    \end{equation*}
\end{lemma}
\begin{proof}[Proof of Lemma~\ref{lm:local}]
    For any $\varepsilon>0$, take a compact subset $K\subset\bR^{d}$ such that  $P(\xi_{n}\notin K)\le \varepsilon$ holds for each $n\in \bN$, and take $N$ sufficiently large, such that $\sup_{\xi\in K}|R_{n}(\xi)|\le \varepsilon$ holds for $n\ge N$. Then we have  $$\bE|R_n(\xi_n)| \le \sup_{\xi\in K} |R_n(\xi)| + C P(\xi_n \notin K)\le (1+C)\varepsilon ,\quad n\ge N.$$ 
\end{proof}

\begin{lemma}\label{lm:rgd1}
    Assume that $\check{\theta}_{0}\in \Theta$ is a fixed parameter in $\Theta$, and $\{\check{\theta}_{n}\}_{n=1}^{\infty}$ is a sequence of random variables such that $\sqrt{n}(\check{\theta}_{n}-\check{\theta}_{0})\to \mathcal{N}(0,A_{1})$ for a positive semidefinite matrix $A_{1}\in \mathbb{R}^{d\times d}$. Conditional on $\check\theta_n$, let $\{Z_{1,n}^{(j)}\}_{j=1}^{n}$ be i.i.d. from $\mathcal{D}(\check{\theta}_{n})$. For a  function $f_{\theta}:\mathcal{Z}\to \mathbb{R}^q$, where $q$ is a positive integer, define $F({\theta})=\mathbb{E}_{\theta}[f_{\theta}(Z)]$, $V_f(\theta)=\mathrm{Cov}_{\theta}[f_\theta(Z)]$. 
    If for any compact subset $K\subset \Theta$, $F(\theta)$ is bounded and continuously differentiable on an open neighborhood of $K$, and  $V_f(\theta)$ is continuous on $K$, and there exists constant $C_K$, 
    such that $\sup_{\theta\in K}\mathbb{E}_{\theta}[\|f_\theta(Z)\|^{3}_{2}]<C_{K}$,
    then we have
    \begin{equation*}
        \sqrt{n}\big(\frac{1}{n}\sum_{j=1}^{n}f_{\check{\theta}_{n}}(Z_{1,n}^{(j)})-F(\check{\theta}_{0})\big) \xrightarrow{d} \mathcal{N}\bigg(0,V_f(\check{\theta}_{0}) + \nabla_{\theta}F(\check{\theta}_{0})A_{1}\big(\nabla_{\theta}F(\check{\theta}_{0})\big)^{\top}\bigg).
    \end{equation*}
\end{lemma}
\begin{proof}[Proof of Lemma~\ref{lm:rgd1}]
    Denote   
    $
        h_{2,n} = \sqrt{n}\big(\frac{1}{n}\sum_{j=1}^{n}f_{\check{\theta}_{n}}(Z_{1,n}^{(j)})-F(\check{\theta}_{0})\big),
    $
    the characteristic function of $h_{2,n}$ is 
    \begin{equation*}
        \begin{aligned}
            \phi_{h_{2,n}}(s)&=\mathbb{E}\bigg[\mathbb{E}_{\check{\theta}_{n}}\big[\exp(is^{\top}h_{2,n})\big|\check{\theta}_{n}\big]\bigg]
            \\ &=\mathbb{E}\bigg[\left\{\mathbb{E}_{\check{\theta}_{n}}\left[\exp\big(\frac{is^\top}{\sqrt{n}} (f_{\check{\theta}_{n}}(Z)-F(\check{\theta}_{0}))\big)\right]\right\}^{n}\bigg].
        \end{aligned}
    \end{equation*}
    Fix a compact subset $K\subset\Theta$. For fixed $s$, when $\sqrt{n}\to \infty$,  by the assumption, for any $\theta\in K$, we have 
    \begin{equation*}
        \bigg|\mathbb{E}_{\theta}[\exp(\frac{1}{\sqrt{n}}{is}^\top f_{{\theta}}(Z))] - 1- \frac{1}{\sqrt{n}}is^\top F({\theta}) + \frac{1}{2n}s^\top \mathbb{E}_{\theta}[f_{\theta}(Z)f_{\theta}^\top(Z)]s\bigg| \leq C\|\frac{s}{\sqrt{n}}\|_{2}^{3}.
    \end{equation*}

    Thus by the boundedness of $F(\theta)$ on $K$, the following expansion holds uniformly for $\theta,\theta^\prime\in K$:
    \begin{equation*}
        \begin{aligned}
            &\mathbb{E}_{\theta}\big[\exp\big(\frac{is^\top}{\sqrt{n}} (f_{\theta}(Z)-F(\theta^\prime))\big)\big]
             \\ =& 1+ \frac{is^\top}{\sqrt{n}}  \big(F(\theta)-F(\theta^\prime)\big) -\frac{1}{2n} s^\top \big[V_f(\theta)+\big(F(\theta)-F(\theta^\prime)\big)\big(F(\theta)- F(\theta^\prime)\big)^\top\big]s + O(\|\frac{s}{\sqrt{n}}\|_{2}^3)
        \end{aligned}
    \end{equation*} 

    For $s\in \mathbb{R}^q,\theta\in\Theta$, define $\chi(s,\theta):=\exp(-\frac{1}{2}s^\top V_{f}(\theta)s)$ and
    \begin{equation*}
        \chi_{n}(s,\theta):=\bigg(\mathbb{E}_{\theta}\big[\exp\big(\frac{is^\top}{\sqrt{n}} (f_{\theta}(Z)-F(\theta))\big)\big]\bigg)^{n},
    \end{equation*}
    Then for any compact set $K\subset\Theta$ and fixed $s$, $\sup_{\theta\in K}|\chi_{n}(s,\theta)-\chi(s,\theta)|\to 0$ as $n\to\infty$.

    Note that $\chi,\chi_{n}$ are bounded, by Lemma~\ref{lm:local}, take $\xi_{n}=\check{\theta}_{n}$ and $R_{n}(\theta)=\chi_{n}(s,\theta)-\chi(s,\theta)$, we have for each fixed $s$,
    \begin{equation*}
        \chi_{n}(s,\check{\theta}_{n}) - \chi(s,\check{\theta}_{n}) \xrightarrow{L_{1}} 0.
    \end{equation*}

    Further, we have 
    \begin{equation*}
        \phi_{h_{2,n}}(s) = \mathbb{E}\bigg[\exp\big(is^\top\sqrt{n}\big[F(\check{\theta}_{n})-F(\check{\theta}_{0})\big]\big)\chi_{n}(s,\check{\theta}_{n})\bigg]
    \end{equation*}
    Applying Delta method to the first term, we have 
    \begin{equation*}
        \sqrt{n}s^\top \big(F(\check{\theta}_{n})-F(\check{\theta}_{0})\big)\xrightarrow{d} 
        \mathcal{N}\bigg(0,s^\top \nabla_\theta F(\check{\theta}_0)A_{1}\big(\nabla_{\theta}F(\check{\theta}_0)\big)^\top  s\bigg).
    \end{equation*}
    For the second term,  we note that $\chi(s,\check{\theta}_{n})\xrightarrow{P}\chi(s,\check{\theta}_{0})$ by continuous mapping theorem. 
    Then by Portmanteau Lemma, we have the following convergence: 
    
    \begin{equation*}
        \begin{aligned}
            \phi_{h_{2,n}}(s)\to \exp\Bigg(-\frac{1}{2}s^{\top}\bigg(V_{f}(\check{\theta}_{0}) + \nabla_{\theta}F(\check{\theta}_{0})A_{1}\big(\nabla_{\theta}F(\check{\theta}_{0})\big)^{\top}\bigg)s\Bigg).
        \end{aligned}
    \end{equation*}

    Note that the right hand side is the characteristic function of a multi-variate normal distribution, by Levy's continuity theorem, the above pointwise convergence of characteristic function implies:
    \begin{equation*}
        \sqrt{n}\big(\frac{1}{n}\sum_{j=1}^{n}f_{\check{\theta}_{n}}(Z_{1,n}^{(j)})-F(\check{\theta}_{0})\big) \xrightarrow{d} \mathcal{N}\bigg(0,V_{f}(\check{\theta}_{0}) + \nabla_{\theta}F(\check{\theta}_{0})A_{1}\big(\nabla_{\theta}F(\check{\theta}_{0})\big)^{\top}\bigg).
    \end{equation*}

\end{proof}

\begin{lemma}\label{rm:consistent}
Under Assumption~\ref{as:consistent2}, for each compact subset $K\subset \Theta$, denote $K_{1}$ its compact convex hull, define $L_{s,K}^\prime:=L_{s,K_{1}}e^{DL_{s,K_{1}}}$ where $D=\text{diam}(K)$ and $L_{s,K_{1}}$ is the upper bound of the  norm of score function $\nabla_{\theta}\log p(z;\theta)$ in Assumption~\ref{as:consistent2}, then for any $\check{\theta},\theta\in K$ and $z\in \mathcal{Z}$, we have 
\begin{equation*}
    \|\dfrac{p(z;\theta)}{p(z;\check{\theta})}-1\|\leq L_{s,K}^\prime\|\theta-\check{\theta}\|_{2}.
\end{equation*}
\end{lemma}
\begin{proof}
    For any $\check{\theta},\theta\in K$ and $z\in \mathcal{Z}$, by mean-value theorem we have 
\begin{equation*}
    |\log p(z;\check{\theta})- \log p(z;\theta)|\leq L_{s,K_{1}}\|\check{\theta}-\theta\|_{2},
\end{equation*}
this leads to 
$$e^{-L_{s,K_{1}}\|\theta-\check{\theta}\|_{2}}\leq p(z;\theta)/p(z;\check{\theta})\leq e^{L_{s,K_{1}}\|\theta-\check{\theta}\|_{2}},$$
then, we have
\begin{equation*}
    \|\dfrac{p(z;\theta)}{p(z;\check{\theta})}-1\|\leq e^{L_{s,K_{1}}\|\theta-\check{\theta}\|_{2}}-1.
\end{equation*}
Since $K$ is bounded, the mean-value theorem gives
\begin{equation*}
    e^{L_{s,K_{1}}\|\theta-\check{\theta}\|_{2}}-1 \leq e^{DL_{s,K_{1}}}L_{s,K_{1}}\|\theta-\check{\theta}\|_{2}.
\end{equation*}
Therefore,
$
    \|\dfrac{p(z;\theta)}{p(z;\check{\theta})}-1\|\leq L_{s,K}^\prime\|\theta-\check{\theta}\|_{2}.
 $
\end{proof}
\subsubsection{Proof of CLT for RRM (Theorem~\ref{thm:cltrrm})}

We first prove the following local expansion lemma with dynamic underlying distribution.
{
\begin{lemma}\label{lm:localexpansion1}
    For a function \(f:\mathcal Z\to\mathbb R\) and
    \(\check\theta\in\Theta\), let
    \[
    \mathbb G_{n,\check\theta}f
    :=\sqrt n\left\{\frac1n\sum_{j=1}^n f(Z_j)
    -\mathbb E_{\check\theta}[f(Z)]\right\},
    \]
    where \(Z_1,\ldots,Z_n\) are i.i.d. from
    \(\mathcal D(\check\theta)\). Under Assumption~\ref{as:pp}, given two random vector sequences $\{\check{\theta}_{n}\}_{n=1}^{\infty},\{\theta_{n}\}_{n=1}^{\infty}$ with $\check{\theta}_{n}\xrightarrow{P} \check{\theta}_{0}$ and $\theta_{n}\xrightarrow{P} \theta_{0}$, if
    \(h_n=O_p^*(1)\), then we have
    \begin{equation*}
        \mathbb G_{n,\check{\theta}_{n}}\left[
        \sqrt n\left(\ell_{\theta_{n}+h_n/\sqrt n}-\ell_{\theta_{n}}\right)
        -h_n^\top\nabla\ell_{\theta_{n}}\right]=o_{P}^{*}(1)
    \end{equation*}
    with
    $\bG_{n,\check{\theta}_{n}}f=\sqrt n\left\{\frac1n\sum_{j=1}^n f(Z_{j,n})
    -\mathbb E_{\check{\theta}_{n}}[f(Z)]\right\}$, where, conditional on \((\check\theta_n,\theta_n)\), the observations \(Z_{1,n},\ldots,Z_{n,n}\) are i.i.d. from \(\mathcal D(\check\theta_n)\).
\end{lemma}
\begin{proof}[Proof of Lemma~\ref{lm:localexpansion1}] We follow the proof of Lemma 19.31 in \cite{van2000asymptotic}. For $M>0$, we define $B_M:=\big\{h\in \mathbb{R}^d: \|h\|_{2}\leq M\big\}$. 

    \textbf{Step 1.} Uniform bound on $B_{M}$. Fix a nonrandom $h\in B_M$.
    For fixed \(\theta,\check{\theta}\in\Theta\) and a constant $h\in \mathbb{R}^d$, denote $$r_{n,\theta,h}(z)=\sqrt{n}\big(\ell_{\theta+h/\sqrt{n}}(z)-\ell_{\theta}(z)\big)-h^{\top}\nabla_{\theta}\ell_{\theta}(z),$$
    then we have $\mathbb{E}_{\check\theta}[\mathbb{G}_{n,\check{\theta}}r_{n,\theta,h}] = 0$. 
    Note that by mean-value theorem, for each $z$, there exists a constant $c_z\in [0,1]$ and  $r_{n,\theta,h}(z) = h^{\top}\big(\nabla_{\theta}\ell_{\theta+\frac{c_zh}{\sqrt{n}}}(z)-\nabla_{\theta}\ell_{\theta}(z)\big)$, under Assumption~\ref{as:pp}, the joint smoothness of $\ell$ implies that $|r_{n,\theta,h}(z)|\leq \frac{\beta}{\sqrt{n}}\|h\|_2^{2}$, thus
    \begin{equation*}
        \text{Var}_{\check{\theta}}(\mathbb{G}_{n,\check{\theta}}r_{n,\theta,h}) = \text{Var}_{\check{\theta}}(r_{n,\theta,h}(Z))\leq \mathbb{E}_{\check{\theta}}\|r_{n,\theta,h}(Z)\|^{2} \leq \frac{\beta^{2}}{n}\|h\|_{2}^{4}.
    \end{equation*}
    
    \textbf{Step 2.} Equicontinuity on $B_{M}$. For fixed \(\theta,\check{\theta}\in\Theta\), $z\in \mathcal{Z}$ and $h,h^\prime\in B_M$,
    \begin{equation*}
        r_{n,\theta,h}(z) = \int_{0}^{1}h^\top\left(\nabla_{\theta}\ell_{\theta+sh/\sqrt{n}}(z)-\nabla_\theta\ell_{\theta}(z)\right)ds
    \end{equation*}
    by the joint smoothness assumption (Assumption~\ref{as:pp}(c)), we have 
    \begin{equation*}
        \begin{aligned}
        &\quad |r_{n,\theta,h}(z)-r_{n,\theta,h^\prime}(z)| \\ &= \bigg|\int_{0}^{1}(h-h^\prime)^\top \big[\nabla_\theta\ell_{\theta+sh/\sqrt{n}}(z)-\nabla_\theta\ell_{\theta}(z)\big]ds + \int_{0}^1(h^\prime)^\top\big[\nabla_\theta\ell_{\theta+sh/\sqrt{n}}(z)-\nabla_\theta\ell_{\theta+sh^\prime/\sqrt{n}}(z)\big]ds\bigg|
        \\ &\leq \bigg|\int_{0}^{1}(h-h^\prime)^\top \big[\nabla_\theta\ell_{\theta+sh/\sqrt{n}}(z)-\nabla_\theta\ell_{\theta}(z)\big]ds\bigg| + \bigg|\int_{0}^1(h^\prime)^\top\big[\nabla_\theta\ell_{\theta+sh/\sqrt{n}}(z)-\nabla_\theta\ell_{\theta+sh^\prime/\sqrt{n}}(z)\big]ds\bigg|
        \\ &\leq \|h-h^\prime\|_{2}\cdot \int_{0}^{1}\|\nabla_{\theta}\ell_{\theta+sh/\sqrt{n}}(z) - \nabla_{\theta}\ell_{\theta}(z)\|_{2}ds 
        + \|h^\prime\|_2\int_{0}^{1}\|\nabla_{\theta}\ell_{\theta+sh/\sqrt{n}}(z)-\nabla_{\theta}\ell_{\theta+sh^\prime/\sqrt{n}}(z)\|_{2}ds
        \\ &\leq \|h-h^\prime\|_2\frac{\|h\|_{2}}{\sqrt{n}}\beta + \|h^\prime\|_2\frac{\beta\|h-h^\prime\|_{2}}{\sqrt{n}}
        \\ &\leq \frac{2M\beta}{\sqrt{n}}\|h-h^\prime\|_{2}.
        \end{aligned}
    \end{equation*}
    Thus, the following inequality holds for any $\check{\theta}\in \Theta$
    \begin{equation*}
        \big|\mathbb{G}_{n,\check{\theta}}\big(r_{n,\theta,h}-r_{n,\theta,h^\prime}\big)\big|\leq \sqrt{n}\widehat{\mathbb{E}}_{\check{\theta},n}|r_{n,\theta,h}-r_{n,\theta,h^\prime}| + \sqrt{n}\mathbb{E}_{\check{\theta}}|r_{n,\theta,h}-r_{n,\theta,h^\prime}|\leq 4M\beta\|h-h^\prime\|_{2}
    \end{equation*}
    Fix $\varepsilon>0$, take $\delta=\frac{\varepsilon}{8M\beta}>0$, and a finite $\delta$-net $\{h_{1},\ldots,h_{N}\}$ of $B_M$. Then for each $h\in B_M$, there exists $j\in [1:N]$, such that $\|h-h_j\|_2\leq \delta$, then for each $\check{\theta}\in \Theta$,
    \begin{equation*}
        \big|\mathbb{G}_{n,\check{\theta}}(r_{n,\theta,h}-r_{n,\theta,h_j})\big|\leq 4M\beta\|h-h_j\|_2\leq 4M\beta\delta\leq\frac{\varepsilon}{2}.
    \end{equation*}
    This shows 
    \begin{equation*}
        \sup_{\theta,\check{\theta}\in \Theta}P^*\big(\sup_{h\in B_M}|\mathbb{G}_{n,\check{\theta}}r_{n,\theta,h}|>\varepsilon\big)\leq \sup_{\theta,\check{\theta}\in \Theta}P\big(\max_{1\leq j\leq N}|\mathbb{G}_{n,\check{\theta}}r_{n,\theta,h_{j}}|>\frac{\varepsilon}{2}\big),
    \end{equation*}
    Using the variance bound in Step 1, we have 
    \begin{equation*}
        \sup_{\theta,\check{\theta}\in \Theta}P\big(\max_{1\leq j\leq N}|\mathbb{G}_{n,\check{\theta}}r_{n,\theta,h_j}|>\frac{\varepsilon}{2}\big)\leq \dfrac{4N\beta^2M^4}{n\varepsilon^{2}}.
    \end{equation*}

    Finally, given random sequence $\check{\theta}_{n}$ and $\theta_{n}$, and general $h_{n}=O_{p}^*(1)$, we have

    $$P\left(
\sup_{\|h\|\le M}
|\mathbb G_{n,\check\theta_n}r_{n,\theta_n,h}|>\varepsilon \,\middle|\,\check\theta_n,\theta_{n} \right) \le\dfrac{4N\beta^2M^4}{n\varepsilon^{2}},$$
taking expectation and by boundedness of $h_{n}$, for any fixed $a>0$, we have  $\limsup_{n}P^*(\|h_{n}\|_{2}>M)\le a$, thus
    \begin{equation*}
        \limsup_{n} P^{*}(|\mathbb{G}_{n,\check{\theta}_{n}}r_{n,\theta_n, h_n}|>\varepsilon)\leq \limsup_{n}P^{*}(\|h_n\|_2>M) + \limsup_{n}P^*(\sup_{h\in B_M}|\mathbb{G}_{n,\check{\theta}_{n}}r_{n,\theta_{n},h}|>\varepsilon)\leq a,
    \end{equation*}
    As \(a\downarrow0\), this proves the desired convergence for the fixed threshold $\varepsilon$.
\end{proof}
}

\begin{lemma}\label{lm:localexpansion2}
    Under Assumption~\ref{as:pp} and \ref{as:consistent2}, for fixed $\check{\theta}_{0}\in \Theta$ and a sequence of random variables  
    assume $\sqrt{n}(\check{\theta}_{n}-\check{\theta}_{0})=O_{p}(1)$ and $h_{n}=O_{p}^{\ast}(1)$ is a random sequence, we have
    \begin{equation*}
        {n}\big(\mathbb{E}_{\check{\theta}_{n}}-\mathbb{E}_{{\check{\theta}_{0}}}\big)\bigg[\ell_{G(\check{\theta}_{0})+\frac{h_{n}}{\sqrt{n}}}(Z)-\ell_{G(\check{\theta}_{0})}(Z)-\frac{h_{n}^{\top}}{\sqrt{n}}\nabla\ell_{G(\check{\theta}_{0})}(Z)\bigg] = o_{p}^*(1)
    \end{equation*}
\end{lemma}
\begin{proof}
By Lemma~\ref{rm:consistent} and Assumption~\ref{as:consistent2}, we have for any compact set $K\subset \Theta$ containing $\check{\theta}_0$, when $\check{\theta}\in K$, for any fixed $h\in \mathbb{R}^{d}$, 
        \begin{equation*}
        \begin{aligned}
            &{n}\bigg|\big(\mathbb{E}_{\check{\theta}}-\mathbb{E}_{{\check{\theta}_{0}}}\big)\big[\ell_{G(\check{\theta}_{0})+\frac{h}{\sqrt{n}}}(Z)-\ell_{G(\check{\theta}_{0})}(Z)-\frac{h^{\top}}{\sqrt{n}}\nabla\ell_{G(\check{\theta}_{0})}(Z)\big]\bigg| 
            \\ \leq &n\mathbb{E}_{\check{\theta}_{0}}\bigg[\bigg|\big(\frac{p(Z;\check{\theta})}{p(Z;\check{\theta}_{0})}-1 \big)\big(\ell_{G(\check{\theta}_{0})+\frac{h}{\sqrt{n}}}(Z)-\ell_{G(\check{\theta}_{0})}(Z)-\frac{h^{\top}}{\sqrt{n}}\nabla_{\theta}\ell_{G(\check{\theta}_{0})}(Z)\big)\bigg|\bigg]
            \\ \leq & \sqrt{n}\mathbb{E}_{\check{\theta}_{0}}\bigg[L_{s,K}^\prime\big\|\check{\theta}-\check{\theta}_{0}\big\|_{2}\times\bigg|\sqrt{n}\big(\ell_{G(\check{\theta}_{0})+\frac{h}{\sqrt{n}}}(Z)-\ell_{G(\check{\theta}_{0})}(Z)\big)-{h^{\top}}\nabla_{\theta}\ell_{G(\check{\theta}_{0})}(Z)\bigg|\bigg]
            \\ \leq & \sqrt{n}\mathbb{E}_{\check{\theta}_{0}}\bigg[L_{s,K}^\prime\big\|\check{\theta}-\check{\theta}_0\big\|_{2}\cdot  \frac{\beta}{\sqrt{n}}\|{h}\|_{2}^{2}\bigg]
            \\ \leq & \beta L_{s,K}^\prime\|h\|_{2}^{2}\cdot \|\check{\theta}-\check{\theta}_{0}\|_{2},
        \end{aligned}
    \end{equation*}

    Define $R_{n}(\check{\theta},h)={n}\big(\mathbb{E}_{\check{\theta}}-\mathbb{E}_{{\check{\theta}_{0}}}\big)\big[\ell_{G(\check{\theta}_{0})+\frac{h}{\sqrt{n}}}-\ell_{G(\check{\theta}_{0})}-\frac{h^{\top}}{\sqrt{n}}\nabla\ell_{G(\check{\theta}_{0})}\big]$. Fix $\varepsilon>0$. By the fact that $\sqrt{n}(\check{\theta}_{n}-\check{\theta}_{0})=O_{p}(1)$ and that $h_{n}=O_{p}^*(1)$, there exists $M>0$
    such that, $\|\check{\theta}_{0}\|_{2}\leq M$,  $\sup_{n}P\big(\|\sqrt{n}(\check{\theta}_{n}-\check{\theta}_{0})\|_{2}>M\big)\leq \varepsilon,$ and $\sup_{n}P^*(\|h_{n}\|_{2}>M)\leq \varepsilon$, then
    \begin{equation*}
        \begin{aligned}
            P^*(|R_{n}(\check{\theta}_{n},h_n)|>2\varepsilon)
            &\leq P^*\big(\sup_{\|h\|_{2}\leq M}\sup_{\sqrt{n}\|\check{\theta}-\check{\theta}_0\|_{2}\leq M}|R_{n}(\check{\theta},h)|>\varepsilon\big) 
            \\ &\quad + P\big(\|\sqrt{n}(\check{\theta}_{n}-\check{\theta}_{0})\|_{2}>M\big) + P^*(\|h_{n}\|_{2}>M).
            \\ &\leq  P^*\big(\sup_{\|h\|_{2}\leq M}\sup_{\sqrt{n}\|\check{\theta}-\check{\theta}_0\|_{2}\leq M}|R_{n}(\check{\theta},h)|>\varepsilon\big) + 2\varepsilon,
        \end{aligned}
    \end{equation*}
    Set $K=B(0,2M)\subset \mathbb{R}^{d}$, we have 
    \begin{equation*}
        P^*\big(\sup_{\|h\|_{2}\leq M}\sup_{\sqrt{n}\|\check{\theta}-\check{\theta}_0\|_{2}\leq M}|R_{n}(\check{\theta},h)|>\varepsilon\big)\leq \frac{2}{\sqrt{n}}\beta M^{3}L_{s,K}^\prime,
    \end{equation*}
    thus when $n\to \infty$, the first term tends to $0$. Therefore, we have proved that 
    \begin{equation*}
        n\big(\mathbb{E}_{\check{\theta}_{n}}-\mathbb{E}_{{\check{\theta}_{0}}}\big)\bigg[\ell_{G(\check{\theta}_{0})+\frac{h_{n}}{\sqrt{n}}}(Z)-\ell_{G(\check{\theta}_{0})}(Z)-\frac{h_{n}^{\top}}{\sqrt{n}}\nabla\ell_{G(\check{\theta}_{0})}(Z)\bigg] = o_{p}^*(1),
    \end{equation*}
\end{proof}

We now make the decomposition preceding Theorem~\ref{thm:cltrrm} precise.

\begin{theorem}[Restatement of Theorem~\ref{thm:cltrrm}]
Under Assumptions~\ref{as:pp} and~\ref{as:consistent2}, for each fixed
\(t\geq1\),
\[
\sqrt n(\widehat\theta_t-\theta_t)\xrightarrow{d}\mathcal N(0,V_t),
\]
where \(V_0=0\) and
 $
V_{t+1}=\nabla G(\theta_t)V_t\nabla G(\theta_t)^\top
+\Sigma_{\theta_t}(\theta_{t+1}),\qquad t\geq0.
$
\end{theorem}

\begin{proof}[Proof of Theorem~\ref{thm:cltrrm}]
    We prove $\sqrt{n}(\widehat{\theta}_{t}-\theta_{t})\to \mathcal{N}(0,V_{t})$ by induction on $t$, where $V_{t}$ is defined as in Theorem~\ref{thm:cltrrm}. 
    By Theorem 5.23 in \citep{van2000asymptotic}, $\sqrt{n}(\widehat{\theta}_{1}-\theta_{1})\xrightarrow{d} \mathcal{N}(0,V_{1})$,
     where $V_{1}=\Sigma_{\theta_{0}}(\theta_{1})$. 
     Assume that $\sqrt{n}(\widehat{\theta}_{t}-\theta_{t})\xrightarrow{d}\mathcal{N}(0,V_{t})$. We prove the result for $t+1$.
     Set $\widetilde{\theta}_{t+1,n}=G(\widehat{\theta}_{t})$. We first localize $\widehat{\theta}_{t+1}$ around $\widetilde{\theta}_{t+1,n}$.
    
     \textbf{Step 1. Asymptotic localization.} We show that $\sqrt{n}(\widehat{\theta}_{t+1}-\widetilde{{\theta}}_{t+1,n})=O_{p}(1)$.

    We follow the proof of Theorem 5.52 in \citep{van2000asymptotic}. 
    The induction hypothesis and the delta method imply that $\sqrt{n}(\widetilde{\theta}_{t+1,n}-\theta_{t+1})$ is asymptotically normal and hence bounded in probability. It remains to show $\sqrt{n}\big(\widehat{\theta}_{t+1}-\widetilde{\theta}_{t+1,n}\big)=O_{p}(1)$.

     Since $\widetilde{\theta}_{t+1,n}=G(\widehat{\theta}_{t})$ minimizes
     $\theta\mapsto\mathbb E_{\widehat\theta_t}[\ell_\theta(Z)]$ and
     $\widehat{\theta}_{t+1}$ minimizes
     $\theta\mapsto\widehat{\mathbb E}_{\widehat\theta_t,n}[\ell_\theta(Z)]$,
     \begin{equation*}
        \mathbb{E}_{\widehat{\theta}_{t}}[\nabla_{\theta}\ell_{\widetilde{\theta}_{t+1,n}}(Z)] = 0,
     \end{equation*}
     and 
    \begin{equation*}
        \widehat{\mathbb{E}}_{\widehat{\theta}_{t},n}\big(\ell_{\widehat{\theta}_{t+1}}-\ell_{\widetilde{\theta}_{t+1,n}}\big)\leq 0
    \end{equation*}

For $n\geq 1$, we define a partition of $\Theta\backslash \{\widetilde{\theta}_{t+1,n}\}$ by setting $S_{j}^{(n)} = \{\theta:2^{j-1}<\sqrt{n}\|\theta-\widetilde{\theta}_{t+1,n}\|\leq 2^{j}\}$ with $j$ ranging over integers.
    For any $\rho>0$, choose a compact subset $K\subset \Theta$ and set $A_{n}=\{\widehat{\theta}_{t}\in K, \widetilde{\theta}_{t+1,n}\in K\}$ so that $P(A_{n}^c)<\rho$ for all sufficiently large $n$. On $A_n$, both $\widehat{\theta}_{t}$ and $\widetilde{\theta}_{t+1,n}$ lie in $K$.

For fixed $M\in \mathbb{R}$, if $\sqrt{n}\|\widehat{\theta}_{t+1}-\widetilde{\theta}_{t+1,n}\|>2^{M}$, then $\widehat{\theta}_{t+1}\in S_{j}^{(n)}$ for some $j>M$, and
    \begin{equation*}
        \inf_{\theta\in S_{j}^{(n)}}\widehat{\mathbb{E}}_{\widehat{\theta}_{t},n}\big(\ell_{\theta}-\ell_{\widetilde{\theta}_{t+1,n}}\big)\leq  \widehat{\mathbb{E}}_{\widehat{\theta}_{t},n}\big(\ell_{\widehat{\theta}_{t+1}}-\ell_{\widetilde{\theta}_{t+1,n}}\big)\leq 0.
    \end{equation*}
    This implies that for any $\eta>0$, the following inequality holds,
    \begin{equation*}
        P(\sqrt{n}\|\widehat{\theta}_{t+1}-\widetilde{\theta}_{t+1,n}\|>2^{M})\leq \sum_{\substack{j>M\\ 2^{j}\leq\sqrt{n}\eta}}P^*\big(A_{n},\inf_{\theta\in S_{j}^{(n)}}\widehat{\mathbb{E}}_{\widehat{\theta}_{t},n}\big(\ell_{\theta}-\ell_{\widetilde{\theta}_{t+1,n}}\big)\leq 0\big) + P(2\|\widehat{\theta}_{t+1}-\widetilde{\theta}_{t+1,n}\|>\eta) + P(A_{n}^c).
    \end{equation*}
    By Proposition~\ref{prop:consistency}, the second term on the right hand side tends to $0$ when $n\to\infty$ for each $\eta>0$.
    By the strong convexity of $\ell$ in $\Theta$, we have
    \begin{equation*}
        {\mathbb{E}}_{\widehat{\theta}_{t}}\big(\ell_{\theta}-\ell_{\widetilde{\theta}_{t+1,n}}\big)\geq \frac{\gamma}{2} \|\theta-\widetilde{\theta}_{t+1,n}\|^{2},
    \end{equation*}
    Conditioning on $A_{n}$ and $\widehat{\theta}_{t}$,
    \begin{equation*}
        \begin{aligned}
            &\sum_{\substack{j>M\\ 2^{j}\leq\sqrt{n}\eta}}P^*\bigg(A_{n},\inf_{\theta\in S_{j}^{(n)}}\widehat{\mathbb{E}}_{\widehat{\theta}_{t},n}\big(\ell_\theta - \ell_{\widetilde{\theta}_{t+1,n}}\big)\leq 0\bigg|\widehat{\theta}_{t}\bigg)
            \\ \leq &\sum_{\substack{j>M\\ 2^{j}\leq\sqrt{n}\eta}}P^*\bigg(A_{n},\inf_{\theta\in S_{j}^{(n)}}\mathbb{G}_{n,\widehat{\theta}_{t}}\big(\ell_\theta-\ell_{\widetilde{\theta}_{t+1,n}}\big)\leq 
            - \frac{\gamma}{2} \sqrt{n}\inf_{\theta\in S_{j}^{(n)}}\|\theta-\widetilde{\theta}_{t+1,n}\|^{2}\bigg|\widehat{\theta}_{t}\bigg)
            \\ \lesssim &  \sum_{\substack{j>M\\ 2^{j}\leq\sqrt{n}\eta}}{2^{-2j}}{\sqrt{n}}\mathbb{E}^*\bigg[\mathbf{1}_{A_{n}}\cdot \big|\inf_{\theta\in S_{j}^{(n)}}\mathbb{G}_{n,\widehat{\theta}_{t}}\big(\ell_\theta-\ell_{\widetilde{\theta}_{t+1,n}}\big)\big|\bigg|\widehat{\theta}_{t}\bigg]
            \\ \lesssim &  \sum_{\substack{j>M\\ 2^{j}\leq\sqrt{n}\eta}}{2^{-2j}}{\sqrt{n}}\mathbb{E}^*\bigg[\mathbf{1}_{A_{n}}\sup_{\theta\in S_{j}^{(n)}}\big|\mathbb{G}_{n,\widehat{\theta}_{t}}(\ell_\theta-\ell_{\widetilde{\theta}_{t+1,n}})\big|\bigg|\widehat{\theta}_{t}\bigg]
        \end{aligned}
    \end{equation*}

    Note that when $j>M$ and $2^j\leq \sqrt{n}\eta$, the set $S_{j}^{(n)}\subset K^{2\eta}$, where $K^{2\eta}:= \{\theta:\text{dist}(\theta,K)\leq 2\eta\}$ is the compact enlargement of the fixed compact set $K$.
    Conditional on $A_{n}$ and $\widehat{\theta}_{t}$, $\widetilde{\theta}_{t+1,n}=G(\widehat{\theta}_{t})$ is fixed and lies in $K$.
    
    For $\delta\in (0,2\eta]$ and $\check{\theta}\in K$, set $\mathcal{F}_{\check{\theta},\delta}=\{\ell(\theta)-\ell(G(\check{\theta})):\|\theta-G(\check{\theta})\|_{2}\leq \delta,\theta\in K^{2\eta}\}$. On $A_{n}$, take $\check{\theta}=\widehat{\theta}_{t}$. Then $G(\check{\theta})\in K$, and $\mathcal{F}_{\check{\theta},\delta}$ has envelope $\delta L_{K^{2\eta}}$, where $L_{K^{2\eta}}$ is given by Assumption~\ref{as:pp}. Since $\sup_{\check{\theta}\in K^{2\eta}}\mathbb{E}_{\check{\theta}}\|L_{K^{2\eta}}(Z)\|^{2}<M_{1}^{2}$, Corollary 19.35 in \citep{van2000asymptotic} gives
    \begin{equation*}
        \begin{aligned}
            \mathbb{E}^*\bigg[\sup_{\|\theta-G(\widehat{\theta}_{t})\|\leq \delta}\big|\mathbb{G}_{n,\widehat\theta_t}(\ell_\theta-\ell_{G(\widehat{\theta}_{t})})\big|\bigg|\widehat{\theta}_{t}\bigg]\lesssim \mathcal{J}_{[]}\big(\delta \|L_{K^{2\eta}}(z)\|_{L_{2}(\widehat{\theta}_{t})},\mathcal{F}_{\widehat{\theta}_{t},\delta},L_{2}(\widehat{\theta}_{t})
        \big),
        \end{aligned}   
   \end{equation*}
   where for any $\check{\theta}\in \Theta$, $L_{2}(\check{\theta})$ stands for $L_{2}$ distance with respect to the distribution $\mathcal{D}(\check{\theta})$.
    By Example 19.7 in \citep{van2000asymptotic}, we know that there exists a constant $C$ that depends only on $K^{2\eta}$, $d$, and the following inequality holds uniformly for $\check{\theta}\in K$,
    \begin{equation*}
        \sup_{\check{\theta}\in K} N_{[]}(\varepsilon,\mathcal{F}_{\check{\theta},\delta},L_{2}({\check{\theta}}))\leq (\frac{C\delta M_{1}}{\varepsilon})^{d}.
    \end{equation*}
    Define the uniform metric entropy over $K$ as below for $a,\delta>0$:
    $$\mathcal{J}_{[]}\big(a,\mathcal{F}_{\check{\theta},\delta},L_{2}\big)=\int_{0}^{a}\sqrt{\log\sup_{\check{\theta}\in K}N_{[]}\big(\varepsilon,\mathcal{F}_{\check{\theta},\delta},L_{2}(\check{\theta})\big)}d\varepsilon,$$ 
    then the following inequality holds for some fixed constant $C_1$ depend on $K^{2\eta}, d$,  
    \begin{equation*}
        \mathcal{J}_{[]}\big(\delta M_{1} ,\mathcal{F}_{\check\theta,\delta},L_{2}\big)\leq \int_{0}^{\delta M_{1}}\sqrt{d\log(\frac{\delta M_{1}}{\varepsilon})+C_{1}}\mathrm{d}\varepsilon = \delta\int_{0}^{M_{1}}\sqrt{d\log(\frac{M_{1}}{\varepsilon})+C_{1}}\mathrm{d}\varepsilon.
    \end{equation*}
    Taking $\delta=\frac{2^{j+1}}{\sqrt{n}}$ gives, for any $M>0$,
    \begin{equation*}
        P^*\big(A_{n},\inf_{\theta\in S_{j}^{(n)}}\widehat{\mathbb{E}}_{\widehat{\theta}_{t},n}\big(\ell_\theta - \ell_{\widetilde{\theta}_{t+1,n}}\big)\leq 0\big)\lesssim 2^{-2j}\sqrt n\,\bE\bigg[\bE\big[\mathbf{1}_{A_{n}}\sup_{\theta\in S_{j}^{(n)}} |\bG_{n,\widehat{\theta}_{t}}(\ell_{\theta}-\ell_{\widetilde{\theta}_{t+1,n}})|\big|\widehat{\theta}_{t}\big]\bigg]
    \end{equation*}

    and thus
    \begin{equation*}
        \sum_{\substack{j>M \\ 2^{j}\leq\sqrt{n}\eta}}P^*\big(A_{n},\inf_{\theta\in S_{j}^{(n)}}\widehat{\mathbb{E}}_{\widehat{\theta}_{t},n}\big(\ell_\theta - \ell_{\widetilde{\theta}_{t+1,n}}\big)\leq 0\big)\lesssim \sum_{j>M}2^{-j} \le {2^{1-M}}.
    \end{equation*}
    Proposition~\ref{prop:consistency} gives $P(2\|\widehat{\theta}_{t+1}-\widetilde{\theta}_{t+1,n}\|>\eta)\to 0$. Hence
    \begin{equation*}
        \limsup_{n}P(\sqrt{n}\|\widehat{\theta}_{t+1}-\widetilde{\theta}_{t+1,n}\|>2^M)\lesssim \rho + 2^{1-M}.
    \end{equation*}

    Letting first $M\to\infty$ and then $\rho\to0$ proves that $\sqrt{n}(\widehat{\theta}_{t+1}-\widetilde{\theta}_{t+1,n})$ is bounded in probability.
    
    \textbf{Step 2. Asymptotic normality.}
    for a fixed function $f:\cZ\to \bR^{d}$, define 
    $g_{t,n}[f] := \sqrt n\left\{\widehat E_{\widehat\theta_t,n}[f(Z)]-E_{\theta_t}[f(Z)]\right\}.$
    Take $\hat{h}_{t+1,n}=\sqrt{n}(\widehat{\theta}_{t+1}-\theta_{t+1})$ and $\tilde{h}_{t+1,n} = -H_{\theta_{t}}(\theta_{t+1})^{-1}g_{t,n}\nabla\ell(\theta_{t+1})$. 
    By Step 1, we have $\sqrt{n}(\widehat{\theta}_{t+1}-\widetilde{\theta}_{t+1,n})=O_{p}(1)$, by induction hypothesis and applying Delta method to function $G$, we have
    $\sqrt{n}(\widetilde{\theta}_{t+1,n}-\theta_{t+1})=O_{p}(1)$. Combining them gives that $\hat{h}_{t+1,n}$ is $O_{p}(1)$.

    For $\check{\theta}\in \Theta$, set $f_{\check{\theta}}=H_{\theta_{t}}(\theta_{t+1})^{-1}\nabla_{\theta}\ell_{\theta_{t+1}}(Z)$ and $F(\check{\theta})=\bE_{\check{\theta}}f_{\check{\theta}}(Z)$. Assumption~\ref{as:consistent2} implies that $F$ is continuously differentiable, while Assumption~\ref{as:pp} gives the required third-moment bound. Note that by Assumption~\ref{as:consistent2}(a) and Lemma~\ref{rm:consistent},  
    we have for each fixed compact set $K_{1}$, \(\left\|\bE_{\vartheta'}[ff^\top]-\bE_\vartheta[ff^\top]
\right\|_2
\le
L'_{s,K_{1}}\|\vartheta'-\vartheta\|_2
\bE_\vartheta\|f\|_2^2.\) Together with continuously differentiability of $F$, we have $\text{Cov}_{\theta}f_{\theta}$ is continuous.
    Lemma~\ref{lm:rgd1} then implies that $\tilde{h}_{t+1,n}$ is asymptotically normal and hence $O_{p}^*(1)$.
    
    Now, by Lemma~\ref{lm:localexpansion1} and Lemma~\ref{lm:localexpansion2}, taking the random sequences in Lemma~\ref{lm:localexpansion1} to be $\check{\theta}_{n}=\hat{\theta}_{t}$ and $\theta_{n}\equiv\theta_{t+1}$ (we remark that $\hat{\theta}_{t}$ is a random sequence depending on sample number per-round $n$), we have
    \begin{equation*}
        \begin{aligned}
           &n\widehat{\mathbb{E}}_{\widehat{\theta}_{t},n}\big[\ell_{\theta_{t+1}+\hat{h}_{t+1,n}/\sqrt{n}}-\ell_{\theta_{t+1}}\big] = \frac{1}{2}\hat{h}_{t+1,n}^{\top}H_{\theta_{t}}(\theta_{t+1})\hat{h}_{t+1,n} + \hat{h}_{t+1,n}^{\top}g_{t,n}[\nabla_{\theta}\ell_{\theta_{t+1}}] + o_{p}(1)
           \\ &n\widehat{\mathbb{E}}_{\widehat{\theta}_{t},n}\big[\ell_{\theta_{t+1}+\widetilde{h}_{t+1,n}/\sqrt{n}}-\ell_{\theta_{t+1}}\big] = -\frac{1}{2}\big(g_{t,n}[\nabla\ell_{\theta_{t+1}}]\big)^{\top}H_{\theta_{t}}(\theta_{t+1})^{-1}\big(g_{t,n}[\nabla\ell_{\theta_{t+1}}]\big) + o_{p}(1). 
        \end{aligned}
    \end{equation*}
    Now by the above formulas and the definition of $\widehat{\theta}_{t+1}$, we have 
    \begin{equation*}
        \big(\hat{h}_{t+1,n}+H_{\theta_{t}}(\theta_{t+1})^{-1}g_{t,n}[\nabla \ell_{\theta_{t+1}}]\big)^{\top}H_{\theta_{t}}(\theta_{t+1})\big(\hat{h}_{t+1,n}+H_{\theta_{t}}(\theta_{t+1})^{-1}g_{t,n}[\nabla \ell_{\theta_{t+1}}]\big) + o_{p}(1)\leq 0,
    \end{equation*}
    since $H_{\theta_{t}}(\theta_{t+1})$ is positive definite, we have 
    \begin{equation*}
        \hat{h}_{t+1,n}+H_{\theta_{t}}(\theta_{t+1})^{-1}g_{t,n}[\nabla_\theta \ell_{\theta_{t+1}}] = o_{p}(1).
    \end{equation*}
    We have by Lemma~\ref{lm:rgd1} and Slutsky Theorem,
    \begin{equation*}
        \sqrt{n}(\widehat{\theta}_{t+1}-\theta_{t+1}) \to \mathcal{N}(0,\Sigma_{\theta_{t}}(\theta_{t+1})+\nabla G(\theta_{t})V_{t}\nabla G(\theta_{t})^{\top}\big),
    \end{equation*}
    The covariance on the right is \(V_{t+1}\) by the recursion in
    Theorem~\ref{thm:cltrrm}.
\end{proof}
\subsubsection{Proof of Corollary~\ref{col:rrmps}}
\begin{proof}[Proof of Corollary~\ref{col:rrmps}]
Since \(\varepsilon\beta<\gamma\), the contraction of \(G\) gives
\[
\|\theta_t-\theta_{\mathrm{PS}}\|_2
\leq\left(\frac{\varepsilon\beta}{\gamma}\right)^t
\|\theta_0-\theta_{\mathrm{PS}}\|_2
\leq B\left(\frac{\varepsilon\beta}{\gamma}\right)^t.
\]
Hence \(\theta_t\in\widehat{\mathcal R}_t\) implies
\[
\theta_{\mathrm{PS}}\in\widehat{\mathcal R}_t
+\mathcal B\left(0,B(\varepsilon\beta/\gamma)^t\right).
\]
Corollary~\ref{col:rrmfeasible} therefore gives
\[
\begin{aligned}
&\liminf_{n\to\infty}\mathbb P\left\{\theta_{\mathrm{PS}}\in
\widehat{\mathcal R}_t+
\mathcal B\left(0,B(\varepsilon\beta/\gamma)^t\right)\right\}\geq\lim_{n\to\infty}
\mathbb P(\theta_t\in\widehat{\mathcal R}_t)=1-\delta.
\end{aligned}
\]

For the bound after the corollary, the optimality of
\(\theta_{\mathrm{PS}}\) gives
\[
\mathbb E_{Z\sim\mathcal D(\theta_{\mathrm{PS}})}
[\nabla\ell(Z;\theta_{\mathrm{PS}})]=0.
\]
Strong convexity and the Cauchy--Schwarz inequality imply
\[
\begin{aligned}
\gamma\|\theta_0-\theta_{\mathrm{PS}}\|_2^2
&\leq
\mathbb E_{Z\sim\mathcal D(\theta_{\mathrm{PS}})}
[\nabla\ell(Z;\theta_0)]^\top(\theta_0-\theta_{\mathrm{PS}})\\
&\leq\sup_{z\in\mathcal Z}\|\nabla\ell(z;\theta_0)\|_2
\|\theta_0-\theta_{\mathrm{PS}}\|_2.
\end{aligned}
\]
Dividing by \(\gamma\|\theta_0-\theta_{\mathrm{PS}}\|_2\) proves the
stated bound when \(\theta_0\neq\theta_{\mathrm{PS}}\); the remaining
case is immediate.
\end{proof}

%% file: Sections/appA_score.tex
\subsubsection{Regularity for Score Matching}
The following conditions justify differentiating under the integral in
Lemma~\ref{lm:K}.

\begin{assumption}\label{as:exchange}
For each \(j\in[d]\), \(\theta\in\Theta\), and \(\psi\in\Psi\), the
following hold, where integrability is with respect to the dominating
measure in Assumption~\ref{as:consistent2}(a).

(a) The functions \(z\mapsto p(z;\theta)s_j(z,\theta;\psi)\) and
\(z\mapsto p(z;\theta)\,\partial_{\theta^{(j)}}s_j(z,\theta;\psi)\) are
integrable.

(b) For almost every \(z\), the map
\(\theta\mapsto p(z;\theta)s_j(z,\theta;\psi)\) is differentiable in
\(\theta^{(j)}\).

(c) For \(\theta'\) in a neighborhood of \(\theta\),
\(\bigl|\partial_{\theta^{(j)}}\{p(z;\theta')s_j(z,\theta';\psi)\}\bigr|\)
is bounded by an integrable function of \(z\) that does not depend on
\(\theta'\).
\end{assumption}

\subsubsection{Proof of Lemma~\ref{lm:K}}
\begin{proof}[Proof of Lemma~\ref{lm:K}]
Integrals are with respect to the dominating measure in
Assumption~\ref{as:consistent2}(a). Fix \(j\in[d]\). Since
\(p(z;\theta)\,\partial_{\theta^{(j)}}\log p(z;\theta)
=\partial_{\theta^{(j)}}p(z;\theta)\), the product rule gives
\[
p(z;\theta)\,\partial_{\theta^{(j)}}\log p(z;\theta)\,s_j(z,\theta;\psi)
=\partial_{\theta^{(j)}}\bigl\{p(z;\theta)s_j(z,\theta;\psi)\bigr\}
-p(z;\theta)\,\partial_{\theta^{(j)}}s_j(z,\theta;\psi).
\]
Integrating over \(z\), and moving the derivative outside the integral by
Assumption~\ref{as:exchange}, we obtain
\[
\mathbb E_{Z\sim\mathcal D(\theta)}
\bigl[\partial_{\theta^{(j)}}\log p(Z;\theta)\,s_j(Z,\theta;\psi)\bigr]
=\frac{\partial}{\partial\theta^{(j)}}
\mathbb E_{Z\sim\mathcal D(\theta)}[s_j(Z,\theta;\psi)]
-\mathbb E_{Z\sim\mathcal D(\theta)}
[\partial_{\theta^{(j)}}s_j(Z,\theta;\psi)].
\]
Summing over \(j\in[d]\) proves the lemma.
\end{proof}

%% file: Sections/appA_perturb.tex
\subsubsection{Policy Perturbation}\label{app:policy_perturbation}
This subsection gives the assumptions and proofs for the policy-perturbation estimator in Section~\ref{sec:RRM}.

For a deterministic \(\theta\), let \(\widehat J_{n,k}(\theta;\psi)\)
denote the expression in \eqref{eq:smobjective}, computed from \(n\)
observations from \(\mathcal D(\theta)\) and \(k\) observations from each
\(\mathcal D(\theta+\eta e_j)\), \(j\in[d]\), plus the constant
\[
\mathbb E_{Z\sim\mathcal D(\theta)}
[\|\nabla_\theta\log p(Z;\theta)\|_2^2].
\]
The constant does not change the minimizer, so the estimator in
Section~\ref{subsec:RRMcov} is also an approximate minimizer of
\(\widehat J_{n,k}\). Define the squared score error
\[
J(\theta;\psi):=\mathbb E_{Z\sim\mathcal D(\theta)}
\bigl[\|\nabla_\theta\log p(Z;\theta)-s(Z,\theta;\psi)\|_2^2\bigr]
\]
and its finite-difference version
\[
\begin{aligned}
J_\eta(\theta;\psi):={}&
\mathbb E_{Z\sim\mathcal D(\theta)}
\left[\|s(Z,\theta;\psi)\|_2^2
+2\sum_{j=1}^d\partial_{\theta^{(j)}}s_j(Z,\theta;\psi)
+\|\nabla_\theta\log p(Z;\theta)\|_2^2\right]\\
&-\frac2\eta\sum_{j=1}^d\left\{
\mathbb E_{Z\sim\mathcal D(\theta+\eta e_j)}
[s_j(Z,\theta+\eta e_j;\psi)]
-\mathbb E_{Z\sim\mathcal D(\theta)}[s_j(Z,\theta;\psi)]\right\}.
\end{aligned}
\]
Lemma~\ref{lm:K} gives \(J_\eta(\theta;\psi)\to J(\theta;\psi)\) as
\(\eta\to0\). The next assumptions make this convergence uniform over the parameters used below. In the following context, for $R>0$, we define $\Psi_{R}:=\{\psi\in \Psi:\|\psi\|_{2}\leq R\}$.

\begin{assumption}\label{as:perturb1}
We assume that the score function $s$, distribution map $\mathcal{D}$ have the following properties:

(a). For any compact subset $K\subset \Theta$, $R>0$, there is a positive constant $C_{K,R}>0$, such that 
    \begin{equation*}
        \sup_{\theta\in K}\sup_{\|\psi\|_{2}<R}
        \left|\frac{\partial^2}{\partial(\theta^{(i)})^2}
        \mathbb{E}_{\theta}[s_{i}(z,\theta;\psi)]\right|
        \leq C_{K,R}<\infty.
    \end{equation*}

(b). (\textit{Enveloping function}) For any compact subset $K$, and $R>0$, there exists a function $L_{K,R}:\mathcal{Z}\to \mathbb{R}$ and $B_{K,R}:\mathcal{Z}\to \mathbb{R}$, such that: denote 
$$g_{0}(z,\theta;\psi):=\|s(z,\theta;\psi)\|^{2}_{2},\ g_{1}^{(i)}(z,\theta,\psi) = s_{i}(z,\theta;\psi),\ g_{2}^{(i)}(z,\theta;\psi) = \partial_{\theta_i}s_{i}(z,\theta;\psi),\ i\in [d],$$
then for any function $g(z,\theta;\psi)\in \{g_{0}(z,\theta;\psi),g_{1}^{(i)}(z,\theta;\psi),g_{2}^{(i)}(z,\theta;\psi):i\in [d]\}$  for any $z\in \mathcal{Z}$, $\theta,\theta^\prime\in K$, $\psi,\psi^\prime\in B_{\Psi}(R)$,  the following inequality holds:

\begin{equation*}
    |g(z,\theta,\psi)|\leq B_{K,R}(z),
\end{equation*}
\begin{equation*}
    |g(z,\theta,\psi)-g(z,\theta^\prime,\psi^\prime)| \leq L_{K,R}(z)\big(\|\theta-\theta^\prime\|_{2}+\|\psi-\psi^\prime\|_{2}\big),
\end{equation*}
with $\sup_{\theta\in K^1}\mathbb{E}_{\theta}[B_{K,R}{(Z)}^{2}+L_{K,R}{(Z)}^{2}]<\infty,$ where $K^1$ is the $1$-enlargement of $K\subset \Theta$.

(c). (\textit{Richness of class.}) For any compact set $K\subset\Theta$, there exists $R_{K}>0$, 
such that for each $\theta\in K$, there exists a $\psi^{\ast}(\theta)\in \Psi_{R_{K}}$ satisfying
\begin{equation*}
    s(z,\theta;\psi^{\ast}(\theta)) = \nabla_{\theta}\log p(z;\theta).
\end{equation*}

(d). (\textit{Approximate minimization and tightness.}) For deterministic
\(\theta\), let \(\widehat\psi_{n,k}(\theta)\) be the empirical
score-matching estimator. For every compact \(K\subset\Theta\), it can be
selected so that, for some \(a_{n,k}\to0\),
\[
\sup_{\theta\in K}\mathbb P\left\{
\widehat J_{n,k}(\theta;\widehat\psi_{n,k}(\theta))
>\inf_{\psi\in\Psi}\widehat J_{n,k}(\theta;\psi)+a_{n,k}\right\}
\longrightarrow0
\]
as \(\min(n,k)\to\infty\). In addition,
\[
\lim_{R\to\infty}\limsup_{\min(n,k)\to\infty}
\sup_{\theta\in K}\mathbb P\{
\|\widehat\psi_{n,k}(\theta)\|_2>R\}=0.
\]
\end{assumption}

\begin{assumption}\label{as:perturb2}
For every compact \(K\subset\Theta\), the following conditions hold.

(a) \((\widetilde\theta,\theta)\mapsto
H_{\widetilde\theta}(\theta)\) is Lipschitz on \(K\times K\).

(b) \((\widetilde\theta,\theta)\mapsto
\mathbb E_{Z\sim\mathcal D(\widetilde\theta)}
[\nabla\ell(Z;\theta)\nabla_\theta\log p(Z;\widetilde\theta)^\top]\)
is Lipschitz on \(K\times K\).

(c) There exists a measurable \(L_K^H:\mathcal Z\to[0,\infty)\) such that,
for all \(\theta,\theta'\in K\) and \(z\in\mathcal Z\),
\[
\|\nabla^2\ell(z;\theta)-\nabla^2\ell(z;\theta')\|_2
\leq L_K^H(z)\|\theta-\theta'\|_2,
\]
and
$
\sup_{\widetilde\theta\in K}
\mathbb E_{Z\sim\mathcal D(\widetilde\theta)}[L_K^H(Z)]<\infty.
$
\end{assumption}
\begin{assumption}\label{as:perturb3}
    For each compact subset $K\subset\Theta$, constant $R>0$, define 
    $$h_{ij}(z;\theta,\check{\theta};\psi):=\partial_{\theta^{(i)}}\ell(z;\theta) s_{(j)}(z,\check{\theta};\psi),$$
    where $i,j\in [d]$. We assume that there exists function $Q_{K,R}:\mathcal{Z}\to \mathbb{R}$, and $L_{K,R}^{\text{prod}}:\mathcal{Z}\to \mathbb{R}$, such that for any $\theta,\check{\theta},\theta^\prime,\check{\theta}^\prime\in K$, $\psi,\psi^\prime\in \Psi_{R}$,
    \begin{equation*}
        \begin{aligned}
            &|h_{ij}(z;\theta,\check{\theta};\psi)|\leq Q_{K,R}(z),
            \\& |h_{ij}(z;\theta,\check{\theta};\psi)-h_{ij}(z;\theta^\prime,\check{\theta}^\prime;\psi^\prime)|\leq L_{K,R}^{\text{prod}}(z)\big(\|\theta-\theta^\prime\|_{2}+\|\check{\theta}-\check{\theta}^\prime\|_{2} + \|\psi-\psi^\prime\|_{2}\big),
        \end{aligned}
    \end{equation*}
    with $\sup_{\check{\theta}\in K}\mathbb{E}_{\check{\theta}}[Q_{K,R}(Z)^{2}+L_{K,R}^{\text{prod}}(Z)^{2}]<\infty$.
\end{assumption}

\subsubsection{Proof of Theorem~\ref{thm:discuss1}}

\begin{lemma}\label{lm:perturb1}
      Under Assumption~\ref{as:exchange}  and \ref{as:perturb1}(a), we have for any compact set $K$ and $R>0$,
    \begin{equation*}
        \sup_{\theta\in K}\sup_{\psi\in {\Psi}_R}|J(\theta;\psi)-{J}_{\eta}(\theta;\psi)|\leq 2d\eta C_{K^{2\eta},R}.
    \end{equation*}
    where $C_{K^{2\eta},R}$ is the constant in Assumption~\ref{as:perturb1} and $K^{2\eta}$ is the compact enlargement $K^{2\eta}:=\{\theta:\text{dist}(\theta,K)\leq 2\eta\}$ of $K$. 
\end{lemma}
\begin{proof}
By mean-value theorem, there exists $\eta^{(i)}\in [0,\eta]$, $1\leq i\leq d$ , such that 
\begin{equation*}
    \begin{aligned}
        \big|J(\theta;\psi)-{J}_{\eta}(\theta;\psi)\big|&= 2\bigg|\sum_{i=1}^{d}\frac{1}{\eta}\bigg({\mathbb{E}}_{\theta+\eta e_i}\big[s_{i}(z,\theta+\eta e_i;\psi)\big]-{\mathbb{E}}_{\theta}\big[s_{i}(z,\theta;\psi)\big]\bigg) - \sum_{i=1}^{d}\big[\frac{\partial}{\partial \theta^{(i)}}\mathbb{E}_{\theta}[s_{i}(z,\theta;\psi)]\big]\bigg|
        \\& = 2\bigg|\sum_{i=1}^{d}\big[\frac{\partial}{\partial \theta^{(i)}}\mathbb{E}_{\theta+\eta^{(i)}e_i}[s_{i}(z,\theta+\eta^{(i)}e_i;\psi)]\big]-
        \sum_{i=1}^{d}\big[\frac{\partial}{\partial \theta^{(i)}}\mathbb{E}_{\theta}[s_{i}(z,\theta;\psi)]\big]\bigg|
        \\& \leq 2C_{K^{2\eta},R}(\sum_{i=1}^{d}\eta^{(i)})
        \\&\leq 2d\eta C_{K^{2\eta},R},
    \end{aligned}
\end{equation*}
the first inequality follows from a direct application of mean-value theorem.
\end{proof}

\begin{theorem}\label{thm:discuss1}
    For fixed \(T\geq1\), under Assumptions~\ref{as:pp},
    \ref{as:exchange}, and~\ref{as:perturb1},
    \begin{equation*}
        \max_{0\leq j\leq T}J(\widehat\theta_j;\widehat\psi_{n,k}(\widehat\theta_j))
        = O_P\left(\frac{1}{\sqrt{n}}+\frac{1}{\eta\sqrt{\min(n,k)}}+\eta+a_{n,k}\right).
    \end{equation*}
\end{theorem}
\begin{proof}
    Fix \(\delta>0\). Proposition~\ref{prop:consistency} gives a compact set
    \(K\) such that
    \[
    P\{\widehat\theta_j\in K,\ 0\leq j\leq T\}>1-\delta.
    \]
    Choose \(R\) larger than the constant in
    Assumption~\ref{as:perturb1}(c) for this set \(K\). By
    Assumption~\ref{as:perturb1}(d), \(R\) may be chosen so that, uniformly
    over deterministic \(\theta\in K\), the fitted parameter belongs to
    \(\Psi_R\) and satisfies the approximate-minimization inequality with
    probability tending to one.

    By Assumption~\ref{as:perturb1} and Dudley's uniform entropy bound, c.f. Corollary 19.35 in \citep{van2000asymptotic}, we have
    \begin{equation*}
        \begin{aligned}
            \sup_{\theta\in K}\mathbb{E}_{\theta}\bigg[\sup_{\psi\in \Psi_{R}}\bigg|\widehat{\mathbb{E}}_{\theta,n}\big[\|s(z,\theta;\psi)\|^{2}\big]-\mathbb{E}_{\theta}\big[\|s(z,\theta;\psi)\|^{2}\big]\bigg|\bigg]&\lesssim \frac{1}{\sqrt{n}},
            \\ \sup_{\theta\in K}\mathbb{E}_{\theta}\bigg[\sup_{\psi\in \Psi_{R}}\bigg|\widehat{\mathbb{E}}_{\theta,n}\big[\dfrac{\partial}{\partial \theta^{(i)}}s_{i}(z,\theta;\psi)\big]-\mathbb{E}_{\theta}\big[\dfrac{\partial}{\partial \theta^{(i)}}s_{i}(z,\theta;\psi)\big]\bigg|\bigg]&\lesssim \frac{1}{\sqrt{n}},
            \\ \sup_{\theta\in K}\mathbb{E}_{\theta}\bigg[\sup_{\psi\in \Psi_{R}}\bigg|\widehat{\mathbb{E}}_{\theta,k}\big[s_{i}(z,\theta;\psi)\big]-\mathbb{E}_{\theta}\big[s_{i}(z,\theta;\psi)\big]\bigg|\bigg]&\lesssim \frac{1}{\sqrt{k}}.
        \end{aligned}
    \end{equation*}
    The same bounds apply to the samples from
    \(\mathcal D(\theta+\eta e_i)\), because these parameters belong to the
    compact enlargement of \(K\) for all sufficiently small \(\eta\).
    Markov's inequality and Lemma~\ref{lm:perturb1} therefore give, for every
    \(\varepsilon>0\), a constant \(C<\infty\) such that, for all
    sufficiently large \(n,k\),
    \[
    \sup_{\theta\in K}P\left\{
    \sup_{\psi\in\Psi_R}|\widehat J_{n,k}(\theta;\psi)-J(\theta;\psi)|
    >C\left(\frac1{\sqrt n}+\frac1{\eta\sqrt{\min(n,k)}}+\eta\right)
    \right\}<\varepsilon.
    \]
    We apply this uniform tail bound separately at each trajectory point.

    Conditional on all observations preceding batch \(j\),
    \(\widehat\theta_j\) is fixed and the samples used to fit
    \(\widehat\psi_{n,k}(\widehat\theta_j)\) are drawn at that deployment
    parameter. The preceding bound and Assumption~\ref{as:perturb1}(d)
    therefore apply at \(\widehat\theta_j\). Since \(T\) is fixed, a union
    bound makes them hold simultaneously for \(0\leq j\leq T\). On this
    event, let \(\psi^\ast(\widehat\theta_j)\in\Psi_R\) be given by
    Assumption~\ref{as:perturb1}(c). Then
    \[
    \begin{aligned}
    J(\widehat\theta_j;\widehat\psi_{n,k}(\widehat\theta_j))
    &\leq 2\sup_{\psi\in\Psi_R}
    |\widehat J_{n,k}(\widehat\theta_j;\psi)
    -J(\widehat\theta_j;\psi)|+a_{n,k},
    \end{aligned}
    \]
    because \(J(\widehat\theta_j;\psi^\ast(\widehat\theta_j))=0\).
    Taking the maximum over the fixed set of indices proves the result.

\end{proof}

\subsubsection{Proof of Consistency of Perturbation-based estimator (Theorem~\ref{thm:discuss2})}
\begin{proof}[Proof of Theorem~\ref{thm:discuss2}]
By Assumption~\ref{as:pp}(d),
\(\widehat H_{\widehat\theta_t}(\widehat\theta_{t+1})\succeq\gamma I_d\)
and \(H_{\widehat\theta_t}(\widehat\theta_{t+1})\succeq\gamma I_d\).
Both inverses therefore have spectral norm at most \(1/\gamma\).

We compare the estimator with \(\nabla G(\theta_t)\) through six terms.
Fix \(\delta>0\). Proposition~\ref{prop:consistency} gives a compact set
\(K\) such that
\[
\mathbb P\{\widehat\theta_t,\widehat\theta_{t+1},
\theta_t,\theta_{t+1}\in K\}>1-\delta.
\]
The uniform tail bound in Assumption~\ref{as:perturb1}(d) then allows us
to choose \(R>0\) such that
\[
\mathbb P\{\|\widehat\psi_{n,k}(\widehat\theta_t)\|_2\le R\}\geq\delta
\]
for all sufficiently large \(n,k\). We work on the intersection of these
two events.

Let 
\begin{equation*}
    \widehat{\nabla G}_{t,1} := -{H}_{\widehat{\theta}_{t}}(\widehat{\theta}_{t+1})^{-1}\widehat{\mathbb{E}}_{\widehat{\theta}_{t},n}\big[\nabla_{\theta}\ell(z;\widehat{\theta}_{t+1})s(z,\widehat{\theta}_{t};\widehat{\psi}_{n,k}(\widehat{\theta}_{t}))^{\top}\big].
\end{equation*}

\textbf{Step 1: Convergence of $\|\widehat{\nabla G}_{t,1}-\widehat{\nabla G}_{t}\|_{2}$}

\begin{equation*}
    \begin{aligned}
        \|\widehat{\nabla G}_{t,1}-\widehat{\nabla G}_{t}\|_{2}&\lesssim \big\|{H}_{\widehat{\theta}_{t}}(\widehat{\theta}_{t+1})^{-1} - \widehat{H}_{\widehat{\theta}_{t}}(\widehat{\theta}_{t+1})^{-1}\big\|_{2}\cdot\widehat{\mathbb{E}}_{\widehat{\theta}_{t},n}\big[\big\|\nabla_{\theta}\ell(z;\widehat{\theta}_{t+1})s(z,\widehat{\theta}_{t};\widehat{\psi}_{n,k}(\widehat{\theta}_{t}))^\top\big\|_{2}\big],
    \end{aligned}
\end{equation*}
We next verify the rate for the empirical Hessian. By
Assumption~\ref{as:perturb2}(c),
\begin{equation*}
\begin{aligned}
&\|\widehat H_{\widehat\theta_t}(\widehat\theta_{t+1})
-H_{\widehat\theta_t}(\widehat\theta_{t+1})\|_2\\
&\quad\leq
\left\{\frac1n\sum_{i=1}^n L_K^H(Z_{t,i})
+\mathbb E_{\widehat\theta_t}[L_K^H(Z)]\right\}
\|\widehat\theta_{t+1}-\theta_{t+1}\|_2\\
&\qquad+
\left\|\frac1n\sum_{i=1}^n\nabla^2\ell(Z_{t,i};\theta_{t+1})
-H_{\widehat\theta_t}(\theta_{t+1})\right\|_2.
\end{aligned}
\end{equation*}
The quantity in braces is \(O_P(1)\). Conditional on the preceding
observations, the last term is an average of independent centered matrices.
Assumption~\ref{as:pp}(c) bounds their spectral norms, so this term is
\(O_P(n^{-1/2})\). Theorem~\ref{thm:cltrrm} now gives
\[
\|\widehat H_{\widehat\theta_t}(\widehat\theta_{t+1})
-H_{\widehat\theta_t}(\widehat\theta_{t+1})\|_2=O_P(n^{-1/2}).
\]
Since both Hessians are bounded below by \(\gamma I_d\),
\[
\begin{aligned}
&\|H_{\widehat\theta_t}(\widehat\theta_{t+1})^{-1}
-\widehat H_{\widehat\theta_t}(\widehat\theta_{t+1})^{-1}\|_2\leq\gamma^{-2}
\|\widehat H_{\widehat\theta_t}(\widehat\theta_{t+1})
-H_{\widehat\theta_t}(\widehat\theta_{t+1})\|_2
=O_P(n^{-1/2}).
\end{aligned}
\]
Assumption~\ref{as:perturb3} also gives
\[
\widehat{\mathbb E}_{\widehat\theta_t,n}
\big[\|\nabla\ell(z;\widehat\theta_{t+1})
s(z,\widehat\theta_t;\widehat\psi_{n,k}(\widehat\theta_t))^\top\|_2\big]
=O_P(1).
\]
This implies that 
\begin{equation*}
    \|\widehat{\nabla G}_{t,1}-\widehat{\nabla G}_{t}\|_{2} = O_P(\frac{1}{\sqrt{n}}).
\end{equation*}
Let 
    \begin{equation*}
        \widehat{\nabla G}_{t,2} := -H_{\widehat{\theta}_{t}}(\widehat{\theta}_{t+1})^{-1}\widehat{\mathbb{E}}_{\widehat{\theta}_{t},n}\big[\nabla_{\theta}\ell(z;{\theta}_{t+1})s(z,\widehat{\theta}_{t};\widehat{\psi}_{n,k}(\widehat{\theta}_{t}))^{\top}\big].
    \end{equation*}
\textbf{Step 2: Convergence of $\|\widehat{\nabla G}_{t,1}-\widehat{\nabla G}_{t,2}\|$}.

    Since \(\|H_{\widehat\theta_t}(\widehat\theta_{t+1})^{-1}\|_2
    \leq1/\gamma\),
    we have
    \begin{equation*}
        \begin{aligned}
            \|\widehat{\nabla G}_{t,1}-\widehat{\nabla G}_{t,2}\|_{2}&\lesssim \|\widehat{\theta}_{t+1}-\theta_{t+1}\|_{2}\widehat{\mathbb{E}}_{\widehat{\theta}_{t},n}\big[\big\|s(z,\widehat{\theta}_{t};\widehat{\psi}_{n,k}(\widehat{\theta}_{t}))\big\|_{2}\big]\\ &=\frac{1}{n}\|\widehat{\theta}_{t+1}-\theta_{t+1}\|_{2}\sum_{l=1}^{n}\big\|s(Z_{t,l},\widehat{\theta}_{t};\widehat{\psi}_{n,k}(\widehat{\theta}_{t}))\big\|_{2}.
        \end{aligned}
    \end{equation*}
    Assumption~\ref{as:perturb1}(b) bounds the last sample average uniformly
    on \(K\times\Psi_R\), so it is \(O_P(1)\). Therefore,
    \begin{equation*}
        \|\widehat{\nabla G}_{t,1}-\widehat{\nabla G}_{t,2}\|_{2} = O_P(\|\widehat{\theta}_{t+1}-\theta_{t+1}\|_{2}),
    \end{equation*}
    thus combining the above result with Theorem \ref{thm:cltrrm} gives
 $
        \|\widehat{\nabla G}_{t,1}-\widehat{\nabla G}_{t,2}\|_{2} = O_P(\frac{1}{\sqrt{n}}).
$   
    Let 
    \begin{equation*}
        \widehat{\nabla G}_{t,3} := -H_{\widehat{\theta}_{t}}(\widehat{\theta}_{t+1})^{-1}\mathbb{E}_{\widehat{\theta}_{t}}\big[\nabla_{\theta}\ell(z;{\theta}_{t+1})s(z,\widehat{\theta}_{t};\widehat{\psi}_{n,k}(\widehat{\theta}_{t}))^{\top}\big].
    \end{equation*}
\textbf{Step 3: Convergence of $\|\widehat{\nabla G}_{t,2}-\widehat{\nabla G}_{t,3}\|$}.

    By the uniform empirical-process bound from
    Assumption~\ref{as:perturb3}, we have
    \begin{equation*}
        \begin{aligned}
            \|\widehat{\nabla G}_{t,2}-\widehat{\nabla G}_{t,3}\|_{2}&\lesssim \bigg\|\widehat{\mathbb{E}}_{\widehat{\theta}_{t},n}\big[\nabla_{\theta}\ell(z;{\theta}_{t+1})s(z,\widehat{\theta}_{t};\widehat{\psi}_{n,k}(\widehat{\theta}_{t}))^{\top}\big]-\mathbb{E}_{\widehat{\theta}_{t}}\big[\nabla_{\theta}\ell(z;{\theta}_{t+1})s(z,\widehat{\theta}_{t};\widehat{\psi}_{n,k}(\widehat{\theta}_{t}))^{\top}\big]\bigg\|_{2}
            \\&=O_P(\frac{1}{\sqrt{n}}).
        \end{aligned}
    \end{equation*}

Let 
\begin{equation*}
        \widehat{\nabla G}_{t,4} := -H_{\theta_{t}}(\theta_{t+1})^{-1}\mathbb{E}_{\widehat{\theta}_{t}}\big[\nabla_{\theta}\ell(z;{\theta}_{t+1})s(z,\widehat{\theta}_{t};\widehat{\psi}_{n,k}(\widehat{\theta}_{t}))^{\top}\big].
    \end{equation*}
\textbf{Step 4: Convergence of $\|\widehat{\nabla G}_{t,3}-\widehat{\nabla G}_{t,4}\|_{2}$}.
By Assumption~\ref{as:perturb3}, we have
$$\mathbb{E}_{\widehat{\theta}_{t}}\big[\nabla_{\theta}\ell(z;{\theta}_{t+1})s(z,\widehat{\theta}_{t};\widehat{\psi}_{n,k}(\widehat{\theta}_{t}))^{\top}\big]=O_P(1),$$
thus
\begin{equation*}
    \begin{aligned}
        \|\widehat{\nabla G}_{t,3}-\widehat{\nabla G}_{t,4}\|_{2}
        &=O_P\big(\|H_{\theta_{t}}(\theta_{t+1})^{-1}-H_{\widehat{\theta}_{t}}(\widehat{\theta}_{t+1})^{-1}\|_{2}\big)
        \\&=O_P\bigg(\big\|H_{\theta_{t}}(\theta_{t+1})^{-1}\big(H_{\theta_{t}}(\theta_{t+1})-H_{\widehat{\theta}_{t}}(\widehat{\theta}_{t+1})\big)H_{\widehat{\theta}_{t}}(\widehat{\theta}_{t+1})^{-1}\big\|_{2}\bigg),
    \end{aligned}
\end{equation*}
Using Assumption~\ref{as:pp}(d) and Assumption~\ref{as:perturb2}(a),
\begin{equation*}
    \|\widehat{\nabla G}_{t,3}-\widehat{\nabla G}_{t,4}\|_{2}\lesssim \|\widehat{\theta}_{t}-\theta_{t}\|_{2} + \|\widehat{\theta}_{t+1}-\theta_{t+1}\|_{2},
\end{equation*}
thus by Theorem \ref{thm:cltrrm},
\begin{equation*}
    \|\widehat{\nabla G}_{t,3}-\widehat{\nabla G}_{t,4}\|_{2} = O_P(\frac{1}{\sqrt{n}}). 
\end{equation*}
Let 
\begin{equation*}
    \widehat{\nabla G}_{t,5} := -H_{\theta_{t}}(\theta_{t+1})^{-1}\mathbb{E}_{\widehat{\theta}_{t}}\big[\nabla_{\theta}\ell(z;{\theta}_{t+1})\nabla_{\theta}\log p(z;\widehat\theta_{t})^\top\big].
\end{equation*}
\textbf{Step 5: Convergence of $\|\widehat{\nabla G}_{t,4}-\widehat{\nabla G}_{t,5}\|_{2}$}.

By Assumption~\ref{as:pp}(b), the loss gradients have uniformly bounded
second moments on \(K\), and
\begin{equation*}
\begin{aligned}
    \|\widehat{\nabla G}_{t,4}-\widehat{\nabla G}_{t,5}\|^{2}&\lesssim \bigg\{\mathbb{E}_{\widehat{\theta}_{t}}\bigg[\big\|\nabla_{\theta}\ell(z;\theta_{t+1})\big\|_{2}\times \big\|s(z,\widehat{\theta}_{t};\widehat{\psi}_{n,k}(\widehat{\theta}_{t}))-\nabla_{\theta}\log p(z;\widehat\theta_{t})\big\|_{2}\bigg]\bigg\}^{2}
    \\&\leq \mathbb{E}_{\widehat{\theta}_{t}}\big[\big\|\nabla_{\theta}\ell(z;\theta_{t+1})\big\|_{2}^{2}\big]
    \\&\ \times \mathbb{E}_{\widehat{\theta}_{t}}\big[\big\|s(z,\widehat{\theta}_{t};\widehat{\psi}_{n,k}(\widehat{\theta}_{t}))-\nabla_{\theta}\log p(z;\widehat\theta_{t})\big\|_{2}^{2}\big]\quad \mathrm{(by\ Cauchy-Schwarz\ inequality)}
    \\& \lesssim \mathbb{E}_{\widehat{\theta}_{t}}\big[\big\|s(z,\widehat{\theta}_{t};\widehat{\psi}_{n,k}(\widehat{\theta}_{t}))-\nabla_{\theta}\log p(z;\widehat\theta_{t})\big\|_{2}^{2}\big]\quad \mathrm{(by\ compactness\ of\ K)}
    \\& = O_P(\frac{1}{\sqrt{n}}+\frac{1}{\eta\sqrt{\min(n,k)}}+\eta+a_{n,k})\quad \mathrm{(by\ Theorem\ \ref{thm:discuss1})}.
\end{aligned}
\end{equation*}

\textbf{Step 6: Convergence of $\|\widehat{\nabla G}_{t}-\nabla G(\theta_{t})\|$}.

Finally, recall that 
\begin{equation*}
    \nabla G(\theta_{t}) = -H_{\theta_{t}}(\theta_{t+1})^{-1}\mathbb{E}_{{\theta}_{t}}\big[\nabla_{\theta}\ell(z;{\theta}_{t+1})\nabla_{\theta}\log p(z;\theta_{t})^\top\big],
\end{equation*}
Assumption~\ref{as:pp}(d) and Assumption~\ref{as:perturb2}(b) give
\[
\|\widehat{\nabla G}_{t,5}-\nabla G(\theta_t)\|_2
\lesssim\|\widehat\theta_t-\theta_t\|_2
=O_P(n^{-1/2}),
\]
where the last equality follows from Theorem~\ref{thm:cltrrm}. Combining the six bounds gives
\begin{equation*}
     \|\widehat{\nabla G}_{t}-\nabla G(\theta_{t})\|_{2}^2 = O_P(\frac{1}{\sqrt{n}}+\frac{1}{\eta\sqrt{\min(n,k)}}+\eta+a_{n,k}).
\end{equation*}

\end{proof}

\subsubsection{Proof of Corollary~\ref{col:rrmfeasible}}
\begin{proof}[Proof of Corollary~\ref{col:rrmfeasible}]
Fix \(t\geq0\). Theorem~\ref{thm:cltrrm} gives
\(\widehat\theta_t\xrightarrow{P}\theta_t\) and
\(\widehat\theta_{t+1}\xrightarrow{P}\theta_{t+1}\). Let \(K\) be a
compact convex set whose interior contains \(\theta_t\) and
\(\theta_{t+1}\). With probability tending to one,
\(\widehat\theta_t,\widehat\theta_{t+1}\in K\).

We first show
 $
\widehat\Sigma_{\widehat\theta_t}(\widehat\theta_{t+1})
\xrightarrow{P}\Sigma_{\theta_t}(\theta_{t+1}).
$
On the event
\(\|\widehat\theta_{t+1}-\theta_{t+1}\|_2\leq r\),
\[
\begin{aligned}
&\|\widehat H_{\widehat\theta_t}(\widehat\theta_{t+1})
-H_{\theta_t}(\theta_{t+1})\|_2\\
&\quad\leq
\frac1n\sum_{i=1}^n
\sup_{\|\theta-\theta_{t+1}\|_2\leq r}
\|\nabla^2\ell(Z_{t,i};\theta)
-\nabla^2\ell(Z_{t,i};\theta_{t+1})\|_2\\
&\qquad+
\left\|\frac1n\sum_{i=1}^n\nabla^2\ell(Z_{t,i};\theta_{t+1})
-H_{\widehat\theta_t}(\theta_{t+1})\right\|_2
+\|H_{\widehat\theta_t}(\theta_{t+1})
-H_{\theta_t}(\theta_{t+1})\|_2.
\end{aligned}
\]
By Assumption~\ref{as:pp}(c), the Hessian is continuous in its parameter
argument and has spectral norm at most \(\beta\). Dominated convergence,
together with Lemma~\ref{rm:consistent}, makes the first term arbitrarily
small as \(r\downarrow0\). Conditional on the preceding observations, the
second term is an average of \(n\) i.i.d. bounded matrices with mean
\(H_{\widehat\theta_t}(\theta_{t+1})\), so it is \(O_P(n^{-1/2})\). The
third term tends to zero by Assumption~\ref{as:perturb2}(a). Hence
\[
\widehat H_{\widehat\theta_t}(\widehat\theta_{t+1})
\xrightarrow{P}H_{\theta_t}(\theta_{t+1}).
\]

The same decomposition applies to the middle factor of
\(\widehat\Sigma_{\widehat\theta_t}(\widehat\theta_{t+1})\). On \(K\),
Assumption~\ref{as:pp}(b)--(c) gives
\(\|\nabla\ell(z;\theta)\|_2\leq L_K(z)\) and
\[
\begin{aligned}
&\|\nabla\ell(z;\widehat\theta_{t+1})
\nabla\ell(z;\widehat\theta_{t+1})^\top
-\nabla\ell(z;\theta_{t+1})
\nabla\ell(z;\theta_{t+1})^\top\|_2\leq
2\beta L_K(z)\|\widehat\theta_{t+1}-\theta_{t+1}\|_2.
\end{aligned}
\]
The third-moment bound on \(L_K\) gives the required uniform
integrability. The conditional weak law and Lemma~\ref{rm:consistent}
therefore yield
\[
\begin{aligned}
&\frac1n\sum_{i=1}^n
\nabla\ell(Z_{t,i};\widehat\theta_{t+1})
\nabla\ell(Z_{t,i};\widehat\theta_{t+1})^\top\\
&\qquad\xrightarrow{P}
\mathbb E_{Z\sim\mathcal D(\theta_t)}
[\nabla\ell(Z;\theta_{t+1})\nabla\ell(Z;\theta_{t+1})^\top]\\
&\qquad=
\operatorname{Cov}_{Z\sim\mathcal D(\theta_t)}
(\nabla\ell(Z;\theta_{t+1})),
\end{aligned}
\]
where the last equality uses
\(\mathbb E_{Z\sim\mathcal D(\theta_t)}
[\nabla\ell(Z;\theta_{t+1})]=0\). Finally,
\(\widehat H_{\widehat\theta_t}(\widehat\theta_{t+1})\succeq\gamma I_d\)
by Assumption~\ref{as:pp}(d), so continuity of matrix inversion proves the
claimed convergence of \(\widehat\Sigma\).

Theorem~\ref{thm:discuss2} gives
\(\widehat{\nabla G}_t\xrightarrow{P}\nabla G(\theta_t)\). Since
\(\widehat V_0=V_0=0\), the recursion in \eqref{eq:defempvar}, induction on
\(t\), and the continuous mapping theorem give
\(\widehat V_t\xrightarrow{P}V_t\) for each fixed \(t\geq1\).

If \(V_t\) is nonsingular, then \(\widehat V_t\) is nonsingular with
probability tending to one and
\(\widehat V_t^{-1}\xrightarrow{P}V_t^{-1}\). Theorem~\ref{thm:cltrrm}
and Slutsky's lemma give
\[
n(\widehat\theta_t-\theta_t)^\top\widehat V_t^{-1}
(\widehat\theta_t-\theta_t)\xrightarrow{d}\chi^2_d.
\]
The continuity of the \(\chi^2_d\) distribution function proves
\(\mathbb P(\theta_t\in\widehat{\mathcal R}_t)\to1-\delta\).
\end{proof}

\subsubsection{Discussion on Assumptions for Perturbation Method}
Now we show that the richness assumption and tight optimizer assumption (Assumption~\ref{as:perturb1}(c-d)) hold when we have a correctly-specified generalized affine score model.

\begin{assumption}\label{as:perturblinear}
    (a) We assume that $s(z,\theta;\psi)$ is linear in $\psi$: Specifically, there exists fixed functions $A:\mathcal{Z}\times\Theta\to \mathbb{R}^{d\times p}$ and $B:\mathcal{Z}\times\Theta\to \mathbb{R}^{d}$, such that for each $\theta,\check{\theta}$,  $s(z,\theta;\psi)=A(z,\theta)\psi + B(z,\theta)$, and for any compact set $K$, there exists constant $\gamma_{K}>0$ such that, for each $\theta,\check{\theta}\in K$, 
    \begin{equation*}
    \mathbb{E}_{\check{\theta}}[A(z,\theta)^\top A(z,\theta)]\succeq \gamma_{K}I_{p}.
    \end{equation*}
    Further, we assume that on any compact subset $K\subset\Theta$ and $R>0$, there exists $C_{K,R}:\mathcal{Z}\to\mathbb{R}$ and $L_{K,R}:\mathcal{Z}\to\mathbb{R}$, such that denote 
\begin{equation*}
\begin{aligned}
    &g_{1}^{(i)}(z,\theta;\psi):=\|A(z,\theta)\psi+B(z,\theta)\|^{2}_{2},\ g_{2}^{(i)}(z,\theta;\psi) = A_{i}(z,\theta)\psi + B_{i}(z,\theta),\ \\ &g_{3}^{(i)}(z,\theta;\psi) = \partial_{\theta^{(i)}}A_{i}(z,\theta)\psi + \partial_{\theta^{(i)}}B_{i}(z,\theta),\ i\in [d],
\end{aligned}
\end{equation*} where we denote $A_{i}$ the $i$-th row of matrix-valued function $A$, and $B_{i}$ the $i$-th component of vector valued function $B$.
Then for any function $g(z,\theta;\psi)\in \{g_{j}^{(i)}(z,\theta;\psi):i\in [d],j\in [3]\}$  for any $z\in \mathcal{Z}$, $\theta,\theta^\prime\in K$, $\psi,\psi^\prime\in \Psi_R$,  the following inequality holds:
\begin{equation*}
    |g(z,\theta)|\leq C_{K,R}(z),
\end{equation*}
\begin{equation*}
    |g(z,\theta;\psi)-g(z,\theta^\prime;\psi^\prime)| \leq L_{K,R}(z)\big(\|\theta-\theta^\prime\|_{2}+\|\psi-\psi^\prime\|_{2}\big),
\end{equation*}
with $\sup_{\theta\in K^1}\mathbb{E}_{\theta}[C_{K,R}{(Z)}^{2}+L_{K,R}{(Z)}^{2}]<\infty,$ where $K^1$ is the $1$-enlargement of $K\subset \Theta$.

    (b) We assume that for a compact subset $K\subset \Theta$, there exists $R_{K}>0$ such that for each ${\theta}\in K$, there exists a $\psi^*(\theta)\in \Psi_{R_{K}}$, such that for any $z\in \mathcal{Z}$, the following identity holds,
    \begin{equation*}
        A(z,\theta)\psi^*(\theta) + B(z,\theta)=\nabla_\theta \log p(z;\theta).
    \end{equation*}
\end{assumption}
\begin{remark}
    We remark that richness assumption (Assumption~\ref{as:perturb1}(c)) holds due to Assumption~\ref{as:perturblinear}(b). Further, under Assumption~\ref{as:perturblinear}, the score matching objectives can be rewritten as 
\begin{equation*}
\begin{aligned}
    {J}_{\eta}(\theta;\psi):&= \mathbb{E}_{\theta}[\|A(z,\theta)\psi + B(z,\theta)\|^{2}] + 2\sum_{i=1}^{d}\mathbb{E}_{\theta}\big[\big(\partial_{\theta^{(i)}}A_{i}(z,\theta)\big)\psi + \partial_{\theta^{(i)}}B_{i}(z,\theta)\big] + \mathbb E_\theta[\|\nabla_\theta\log p(z;\theta)\|_2^2]
        \\ &-\frac{2}{\eta}\sum_{i=1}^{d}\bigg({\mathbb{E}}_{\theta+\eta e_i}\big[A_{i}(z,\theta+\eta e_i)\big]-{\mathbb{E}}_{\theta}\big[A_{i}(z,\theta)\big]\bigg)\psi 
        \\&-\frac{2}{\eta}\sum_{i=1}^{d}\bigg({\mathbb{E}}_{\theta+\eta e_i}\big[B_{i}(z,\theta+\eta e_i)\big]-{\mathbb{E}}_{\theta}\big[B_{i}(z,\theta)\big]\bigg),
       \\ \widehat{J}_{n,k}(\theta;\psi):&= \widehat{\mathbb{E}}_{\theta,n}[\|A(z,\theta)\psi + B(z,\theta)\|^{2}] + 2\sum_{i=1}^{d}\widehat{\mathbb{E}}_{\theta,n}\big[\big(\partial_{\theta^{(i)}}A_{i}(z,\theta)\big)\psi + \partial_{\theta^{(i)}}B_{i}(z,\theta)\big] + \mathbb E_\theta[\|\nabla_\theta\log p(z;\theta)\|_2^2]
        \\ &-\frac{2}{\eta}\sum_{i=1}^{d}\bigg(\widehat{\mathbb{E}}_{\theta+\eta e_i,k}\big[A_{i}(z,\theta+\eta e_i)\big]-\widehat{\mathbb{E}}_{\theta,n}\big[A_{i}(z,\theta)\big]\bigg)\psi 
        \\&- \frac{2}{\eta}\sum_{i=1}^{d}\bigg(\widehat{\mathbb{E}}_{\theta+\eta e_i,k}\big[B_{i}(z,\theta+\eta e_i)\big]-\widehat{\mathbb{E}}_{\theta,n}\big[B_{i}(z,\theta)\big]\bigg).
\end{aligned}
\end{equation*}
We note that $\widehat{J}_{n,k}(\theta,\psi)$ is convex in $\psi$, and $J_{\eta}(\theta,\psi),J(\theta,\psi)$ are $\gamma_{K}$-strictly-convex in $\psi$ given compact subset $K$.
\end{remark}

We now prove that under this assumption, the tight optimizer assumption (Assumption~\ref{as:perturb1}(d)) can be achieved.
\begin{lemma}\label{lm:perturblinear}
    Under Assumption~\ref{as:exchange} and \ref{as:perturb1}(a) and Assumption~\ref{as:perturblinear}, we define $\widehat{\psi}_{n,k}(\theta)$ as below: 
    when $\widehat{J}_{n,k}$ is strictly-convex in $\psi$, we take $\widehat{\psi}_{n,k}(\theta)$ to be the empirical minimizer of $\widehat{J}_{n,k}$, otherwise we set it to be $0$. Then for any compact subset $K\subset \Theta$, there exists $R_{K,1}>0$ such that as $\eta\to 0$, $\min(n,k)\to \infty$ and $\eta\sqrt{\min(n,k)}\to \infty$, we have $$\sup_{\theta\in K}P(\|\widehat{\psi}_{n,k}(\theta)\|>R_{K,1})\to 0.$$
\end{lemma}

\begin{proof}[Proof of Lemma~\ref{lm:perturblinear}]
    Fix $\delta>0$. By Assumption~\ref{as:perturblinear}(a), for a fixed compact subset $K\subset\Theta$ and $\theta\in K$, the quadratic term in $\widehat{J}_{n,k}(\theta;\psi)$ is 
    \begin{equation*}
        \psi^{\top}\widehat{\mathbb{E}}_{\theta,n}[A(z,\theta)^\top A(z,\theta)]\psi,
    \end{equation*}
    by uniform law of large numbers, when $\min(n,k)\to \infty$, the matrix $\widehat{\mathbb{E}}_{\theta,n}[A(z,\theta)^\top A(z,\theta)]$ converges in probability uniformly to $\mathbb{E}_{\theta}[A(z,\theta)^\top A(z,\theta)]$, which is positive definite, thus 
    \begin{equation*}
        \inf_{\check{\theta}\in K}P\big(\widehat{J}_{n,k}(\check{\theta},\psi)\text{ is strongly convex in }\psi\big)\to 1,
    \end{equation*}
    as $\min(n,k)\to \infty$. This implies that the $\widehat{\psi}_{n,k}(\check{\theta})$ is exact minimizer with probability converging to $1$. 
    Therefore we shall assume $\widehat{J}_{n,k}$ is strongly convex, and thus $\widehat{\psi}_{n,k}(\theta)$ is uniquely defined.
    By Assumption~\ref{as:perturblinear}(b), for fixed compact set $K$ there exists a constant $R_{K}$, such that $\psi^*(\theta)$ the minimizer of $J(\theta;\psi)$ lies in the ball $\Psi_{R_{K}}$ for any $\theta$.
    By strong convexity, for $\psi\in \Psi$, we have 
    \begin{equation*}
        J(\theta;\psi)-J(\theta;\psi^\ast(\theta))\geq \frac{\gamma_{K}}{2}\|\psi-\psi^*(\theta)\|_{2}^{2},
    \end{equation*} 
    Thus take $R_{K,1}=R_{K}+1$, then for any $\psi$ with $\|\psi\|_{2}=R_{K,1}$,
    \begin{equation*}
        J(\theta;\psi) - J(\theta;\psi^*(\theta))\geq \gamma_{K}/2.
    \end{equation*}
    
    Now for notation simplicity, we write $\widehat{\psi}_{n,k}$ and $\psi^*$ for $\widehat{\psi}_{n,k}(\theta)$ and $\psi^*(\theta)$ respectively.     
    When $\widehat{\psi}_{n,k}$ is outside $\Psi_{R_{K,1}}$, set $\bar{\psi}_{n,k}$ to be the intersection of the sphere $\partial \Psi_{R_{K,1}}$ and the line segment connecting $\widehat{\psi}_{n,k}$ and $\psi^*$.
    Then since $\|\bar{\psi}_{n,k}\|_{2}=R_{K,1}$, we have
    \begin{equation*}
        \begin{aligned}
            P(\|\widehat{\psi}_{n,k}(\theta)\|_{2}> R_{K,1})
            &\leq P\big(J(\theta,\bar{\psi}_{n,k})-J(\theta,\psi^*)\geq \frac{\gamma_{K}}{2}\big)
        \end{aligned}
    \end{equation*}
    Since $\widehat{\psi}_{n,k}$ is a global minimizer of
    $\widehat{J}_{n,k}$, convexity gives
    $\widehat{J}_{n,k}(\theta;\bar{\psi}_{n,k})
    \leq\widehat{J}_{n,k}(\theta;\psi^\ast)$. Therefore,
        \begin{equation*}
        \begin{aligned}
            J(\theta,\bar{\psi}_{n,k}) - J(\theta,{\psi}^*) &= J(\theta,\bar{\psi}_{n,k})-\widehat{J}_{n,k}(\theta,\bar{\psi}_{n,k}) + \big(\widehat{J}_{n,k}(\theta,\bar{\psi}_{n,k})-\widehat{J}_{n,k}(\theta,\psi^*)\big) - \big(J(\theta,\psi^*)-\widehat{J}_{n,k}(\theta,\psi^*) \big)
            \\ &\leq J(\theta,\bar{\psi}_{n,k})-\widehat{J}_{n,k}(\theta,\bar{\psi}_{n,k}) - \big(J(\theta,\psi^*)-\widehat{J}_{n,k}(\theta,\psi^*) \big)
            \\ &\leq 2\sup_{\psi\in \Psi_{R_{K,1}}}\big|J(\theta,\psi)-\widehat{J}_{n,k}(\theta,\psi)\big|
            \\ &\leq 2\sup_{\psi\in \Psi_{R_{K,1}}}\big|J_\eta(\theta,\psi)-\widehat{J}_{n,k}(\theta,\psi)\big| + 2\sup_{\psi\in \Psi_{R_{K,1}}}\big|J_\eta(\theta,\psi)-{J}(\theta,\psi)\big|
            \\ &\leq 2\sup_{\psi\in \Psi_{R_{K,1}}}\big|J_\eta(\theta,\psi)-\widehat{J}_{n,k}(\theta,\psi)\big| + C_{K}d\eta,
        \end{aligned}
    \end{equation*}
where the last inequality follows from Lemma~\ref{lm:perturb1}. Note that under Assumption~\ref{as:perturblinear}, the score function $s(z,\theta;\psi)$ satisfies Assumption~\ref{as:perturb1}(b), thus as in the proof of Theorem~\ref{thm:discuss1}, we have  the following bound holds under \(P_\theta\), uniformly over \(\theta\in K\):
\begin{equation*}
        \sup_{\psi\in \Psi_{R}}\big|\widehat{J}_{n,k}(\theta;\psi)-{J}_{\eta}(\theta;\psi)\big| = O_P(\frac{1}{\sqrt{n}}+\frac{1}{\eta\sqrt{\min(n,k)}}).
    \end{equation*}
Thus when $\min(k,n)\to\infty$, and $\eta\to 0$, we have
    \begin{equation*}
        \begin{aligned}
            \sup_{\theta\in K}P\big(\|\widehat{\psi}_{n,k}\|>R_{K,1}\big)
            &\leq \sup_{\theta\in K}P_{\theta}\big(\sup_{\psi\in \Psi_{R_{K,1}}}\big|J_\eta(\theta,\psi)-\widehat{J}_{n,k}(\theta,\psi)\big|+C_{K}d\eta>\frac{\gamma_{K}}{2}\big)\to 0.
        \end{aligned}
    \end{equation*}

\end{proof}

%% file: Sections/appA_rgd.tex
\subsection{Proofs for Section~\ref{sec:RGD}}\label{subsec:cltrgd}

\subsubsection{Proof of Theorem~\ref{thm:cltrgd}}
\begin{proof}[Proof of Theorem~\ref{thm:cltrgd}]
For \(\theta\in\Theta\) and \(z\in\mathcal Z\), let
\[
f_\theta(z):=\theta-\alpha\nabla\ell(z;\theta).
\]
Then \(\widehat\theta_{t+1}=n^{-1}\sum_{i=1}^n
f_{\widehat\theta_t}(Z_{t,i})\), and the two quantities in
Lemma~\ref{lm:rgd1} are
\[
F(\theta)=\theta-\alpha S(\theta),\qquad
V_f(\theta)=\alpha^2
\operatorname{Cov}_{Z\sim\mathcal D(\theta)}
\bigl(\nabla\ell(Z;\theta)\bigr).
\]
In particular, \(\theta_{t+1}=F(\theta_t)\) and
\(\nabla F(\theta)=I_d-\alpha\nabla S(\theta)\).

We verify the conditions of Lemma~\ref{lm:rgd1}. Fix a compact set
\(K\subset\Theta\). By Assumption~\ref{as:consistent2}(b), \(S\) is
continuously differentiable, and hence \(F\) is continuously differentiable
and bounded on \(K\). To prove that \(V_f\) is continuous, it remains to
consider the second moment of the gradients. For \(\theta,\theta'\in K\),
the difference between this second moment at \(\theta'\) and at \(\theta\)
is
\[
\begin{aligned}
&\mathbb E_{Z\sim\mathcal D(\theta')}
\left[\nabla\ell(Z;\theta')\nabla\ell(Z;\theta')^\top
-\nabla\ell(Z;\theta)\nabla\ell(Z;\theta)^\top\right]\\
&\quad+
\mathbb E_{Z\sim\mathcal D(\theta)}
\left[\nabla\ell(Z;\theta)\nabla\ell(Z;\theta)^\top
\left\{\frac{p(Z;\theta')}{p(Z;\theta)}-1\right\}\right].
\end{aligned}
\]
Assumption~\ref{as:pp}(b)--(c) bounds the spectral norm of the matrix in
the first expectation by
\(2\beta L_K(Z)\|\theta'-\theta\|_2\). Lemma~\ref{rm:consistent} bounds
the factor in braces in the second expectation by a constant multiple of
\(\|\theta'-\theta\|_2\). The moment bound in
Assumption~\ref{as:pp}(b) therefore proves continuity. The same assumption
also gives
\[
\sup_{\theta\in K}
\mathbb E_{Z\sim\mathcal D(\theta)}
\bigl[\|f_\theta(Z)\|_2^3\bigr]<\infty.
\]

We now proceed by induction on \(t\). For \(t=0\),
\(\widehat\theta_0=\theta_0\) and \(V_0=0\). Suppose that
\(\sqrt n(\widehat\theta_t-\theta_t)\xrightarrow{d}\mathcal N(0,V_t)\).
Conditional on all preceding observations and estimators,
\(Z_{t,1},\ldots,Z_{t,n}\) are i.i.d. from
\(\mathcal D(\widehat\theta_t)\). Lemma~\ref{lm:rgd1}, with
\(\check\theta_n=\widehat\theta_t\), \(\check\theta_0=\theta_t\), and
\(A_1=V_t\), gives
\[
\begin{aligned}
\sqrt n(\widehat\theta_{t+1}-\theta_{t+1})
&=\sqrt n\left\{\frac1n\sum_{i=1}^n
f_{\widehat\theta_t}(Z_{t,i})-F(\theta_t)\right\}\\
&\xrightarrow{d}\mathcal N\left(0,
V_f(\theta_t)+\nabla F(\theta_t)V_t\nabla F(\theta_t)^\top\right).
\end{aligned}
\]
The covariance on the right is \(V_{t+1}\) by the recursion in
Theorem~\ref{thm:cltrgd}.
\end{proof}

\subsubsection{Proof of Theorem~\ref{thm:rgdperturb}}
\begin{proof}[Proof of Theorem~\ref{thm:rgdperturb}]
We first derive the formula for \(\nabla S\) in
Section~\ref{subsec:RGDcov}. The map
\[
(\widetilde\theta,\theta)\longmapsto
\mathbb E_{Z\sim\mathcal D(\widetilde\theta)}
[\nabla\ell(Z;\theta)]
\]
is continuously differentiable by Assumption~\ref{as:consistent2}(b), and
\(S(\theta)\) is its value at \(\widetilde\theta=\theta\). Its Jacobian in
the second argument is \(H_\theta(\theta)\). Its Jacobian in the first
argument is
\[
\mathbb E_{Z\sim\mathcal D(\theta)}
[\nabla\ell(Z;\theta)\nabla_\theta\log p(Z;\theta)^\top],
\]
by the likelihood-ratio argument in the proof of
Proposition~\ref{prop:differentiability}. Adding the two Jacobians gives the
formula in Section~\ref{subsec:RGDcov}.

Fix \(t\geq0\). Theorem~\ref{thm:cltrgd} gives
\(\|\widehat\theta_t-\theta_t\|_2=O_P(n^{-1/2})\), with equality to zero
when \(t=0\). The proof of Theorem~\ref{thm:discuss1}, applied to the finite
set \(\{\widehat\theta_j:0\leq j\leq t\}\) and using
Theorem~\ref{thm:cltrgd} in place of Proposition~\ref{prop:consistency},
gives
\[
J\bigl(\widehat\theta_t;
\widehat\psi_{n,k}(\widehat\theta_t)\bigr)
=O_P\left(\frac1{\sqrt n}
+\frac1{\eta\sqrt{\min(n,k)}}+\eta+a_{n,k}\right).
\]

Let \(K\) be a compact set whose interior contains \(\theta_t\). By the
preceding convergence and Assumption~\ref{as:perturb1}(d), we may restrict
attention to an event on which \(\widehat\theta_t\in K\) and
\(\|\widehat\psi_{n,k}(\widehat\theta_t)\|_2\leq R\), for a fixed
\(R<\infty\), with probability arbitrarily close to one. On this event, we
split \(\widehat{\nabla S}_t-\nabla S(\theta_t)\) into four terms.

\emph{Hessian term.}
Conditional on all preceding observations and estimators,
\(\nabla^2\ell(Z_{t,i};\widehat\theta_t)\), \(i\in[n]\), are i.i.d. with
mean \(H_{\widehat\theta_t}(\widehat\theta_t)\). Their spectral norms are
bounded by \(\beta\) under Assumption~\ref{as:pp}(c), so
\[
\left\|\frac1n\sum_{i=1}^n
\nabla^2\ell(Z_{t,i};\widehat\theta_t)
-H_{\widehat\theta_t}(\widehat\theta_t)\right\|_2
=O_P(n^{-1/2}).
\]
Assumption~\ref{as:perturb2}(a) also gives
\[
\|H_{\widehat\theta_t}(\widehat\theta_t)
-H_{\theta_t}(\theta_t)\|_2
\lesssim\|\widehat\theta_t-\theta_t\|_2
=O_P(n^{-1/2}).
\]

\emph{Sampling term.}
For the product of the loss gradient and the fitted score,
Assumption~\ref{as:perturb3} gives a square-integrable envelope and a
square-integrable Lipschitz bound over \(K\times K\times\Psi_R\). The same
uniform empirical-process bound used in the proof of
Theorem~\ref{thm:discuss1} therefore gives
\[
\begin{aligned}
&\left\|\frac1n\sum_{i=1}^n
\nabla\ell(Z_{t,i};\widehat\theta_t)
s\bigl(Z_{t,i},\widehat\theta_t;
\widehat\psi_{n,k}(\widehat\theta_t)\bigr)^\top\right.\\
&\quad\left.-
\mathbb E_{Z\sim\mathcal D(\widehat\theta_t)}
\left[\nabla\ell(Z;\widehat\theta_t)
s\bigl(Z,\widehat\theta_t;
\widehat\psi_{n,k}(\widehat\theta_t)\bigr)^\top\right]
\right\|_2
=O_P(n^{-1/2}).
\end{aligned}
\]

\emph{Score term.}
By the Cauchy--Schwarz inequality, the squared spectral norm of
\[
\mathbb E_{Z\sim\mathcal D(\widehat\theta_t)}
\left[\nabla\ell(Z;\widehat\theta_t)
\left\{s\bigl(Z,\widehat\theta_t;
\widehat\psi_{n,k}(\widehat\theta_t)\bigr)
-\nabla_\theta\log p(Z;\widehat\theta_t)\right\}^\top\right]
\]
is bounded by
\[
\mathbb E_{Z\sim\mathcal D(\widehat\theta_t)}
[\|\nabla\ell(Z;\widehat\theta_t)\|_2^2]
J\bigl(\widehat\theta_t;
\widehat\psi_{n,k}(\widehat\theta_t)\bigr).
\]
The first factor is bounded on \(K\) by Assumption~\ref{as:pp}(b), so this
term contributes
\(O_P(n^{-1/2}+\{\eta\sqrt{\min(n,k)}\}^{-1}+\eta+a_{n,k})\) to the
squared error.

\emph{Distribution term.}
Assumption~\ref{as:perturb2}(b) gives
\[
\begin{aligned}
&\left\|\mathbb E_{Z\sim\mathcal D(\widehat\theta_t)}
[\nabla\ell(Z;\widehat\theta_t)
\nabla_\theta\log p(Z;\widehat\theta_t)^\top]
-\mathbb E_{Z\sim\mathcal D(\theta_t)}
[\nabla\ell(Z;\theta_t)
\nabla_\theta\log p(Z;\theta_t)^\top]\right\|_2\\
&\qquad\lesssim\|\widehat\theta_t-\theta_t\|_2
=O_P(n^{-1/2}).
\end{aligned}
\]
The squared norm of the sum is at most four times the sum of the four
squared norms. Combining the preceding bounds proves the result.
\end{proof}

\subsubsection{Proof of Corollary~\ref{col:rgdfeasible}}
\begin{proof}[Proof of Corollary~\ref{col:rgdfeasible}]
Fix \(t\geq0\). By Theorem~\ref{thm:cltrgd},
\(\widehat\theta_t\xrightarrow{P}\theta_t\) and
\(\widehat\theta_{t+1}\xrightarrow{P}\theta_{t+1}\). The REGD update
therefore gives
\[
\frac{\widehat\theta_t-\widehat\theta_{t+1}}{\alpha}
\xrightarrow{P}
\frac{\theta_t-\theta_{t+1}}{\alpha}=S(\theta_t).
\]
Conditional on all preceding observations and estimators,
\[
\frac1n\sum_{i=1}^n
\nabla\ell(Z_{t,i};\widehat\theta_t)
\nabla\ell(Z_{t,i};\widehat\theta_t)^\top
\]
is an average of i.i.d. matrices. On a compact neighborhood \(K\) of
\(\theta_t\), their spectral norms are bounded by \(L_K(Z)^2\). The
third-moment bound in Assumption~\ref{as:pp}(b) gives uniform
integrability, and the conditional weak law of large numbers gives
\[
\begin{aligned}
&\frac1n\sum_{i=1}^n
\nabla\ell(Z_{t,i};\widehat\theta_t)
\nabla\ell(Z_{t,i};\widehat\theta_t)^\top-
\mathbb E_{Z\sim\mathcal D(\widehat\theta_t)}
[\nabla\ell(Z;\widehat\theta_t)
\nabla\ell(Z;\widehat\theta_t)^\top]
\xrightarrow{P}0.
\end{aligned}
\]
The continuity established in the proof of Theorem~\ref{thm:cltrgd} then
shows that the last two terms in \eqref{eq:rgdempvar} converge in
probability to
\[
\alpha^2\operatorname{Cov}_{Z\sim\mathcal D(\theta_t)}
\bigl(\nabla\ell(Z;\theta_t)\bigr).
\]

Theorem~\ref{thm:rgdperturb} gives
\(\widehat{\nabla S}_t\xrightarrow{P}\nabla S(\theta_t)\). Since
\(\widehat V_0=V_0=0\), induction in the recursion
\eqref{eq:rgdempvar} and the continuous mapping theorem give
\(\widehat V_t\xrightarrow{P}V_t\) for each fixed \(t\geq1\).
If \(V_t\) is nonsingular, then
\(\widehat V_t\) is nonsingular with probability tending to one and
\(\widehat V_t^{-1}\xrightarrow{P}V_t^{-1}\). Theorem~\ref{thm:cltrgd}
and Slutsky's lemma give
\[
n(\widehat\theta_t-\theta_t)^\top\widehat V_t^{-1}
(\widehat\theta_t-\theta_t)\xrightarrow{d}\chi_d^2,
\]
which proves the stated coverage.
\end{proof}

\subsubsection{Proof of Corollary~\ref{col:rgdps}}
\begin{proof}[Proof of Corollary~\ref{col:rgdps}]
We first verify the two claims about \(\rho_{\mathrm{RGD}}\). Since
\(\alpha\leq2/(\beta+\gamma)\) and
\((\beta+\gamma)^2\geq4\beta\gamma\),
\[
\frac{\alpha\beta\gamma}{\beta+\gamma}
\leq\frac{2\beta\gamma}{(\beta+\gamma)^2}
\leq\frac12.
\]
It follows that \(\rho_{\mathrm{RGD}}\geq1/2\). The condition on
\(\varepsilon\) makes the term in braces in the definition of
\(\rho_{\mathrm{RGD}}\) positive, so \(\rho_{\mathrm{RGD}}<1\). It also
implies \(\varepsilon<\gamma/\beta\), and hence
\(\theta_{\mathrm{PS}}\) is well defined. By
\citet[Theorem~3.8]{perdomo2020performative},
\[
\|\theta_t-\theta_{\mathrm{PS}}\|_2
\leq\rho_{\mathrm{RGD}}^t
\|\theta_0-\theta_{\mathrm{PS}}\|_2
\leq B\rho_{\mathrm{RGD}}^t.
\]
Thus \(\theta_t\in\widehat{\mathcal R}_t\) implies
\(\theta_{\mathrm{PS}}\in\widehat{\mathcal R}_t+
\mathcal B(0,B\rho_{\mathrm{RGD}}^t)\). Corollary~\ref{col:rgdfeasible}
therefore gives
\[
\begin{aligned}
&\liminf_{n\to\infty}\mathbb P\left\{\theta_{\mathrm{PS}}\in
\widehat{\mathcal R}_t+\mathcal B(0,B\rho_{\mathrm{RGD}}^t)\right\}\geq
\lim_{n\to\infty}\mathbb P(\theta_t\in\widehat{\mathcal R}_t)
=1-\delta.
\end{aligned}
\]
\end{proof}

%% file: Sections/appA_PPI.tex
\subsection{Details of Appendix~\ref{sec:ppi}: Theory of Prediction-Powered Inference under Performativity}\label{app:resultsforppi}

As in Section~\ref{sec:ppi}, \(\widehat\theta_t\) and
\(\widehat\theta_t(\lambda)\) denote the PPI iterates
\(\widehat\theta_t^{\mathrm{PPI}}\) and
\(\widehat\theta_t^{\mathrm{PPI}}(\lambda)\). For
\(\check\theta,\theta\in\Theta\), let
\begin{align*}
\mathcal L_{\check\theta}(\theta)
&:=\mathbb E_{Z\sim\mathcal D(\check\theta)}[\ell(Z;\theta)],
&
\mathcal L_{\check\theta}^{f}(\theta)
&:=\mathbb E_{Z\sim\mathcal D(\check\theta)}[\ell(X,f(X);\theta)],\\
\mathcal L_{\check\theta,n}(\theta)
&:=\frac1n\sum_{i=1}^n\ell(X_i,Y_i;\theta),
&
\mathcal L_{\check\theta,n}^{f}(\theta)
&:=\frac1n\sum_{i=1}^n\ell(X_i,f(X_i);\theta),\\
\widetilde{\mathcal L}_{\check\theta,N}^{f}(\theta)
&:=\frac1N\sum_{j=1}^N\ell(X_j^u,f(X_j^u);\theta),
&
\widehat{\mathcal L}_{\check\theta,n,N}(\theta;\lambda)
&:=\mathcal L_{\check\theta,n}(\theta)
+\lambda\{\widetilde{\mathcal L}_{\check\theta,N}^{f}(\theta)
-\mathcal L_{\check\theta,n}^{f}(\theta)\}.
\end{align*}
Here \((X_i,Y_i)\) are i.i.d. from \(\mathcal D(\check\theta)\), and
\(X_j^u\) are independently drawn from its \(X\)-marginal.

\begin{assumption}\label{as:ppias2}
The following conditions hold.

(a) For some \(\gamma^f>0\) and every \(x\in\mathcal X\), the map
\(\theta\mapsto\ell(x,f(x);\theta)\) is \(\gamma^f\)-strongly convex.

(b) For every \((x,y)\in\mathcal Z\), the map
\(\theta\mapsto\ell(x,y;\theta)-\ell(x,f(x);\theta)\) is convex.

(c) For every compact \(K\subset\Theta\), there exist measurable
\(A_K^f,L_K^f:\mathcal X\to[0,\infty)\) such that, for all
\(\theta,\theta'\in K\) and \(x\in\mathcal X\),
\[
|\ell(x,f(x);\theta)|\leq A_K^f(x),\qquad
|\ell(x,f(x);\theta)-\ell(x,f(x);\theta')|
\leq L_K^f(x)\|\theta-\theta'\|_2,
\]
and
\[
\sup_{\widetilde\theta\in K}
\mathbb E_{Z\sim\mathcal D(\widetilde\theta)}
[A_K^f(X)^3+L_K^f(X)^3]<\infty.
\]
\end{assumption}
\begin{remark}
Assumption~\ref{as:ppias2}(a)--(b) holds when
\(\ell(x,y;\theta)=\frac12(y-\theta^\top x)^2
+\frac{\gamma^f}{2}\|\theta\|_2^2\).
\end{remark}
For \(\theta\in\Theta\), write \(\ell_\theta(Z):=\ell(Z;\theta)\) and
\(\ell_\theta^f(Z):=\ell(X,f(X);\theta)\). We write
\(\mathbb E_\theta^u\) for expectation under the \(X\)-marginal of
\(\mathcal D(\theta)\). Under Assumption~\ref{as:ppias2},
\(\ell_\theta(z)-\ell_\theta^f(z)\) is convex in \(\theta\) for every
\(z\in\mathcal Z\). For a function \(h:\mathcal Z\to\mathbb R^q\) and
\(\check\theta\in\Theta\), define the labeled and unlabeled empirical
processes by
\begin{equation*}
    \begin{aligned}
        &\mathbb{G}_{n,\check{\theta}}h := \sqrt{n}\big(\frac{1}{n}\sum_{j=1}^{n}h(Z_{j})-\mathbb{E}_{\check{\theta}}h(Z)\big),
        \\& {\mathbb{G}}^{f}_{N,\check{\theta}}h := \sqrt{N}\big(\frac{1}{N}\sum_{j=1}^{N}h(X_{j}^{u},f(X_{j}^{u}))-\mathbb{E}_{\check{\theta}}^{u}h(X,f(X))\big),
    \end{aligned}
    \end{equation*}
    where $\{Z_{j}\}_{j=1}^{n}$ are i.i.d. from $\mathcal{D}(\check{\theta})$ and
     $\{X_{j}^{u}\}_{j=1}^{N}$ are i.i.d. from $\mathcal{D}_{X}(\check{\theta})$. 
\begin{lemma}\label{lm:ppiconsistent1}
    Under Assumption~\ref{as:pp} and \ref{as:ppias2}, for any compact subset $K_{1}, K_{2}\subset \Theta$, constant $\delta>0$, when $n\to \infty$ and $\frac{n}{N}\to r$ for some $r\geq 0$, the following uniform LLN holds:
    \begin{equation*}
        \begin{aligned}
            \sup_{\check{\theta}\in K_{1}}P_{\check{\theta}}\big(\sup_{\lambda\in[0,1]}\sup_{\theta\in K_{2}}\big|\widehat{\mathcal L}_{\check{\theta},n,N}(\theta;\lambda)-\mathcal L_{\check{\theta}}(\theta)\big|>\delta\big) \to 0.
        \end{aligned}
    \end{equation*}
\end{lemma}
\begin{proof}[Proof of Lemma~\ref{lm:ppiconsistent1}]
    Since $\widehat{\mathcal L}_{\check{\theta},n,N}(\theta;\lambda)=\mathcal L_{\check{\theta},n}(\theta)+\lambda\big(\widetilde{\mathcal L}_{\check{\theta},N}^f(\theta)-\mathcal L_{\check{\theta},n}^f(\theta)\big)$, we have
    \begin{equation*}
        \widehat{\mathcal L}_{\check{\theta},n,N}(\theta;\lambda) - \mathcal L_{\check{\theta}}(\theta)=\mathcal L_{\check{\theta},n}(\theta)-\mathcal L_{\check{\theta}}(\theta)+\lambda\big(\widetilde{\mathcal L}_{\check{\theta},N}^f(\theta)-\mathcal L_{\check{\theta}}^{f}(\theta)\big)-\lambda\big(\mathcal L_{\check{\theta},n}^f(\theta)-\mathcal L_{\check{\theta}}^{f}(\theta)\big).
    \end{equation*}
By the proof of Lemma~\ref{lm:consistency1}, the three differences converge uniformly to $0$ over $\lambda\in[0,1]$ as $n,N\to\infty$.
\end{proof}

Write $\widehat{\mathcal L}_{\check{\theta}}(\theta;\lambda)$ for $\widehat{\mathcal L}_{\check{\theta},n,N}(\theta;\lambda)$. The population objective remains $\mathcal L_{\check{\theta}}$, with minimizer $G(\check{\theta})$.

\begin{proposition}[Consistency of the PPI iterates]\label{prop:consistencyppi}
Suppose Assumptions~\ref{as:pp} and~\ref{as:ppias2} hold,
\(n/N\to r\in[0,\infty)\), and \(\widehat\lambda_s\in[0,1]\) for all
\(s\). Then, for each fixed \(t\geq1\),
\[
\sup_{\lambda\in[0,1]}
\|\widehat\theta_t(\lambda)-\theta_t\|_2\xrightarrow{P}0.
\]
In particular, \(\widehat\theta_t\xrightarrow{P}\theta_t\).
\end{proposition}

\begin{proof}[Proof of Proposition~\ref{prop:consistencyppi}]
For \(\check\theta\in\Theta\) and \(\lambda\in[0,1]\), let
\[
\widehat G_\lambda(\check\theta)
:=\argmin_{\theta\in\Theta}
\widehat{\mathcal L}_{\check\theta}(\theta;\lambda).
\]
We prove the result by induction, starting from
\(\widehat\theta_0=\theta_0\).

\emph{Step 1: uniform localization.} Fix a compact set
\(K_1\subset\Theta\). Since
\[
\widehat{\mathcal L}_{\check\theta}(\theta;\lambda)
=(1-\lambda)\mathcal L_{\check\theta,n}(\theta)
+\lambda\widehat{\mathbb E}_{\check\theta,n}
[\ell_\theta-\ell_\theta^f]
+\lambda\widetilde{\mathcal L}_{\check\theta,N}^f(\theta),
\]
Assumptions~\ref{as:pp} and~\ref{as:ppias2} imply that this objective is
\(\min\{\gamma,\gamma^f\}\)-strongly convex. Hence
\[
\min\{\gamma,\gamma^f\}
\|\widehat G_\lambda(\check\theta)-G(\check\theta)\|_2
\leq
\|\nabla_\theta\widehat{\mathcal L}_{\check\theta}
(G(\check\theta);\lambda)\|_2.
\]
The Lipschitz continuity of \(G\) gives a compact set \(K\) containing
\(K_1\) and \(G(K_1)\). Uniformly over \(\lambda\in[0,1]\), the expression
on the right is bounded by the sum of
\[
\|\nabla\mathcal L_{\check\theta,n}(G(\check\theta))\|_2,
\quad
\|\widehat{\mathbb E}_{\check\theta,n}
[\nabla\ell_{G(\check\theta)}-\nabla\ell_{G(\check\theta)}^f]\|_2,
\quad
\|\nabla\widetilde{\mathcal L}_{\check\theta,N}^f
(G(\check\theta))\|_2.
\]
Their second moments are bounded uniformly over \(\check\theta\in K_1\) by
the moment conditions on \(L_K\) and \(L_K^f\). Therefore
\[
\lim_{R\to\infty}\limsup_{n,N\to\infty}
\sup_{\check\theta\in K_1}
P_{\check\theta}\left\{
\sup_{\lambda\in[0,1]}
\|\widehat G_\lambda(\check\theta)-G(\check\theta)\|_2>R
\right\}=0.
\]

\emph{Step 2: uniform consistency.} Suppose
\(\widehat\theta_t\xrightarrow{P}\theta_t\), which holds for \(t=0\).
Choose compact sets \(K_1,K_2\subset\Theta\) so that, with probability
tending to one, \(\widehat\theta_t\in K_1\) and
\(\widehat G_\lambda(\widehat\theta_t)\in K_2\) for every
\(\lambda\in[0,1]\); Step~1 permits this choice. For
\(\check\theta\in K_1\), the minimizing property of
\(\widehat G_\lambda(\check\theta)\) and the \(\gamma\)-strong convexity of
\(\mathcal L_{\check\theta}\) give
\[
\|\widehat G_\lambda(\check\theta)-G(\check\theta)\|_2^2
\leq\frac4\gamma
\sup_{\theta\in K_2}
|\widehat{\mathcal L}_{\check\theta}(\theta;\lambda)
-\mathcal L_{\check\theta}(\theta)|.
\]
Lemma~\ref{lm:ppiconsistent1}, uniformly in \(\lambda\), therefore yields
\[
\sup_{\lambda\in[0,1]}
\|\widehat G_\lambda(\widehat\theta_t)
-G(\widehat\theta_t)\|_2\xrightarrow{P}0.
\]
Since \(G\) is continuous,
\[
\begin{aligned}
\sup_{\lambda\in[0,1]}
\|\widehat\theta_{t+1}(\lambda)-\theta_{t+1}\|_2
&\leq
\sup_{\lambda\in[0,1]}
\|\widehat G_\lambda(\widehat\theta_t)-G(\widehat\theta_t)\|_2+\|G(\widehat\theta_t)-G(\theta_t)\|_2
\xrightarrow{P}0.
\end{aligned}
\]
This completes the induction. Taking
\(\lambda=\widehat\lambda_{t+1}\) gives the final assertion.
\end{proof}

\begin{lemma}\label{lm:localexpansion1ppi}
Under Assumptions~\ref{as:pp} and~\ref{as:ppias2}, let
$\{\check\theta_n\}$ and $\{\theta_n\}$ be random vector sequences
in $\Theta$ such that
\[
\check\theta_n\xrightarrow{P}\check\theta_0,
\qquad
\theta_n\xrightarrow{P}\theta_0.
\]
Conditional on $(\check\theta_n,\theta_n)$, suppose that the labeled
observations $Z_{1,n},\ldots,Z_{n,n}$ are i.i.d.\ from
$\mathcal D(\check\theta_n)$, and the unlabeled covariates
$X^u_{1,n},\ldots,X^u_{N,n}$ are i.i.d.\ from its $X$-marginal
and independent of the labeled observations. If $h_n=O_P^*(1)$, $n,N\to\infty$, and
$n/N\to r\in[0,\infty)$, then
\[
\begin{aligned}
&\mathbb G_{n,\check\theta_n}
\left[
\sqrt n\left(
\ell_{\theta_n+h_n/\sqrt n}-\ell_{\theta_n}
\right)
-h_n^\top\nabla\ell_{\theta_n}
\right]
=o_P^*(1),\\
&\mathbb G_{n,\check\theta_n}
\left[
\sqrt n\left(
\ell^f_{\theta_n+h_n/\sqrt n}-\ell^f_{\theta_n}
\right)
-h_n^\top\nabla\ell^f_{\theta_n}
\right]
=o_P^*(1),\\
&\sqrt{\frac nN}\,
\mathbb G^f_{N,\check\theta_n}
\left[
\sqrt n\left(
\ell^f_{\theta_n+h_n/\sqrt n}-\ell^f_{\theta_n}
\right)
-h_n^\top\nabla\ell^f_{\theta_n}
\right]
=o_P^*(1).
\end{aligned}
\]
Here the empirical processes are centered at the expectations under
$\mathcal D(\check\theta_n)$ and its $X$-marginal, respectively.
\end{lemma}

\begin{proof}[Proof of Lemma~\ref{lm:localexpansion1ppi}]
The first convergence follows from Lemma~\ref{lm:localexpansion1}.
Its proof, applied to $z=(x,f(x))$, gives the second convergence.
We prove the third using the same argument.

For $M>0$, let
\[
B_M=\{h\in\mathbb R^d:\|h\|_2\le M\}.
\]
For fixed $\theta,\check\theta\in\Theta$ and nonrandom $h\in B_M$,
define
\[
r^f_{n,\theta,h}(x)
:=
\sqrt n\left\{
\ell(x,f(x);\theta+h/\sqrt n)
-\ell(x,f(x);\theta)
\right\}
-h^\top\nabla_\theta\ell(x,f(x);\theta).
\]
By Assumption~\ref{as:pp}(c),
\[
|r^f_{n,\theta,h}(x)|
\le \frac{\beta}{\sqrt n}\|h\|_2^2.
\]
The centered empirical process has mean zero, and
\[
\begin{aligned}
\operatorname{Var}_{\check\theta}\left(
\sqrt{\frac nN}\,
\mathbb G^f_{N,\check\theta}r^f_{n,\theta,h}
\right)
&=
\frac nN\,
\operatorname{Var}_{\check\theta}
\left(r^f_{n,\theta,h}(X)\right)\le \frac{\beta^2}{N}\|h\|_2^4.
\end{aligned}
\]
Thus, for each fixed $h$, the process converges to zero in probability
uniformly over $\theta,\check\theta$.

To make this convergence uniform over $B_M$, use the integral
representation
\[
r^f_{n,\theta,h}(x)
=
\int_0^1
h^\top\left\{
\nabla_\theta\ell(x,f(x);\theta+sh/\sqrt n)
-\nabla_\theta\ell(x,f(x);\theta)
\right\}\,ds.
\]
As in the proof of Lemma~\ref{lm:localexpansion1},
Assumption~\ref{as:pp}(c) gives, for $h,h'\in B_M$,
\[
|r^f_{n,\theta,h}(x)-r^f_{n,\theta,h'}(x)|
\le
\frac{2M\beta}{\sqrt n}\|h-h'\|_2.
\]
Consequently,
\[
\left|
\sqrt{\frac nN}\,
\mathbb G^f_{N,\check\theta}
\left(r^f_{n,\theta,h}-r^f_{n,\theta,h'}\right)
\right|
\le 4M\beta\|h-h'\|_2.
\]

Fix $\varepsilon>0$, set $\delta=\varepsilon/(8M\beta)$, and let
$\{h_1,\ldots,h_J\}$ be a finite $\delta$-net of $B_M$.
The net cardinality $J$ is independent of
$n,N,\theta,\check\theta$.
The preceding inequality, a union bound, and the variance bound yield
\[
\begin{aligned}
&\sup_{\theta,\check\theta\in\Theta}
P_{\check\theta}\left(
\sup_{h\in B_M}
\left|
\sqrt{\frac nN}\,
\mathbb G^f_{N,\check\theta}r^f_{n,\theta,h}
\right|>\varepsilon
\right)\le
\sup_{\theta,\check\theta\in\Theta}
P_{\check\theta}\left(
\max_{1\le j\le J}
\left|
\sqrt{\frac nN}\,
\mathbb G^f_{N,\check\theta}r^f_{n,\theta,h_j}
\right|>\frac{\varepsilon}{2}
\right)\le
\frac{4J\beta^2M^4}{N\varepsilon^2}.
\end{aligned}
\]
The supremum over $B_M$ is measurable: by the displayed Lipschitz
bound, it equals the supremum over a countable dense subset.

By the conditional sampling assumption, the same bound applies
almost surely after conditioning on $(\check\theta_n,\theta_n)$:
\[
P\left(
\sup_{h\in B_M}
\left|
\sqrt{\frac nN}\,
\mathbb G^f_{N,\check\theta_n}r^f_{n,\theta_n,h}
\right|>\varepsilon
\,\middle|\,
\check\theta_n,\theta_n
\right)
\le
\frac{4J\beta^2M^4}{N\varepsilon^2}.
\]
Averaging this inequality and localizing $h_n$ gives
\[
\begin{aligned}
&P^*\left(
\left|
\sqrt{\frac nN}\,
\mathbb G^f_{N,\check\theta_n}r^f_{n,\theta_n,h_n}
\right|>\varepsilon
\right)\quad\le
P^*(\|h_n\|_2>M)
+\frac{4J\beta^2M^4}{N\varepsilon^2}.
\end{aligned}
\]
For any $a>0$, tightness of $h_n$ permits choosing $M$ such that
\[
\limsup_n P^*(\|h_n\|_2>M)\le a.
\]
For this fixed $M$, letting $n,N\to\infty$ bounds the limiting upper
probability by $a$. Letting $a\downarrow0$ proves the third convergence.
\end{proof}

\begin{lemma}\label{lm:rrm_clt_ppi}
Suppose
\(\sqrt n(\check\theta_n-\check\theta_0)\xrightarrow{d}
\mathcal N(0,A_1)\) for a positive semidefinite matrix
\(A_1\in\mathbb R^{d\times d}\), and
\(\lambda_n\xrightarrow{P}\lambda_0\). Conditional on
\(\check\theta_n\), the labeled observations are i.i.d. from
\(\mathcal D(\check\theta_n)\), and the unlabeled covariates are i.i.d.
from its \(X\)-marginal and independent of the labeled observations. For
\(h_\theta:\mathcal Z\to\mathbb R^q\), define
\[
\begin{aligned}
F_h(\theta)&:=\mathbb E_\theta[h_\theta(Z)],
&h_\theta^f(Z)&:=h_\theta(X,f(X)),\\
F_h^f(\theta)&:=\mathbb E_\theta[h_\theta^f(Z)],
&V_{\lambda,h}(\theta)&:=\operatorname{Cov}_\theta
\{h_\theta(Z)-\lambda h_\theta^f(Z)\},\\
&&V_{\lambda,h}^f(\theta)&:=\lambda^2
\operatorname{Cov}_\theta\{h_\theta^f(Z)\}.
\end{aligned}
\]
Suppose, for every compact \(K\subset\Theta\), that \(F_h\) is bounded
and continuously differentiable on an open neighborhood of \(K\),
\(V_{\lambda,h}\) and \(V_{\lambda,h}^f\) are continuous on \(K\), and
\[
\sup_{\theta\in K}\mathbb E_\theta
[\|h_\theta(Z)\|_2^3+\|h_\theta^f(Z)\|_2^3]<\infty.
\]
Then, as \(n,N\to\infty\) with \(n/N\to r\geq0\),
\[
\begin{aligned}
&\mathbb G_{n,\check\theta_n}
[h_{\check\theta_n}-\lambda_n h_{\check\theta_n}^f]
+\lambda_n\sqrt{\frac nN}\mathbb G^f_{N,\check\theta_n}
h_{\check\theta_n}
+\sqrt n\{F_h(\check\theta_n)-F_h(\check\theta_0)\}\\
&\quad\xrightarrow{d}\mathcal N\left(0,
V_{\lambda_0,h}(\check\theta_0)
+rV_{\lambda_0,h}^f(\check\theta_0)
+\nabla F_h(\check\theta_0)A_1\nabla F_h(\check\theta_0)^\top
\right).
\end{aligned}
\]
\end{lemma}
\begin{proof}[Proof of Lemma~\ref{lm:rrm_clt_ppi}]
Set \(g_\theta=h_\theta-\lambda_0h_\theta^f\) and
\(F_g(\theta)=\mathbb E_\theta[g_\theta(Z)]\). For fixed
\(s\in\mathbb R^q\), define
\[
\begin{aligned}
\chi(s,\theta)
&:=\exp\{-\tfrac12s^\top V_{\lambda_0,h}(\theta)s\},\\
\chi_n(s,\theta)
&:=\left[\mathbb E_\theta\exp\left\{
\frac{is^\top}{\sqrt n}(g_\theta(Z)-F_g(\theta))\right\}\right]^n,\\
\chi^f(s,\theta)
&:=\exp\{-\tfrac r2s^\top V_{\lambda_0,h}^f(\theta)s\},\\
\chi_{n,N}^f(s,\theta)
&:=\left[\mathbb E_\theta^u\exp\left\{
\frac{i\sqrt n\lambda_0s^\top}{N}
(h_\theta^f(Z)-F_h^f(\theta))\right\}\right]^N.
\end{aligned}
\]
The third-moment bound and the continuity of the two covariances give,
uniformly on compact sets,
\(\chi_n(s,\theta)-\chi(s,\theta)\to0\) and
\(\chi_{n,N}^f(s,\theta)-\chi^f(s,\theta)\to0\). Since
\(\check\theta_n\xrightarrow{P}\check\theta_0\), Lemma~\ref{lm:local}
and the continuity of \(\chi\) and \(\chi^f\) yield
\[
\chi_n(s,\check\theta_n)\xrightarrow{L^1}
\chi(s,\check\theta_0),\qquad
\chi_{n,N}^f(s,\check\theta_n)\xrightarrow{L^1}
\chi^f(s,\check\theta_0).
\]

Let \(\xi_n\) denote the random vector on the left side of the conclusion,
and let \(\xi_n^0\) be the same vector with \(\lambda_n\) replaced by
\(\lambda_0\). Conditional independence of the two samples gives
\[
\begin{aligned}
\mathbb E[e^{is^\top\xi_n^0}]
=\mathbb E\big[&\exp\{i\sqrt n s^\top
(F_h(\check\theta_n)-F_h(\check\theta_0))\}\times\chi_n(s,\check\theta_n)
\chi_{n,N}^f(s,\check\theta_n)\big].
\end{aligned}
\]
The delta method and the preceding \(L^1\) convergences show that this
characteristic function converges to
\[
\exp\left[-\frac12s^\top\left\{
V_{\lambda_0,h}(\check\theta_0)
+rV_{\lambda_0,h}^f(\check\theta_0)
+\nabla F_h(\check\theta_0)A_1\nabla F_h(\check\theta_0)^\top
\right\}s\right].
\]

It remains to replace \(\lambda_0\) by \(\lambda_n\). Their difference is
\((\lambda_0-\lambda_n)\Delta_n\), where
\[
\Delta_n:=\mathbb G_{n,\check\theta_n}h_{\check\theta_n}^f
-\sqrt{\frac nN}\mathbb G^f_{N,\check\theta_n}h_{\check\theta_n}.
\]
For every compact \(K\subset\Theta\), the third-moment condition gives
\[
\sup_{\theta\in K}\mathbb E_\theta
[\|h_\theta^f(Z)\|_2^2]<\infty,
\]
and the same bound holds under the \(X\)-marginal. Since
\(\check\theta_n\) lies in such a compact set with probability tending to
one, the conditional second moments of both empirical processes are
bounded, and hence \(\Delta_n=O_p(1)\). Therefore
\((\lambda_0-\lambda_n)\Delta_n=o_p(1)\). The characteristic functions of
\(\xi_n\) and \(\xi_n^0\) have the same limit, which proves the result.
\end{proof}
\subsubsection{Proof of Theorem~\ref{thm:ppiclt}}

\begin{proof}[Proof of Theorem~\ref{thm:ppiclt}]
We prove the result by induction on \(t\geq0\). For \(t=0\),
\(\widehat\theta_0=\theta_0\) and \(V_0=0\). Suppose the conclusion holds
at round \(t\). Conditional on all preceding observations and estimators,
\(Z_{t,1},\ldots,Z_{t,n}\) are i.i.d. from
\(\mathcal D(\widehat\theta_t)\), and
\(X^u_{t,1},\ldots,X^u_{t,N}\) are independently drawn from its
\(X\)-marginal. Below, \(\lambda=\widehat\lambda_{t+1}\).

\emph{Step 1: localization.} We first show
\[
\sqrt n\{\widehat\theta_{t+1}-G(\widehat\theta_t)\}=O_p(1).
\]
The localization argument in Step~1 of the proof of
Theorem~\ref{thm:cltrrm} applies with
Proposition~\ref{prop:consistencyppi} in place of
Proposition~\ref{prop:consistency}. The centered PPI objective is the sum of
a labeled and an unlabeled empirical process. On
\(\{\theta:\|\theta-G(\widehat\theta_t)\|_2\leq\delta\}\), their increments
have envelopes \(\delta\{L_K(z)+L_K^f(x)\}\) and \(\delta L_K^f(x)\),
respectively. The same bracketing bound therefore applies to both processes,
uniformly over \(\lambda\in[0,1]\); moreover,
\(\sqrt{n/N}\) is bounded.

\emph{Step 2: the leading term.} Let
\[
\begin{aligned}
\widetilde h:={}&-H_{\theta_t}(\theta_{t+1})^{-1}\sqrt n\Bigg[
\frac1n\sum_{i=1}^n\Big\{\nabla\ell(Z_{t,i};\theta_{t+1})
-\lambda\nabla\ell(X_{t,i},f(X_{t,i});\theta_{t+1})\Big\}\\
&\hspace{43mm}+\frac\lambda N\sum_{j=1}^N
\nabla\ell(X^u_{t,j},f(X^u_{t,j});\theta_{t+1})\Bigg].
\end{aligned}
\]
The expression in brackets is the gradient of the PPI objective at
\(\theta_{t+1}\). Since
\(\mathbb E_{\theta_t}[\nabla\ell(Z;\theta_{t+1})]=0\) and the labeled and
unlabeled prediction terms have the same conditional mean, its scaled value
is the random vector in Lemma~\ref{lm:rrm_clt_ppi} with
\[
h_\vartheta(z)=\nabla\ell(z;\theta_{t+1}),\quad
\check\theta_n=\widehat\theta_t,\quad
\check\theta_0=\theta_t,\quad
A_1=V_t,\quad
\lambda_n=\widehat\lambda_{t+1},\quad
\lambda_0=\lambda_{t+1}.
\]
We verify the conditions of the lemma. The map
\(F_h(\vartheta)=\mathbb E_{Z\sim\mathcal D(\vartheta)}
[\nabla\ell(Z;\theta_{t+1})]\) is continuously differentiable by
Assumption~\ref{as:consistent2}(b), and the proof of
Proposition~\ref{prop:differentiability} gives
\[
\nabla F_h(\theta_t)
=-H_{\theta_t}(\theta_{t+1})\nabla G(\theta_t).
\]
Lemma~\ref{rm:consistent} gives continuity of the labeled covariances.
Integrating its density-ratio bound over \(y\) gives the same bound for the
\(X\)-marginal densities and hence continuity of the unlabeled covariance.
The third-moment condition follows from Assumptions~\ref{as:pp}(b)
and~\ref{as:ppias2}(c). Lemma~\ref{lm:rrm_clt_ppi} and the definition of
\(\Sigma_{\theta_t}(\theta_{t+1};\lambda_{t+1})\) now give
\[
\widetilde h\xrightarrow{d}\mathcal N(0,V_{t+1}).
\]

\emph{Step 3: expansion.} Let
\(\widehat h=\sqrt n(\widehat\theta_{t+1}-\theta_{t+1})\). Step~1, the
induction hypothesis, and the delta method applied to \(G\) imply that
\(\widehat h=O_p(1)\). For \(h=O_p(1)\), consider \(n\) times the increase
of the PPI objective from \(\theta_{t+1}\) to
\(\theta_{t+1}+h/\sqrt n\). The two prediction terms have the same
conditional mean, so only the labeled loss contributes to the population
part. The center-independent form of Lemma~\ref{lm:localexpansion1ppi}
controls the three empirical
remainders, and Lemma~\ref{lm:localexpansion2} replaces the current
distribution by \(\mathcal D(\theta_t)\) in the remaining second-order
term. Uniformly for \(h=O_p(1)\), the increase is therefore
\[
\begin{aligned}
&-h^\top H_{\theta_t}(\theta_{t+1})\widetilde h
+n\,\mathbb E_{Z\sim\mathcal D(\theta_t)}
\left[\ell(Z;\theta_{t+1}+h/\sqrt n)-\ell(Z;\theta_{t+1})\right]+o_p(1)\\
&\quad=-h^\top H_{\theta_t}(\theta_{t+1})\widetilde h
+\frac12h^\top H_{\theta_t}(\theta_{t+1})h+o_p(1).
\end{aligned}
\]
The second equality uses
\(\mathbb E_{\theta_t}[\nabla\ell(Z;\theta_{t+1})]=0\) and
Assumption~\ref{as:pp}(c). Evaluating this expansion at \(\widehat h\) and
\(\widetilde h\), the minimizing property of \(\widehat\theta_{t+1}\)
gives
\[
\frac12(\widehat h-\widetilde h)^\top
H_{\theta_t}(\theta_{t+1})(\widehat h-\widetilde h)\leq o_p(1).
\]
Since \(H_{\theta_t}(\theta_{t+1})\succeq\gamma I_d\),
\(\widehat h=\widetilde h+o_p(1)\). Step~2 and Slutsky's lemma prove the
claim at round \(t+1\).
\end{proof}

\subsubsection{Proof of Corollary~\ref{col:ppiclt}}
\begin{proof}[Proof of Corollary~\ref{col:ppiclt}]
Fix \(t\geq0\). Proposition~\ref{prop:consistencyppi} gives
\[
\widehat\theta_t\xrightarrow{P}\theta_t,\qquad
\widehat\theta_{t+1}(\widetilde\lambda_{t+1})\xrightarrow{P}\theta_{t+1},
\qquad
\widehat\theta_{t+1}\xrightarrow{P}\theta_{t+1}.
\]

\emph{Step 1: plug-in covariance.} We first show that, when \(\theta\) is
either \(\widehat\theta_{t+1}(\widetilde\lambda_{t+1})\) or
\(\widehat\theta_{t+1}\),
\[
\sup_{\lambda\in[0,1]}
\left\|
\widehat\Sigma_{\widehat\theta_t}(\theta;\lambda)
-\Sigma_{\theta_t}(\theta_{t+1};\lambda)
\right\|_2\xrightarrow{P}0.
\]
Both matrices are polynomials of degree two in \(\lambda\). Their
coefficients are products of the inverse Hessian with covariances and
cross-covariances of \(\nabla\ell(Z;\theta)\) and
\(\nabla\ell(X,f(X);\theta)\). The proof of
Corollary~\ref{col:rrmfeasible} gives convergence of the Hessian and each
coefficient involving \(\nabla\ell(Z;\theta)\). The same argument applies
to the remaining coefficients with \(L_K(z)\) replaced by
\(L_K(z)+L_K^f(x)\), using Assumption~\ref{as:pp}(c) also at
\(z=(x,f(x))\). Integrating the density-ratio bound in
Lemma~\ref{rm:consistent} over \(y\) gives the corresponding bound for the
\(X\)-marginal, so the argument also applies to the \(n+N\) covariates.
Since \(n/N\to r\) and \(\lambda\in[0,1]\), convergence of the finitely many
coefficients gives the displayed uniform convergence.

\emph{Step 2: the weights.} Since
\(\operatorname{Tr}(A^\top)=\operatorname{Tr}(A)\),
\[
\begin{aligned}
\operatorname{Tr}\Sigma_{\theta_t}(\theta_{t+1};\lambda)
={}&\operatorname{Tr}\Sigma_{\theta_t}(\theta_{t+1})\\
&-2\lambda\operatorname{Tr}\Bigl\{
H_{\theta_t}(\theta_{t+1})^{-1}
\operatorname{Cov}_{Z\sim\mathcal D(\theta_t)}
\bigl(\nabla\ell(Z;\theta_{t+1}),
\nabla\ell(X,f(X);\theta_{t+1})\bigr)
H_{\theta_t}(\theta_{t+1})^{-1}\Bigr\}\\
&+(1+r)\lambda^2\operatorname{Tr}\Bigl\{
H_{\theta_t}(\theta_{t+1})^{-1}
\operatorname{Cov}_{Z\sim\mathcal D(\theta_t)}
\bigl(\nabla\ell(X,f(X);\theta_{t+1})\bigr)
H_{\theta_t}(\theta_{t+1})^{-1}\Bigr\}.
\end{aligned}
\]
The coefficient of \(\lambda^2\) is positive by the condition in the
corollary and the positive definiteness of
\(H_{\theta_t}(\theta_{t+1})\). Thus \(\lambda_{t+1}^*\) is the ratio of the
trace multiplying \(-2\lambda\) to the coefficient of \(\lambda^2\),
truncated to \([0,1]\). The sample criterion has the same quadratic form,
with the covariances specified in Section~\ref{subsec:PPIcov} and \(r\)
replaced by \(n/N\). Step~1, evaluated at
\(\widehat\theta_{t+1}(\widetilde\lambda_{t+1})\), shows that its
coefficients converge to the population coefficients. With probability
tending to one, its quadratic coefficient is positive. Continuity of the
truncated ratio then gives
\[
\widehat\lambda_{t+1}\xrightarrow{P}\lambda_{t+1}^*.
\]
The same argument holds at every fixed round. Therefore
Theorem~\ref{thm:ppiclt} applies with
\(\lambda_s=\lambda_s^*\) at each fixed \(s\).

\emph{Step 3: covariance and coverage.} Theorems~\ref{thm:discuss1}
and~\ref{thm:discuss2} continue to hold along the PPI trajectory: their
proofs use Theorem~\ref{thm:cltrrm} and
Proposition~\ref{prop:consistency} only through the
\(\sqrt n\)-consistency and consistency of the iterates, which
Theorem~\ref{thm:ppiclt} and Proposition~\ref{prop:consistencyppi}
provide. Hence
\(\widehat{\nabla G}_t\xrightarrow{P}\nabla G(\theta_t)\).
Steps~1 and~2 also give
\[
\widehat\Sigma_{\widehat\theta_t}
(\widehat\theta_{t+1};\widehat\lambda_{t+1})
\xrightarrow{P}
\Sigma_{\theta_t}(\theta_{t+1};\lambda_{t+1}^*).
\]
Since \(\widehat V_0=V_0=0\), induction in
\eqref{eq:ppiempvar} and the continuous mapping theorem yield
\(\widehat V_t\xrightarrow{P}V_t\). If \(V_t\) is nonsingular, then
\(\widehat V_t\) is positive definite with probability tending to one and
\(\widehat V_t^{-1}\xrightarrow{P}V_t^{-1}\). Theorem~\ref{thm:ppiclt}
and Slutsky's lemma give
\[
n(\widehat\theta_t-\theta_t)^\top\widehat V_t^{-1}
(\widehat\theta_t-\theta_t)\xrightarrow{d}\chi_d^2,
\]
which proves the stated coverage.
\end{proof}

\subsubsection{Proof of Corollary~\ref{col:ppips}}
\begin{proof}[Proof of Corollary~\ref{col:ppips}]
The population trajectory is the RRM trajectory. Hence the proof of
Corollary~\ref{col:rrmps} applies with
Corollary~\ref{col:rrmfeasible} replaced by
Corollary~\ref{col:ppiclt}.
\end{proof}

%% file: Sections/app_exp_new.tex
\section{Experimental Details}\label{app:expjmlr}
In this section, we provide implementation details for the experiments in Section~\ref{sec:exp}.

\subsection{Pure-Synthetic Experiment}\label{app:pureexp}
\subsubsection{Data Generation}
We use the mixture distribution and conditional outcome model described in Section~\ref{sec:puresyntheticexp}, that is, when deploying \(\theta\):
$$
X\sim(1-a(\theta))\,\mathrm{Unif}(0,1)+a(\theta)\,B(5,2),\quad 
Y\mid X=x\sim\mathrm{Ber} \big(\sigma(-1+2x)\big).
$$
The predictor is fitted using regularized logistic loss,
$$\ell((x,y);\theta)
=\log(1+e^{\theta x})-y\theta x+\frac{\tau}{2}\|\theta\|_2^2,$$
where $\tau=1$ is the regularization parameter. We take $\theta_{0}=1$ and learning rate $\alpha=0.5$.

\subsubsection{Score estimation.}
We implement the score estimator in Section~\ref{sec:RRM}.
In the mild-feedback setting, each deployment uses 20,000
independent auxiliary observations from each of the current
and perturbed distributions, totaling 40,000 observations per
score fit, with perturbation scale $\eta=0.1$. The basis
estimator uses $b(x)=(1,2x-1)^\top$ and the score model
$s(z,\theta;\psi)=\psi^\top b(x)$. We remark that this linear basis does not contain the exact distributional score
of the mixture model. The near-nominal trajectory coverage observed in this setting is consistent with the small contribution of propagated uncertainty, despite the score approximation error.  

In the strong-feedback setting, to better resolve the sharper deployment response, we use a smaller perturbation scale $\eta=0.01$ and increase the auxiliary sample size
to 100,000 observations per distribution to limit estimation
variability. We use a correctly specified basis to isolate the role
of uncertainty propagation in coverage. The effects of basis
specification and auxiliary sample size are examined in the
score-matching ablation later in Appendix~\ref{app:score matching}. 
The basis score model is
\[
s(z,\theta;\psi)=\psi_0+\psi_1 s_\theta(x),
\qquad
s_\theta(x)=\partial_\theta\log p_\theta(x),
\]
where $p_\theta$ is the deployment-dependent covariate density.
This basis contains the exact distributional score, allowing
us to examine covariance estimation under a correctly specified
score model. Its coefficients are nevertheless estimated from
the auxiliary samples.

In both settings, the auxiliary batches are independent of
the labeled observations used for the predictor update, and
the score-model coefficients are refitted at every deployment.

\subsubsection{Contraction and bias bounds.}
We verify the contraction conditions for the synthetic experiment.
Since $\tau=1$,
\[
\nabla_\theta\ell((x,y);\theta)
=x\{\sigma(\theta x)-y\}+\theta,
\]
\[
\nabla_\theta^2\ell((x,y);\theta)
=1+x^2\sigma'(\theta x)\in[1,5/4].
\]
Thus the loss is $1$-strongly convex and $5/4$-smooth in $\theta$. To bound its smoothness in the data, let $c_\sigma:=\sup_{u\in\mathbb R}|u|\sigma'(u).$ For $u\ge0$, the derivative of $u\sigma'(u)$ has the sign of
$1-u\tanh(u/2)$. Consequently, the maximum is attained at the unique positive solution of $u\tanh(u/2)=1,$ giving $u\approx1.543405$ and $c_\sigma\approx0.223872$. We have
\[
\partial_x\nabla_\theta\ell((x,y);\theta)
=\sigma(\theta x)-y+\theta x\sigma'(\theta x).
\]
Therefore, $\left|\partial_x\nabla_\theta\ell((x,y);\theta)\right|
\le1+c_\sigma$, and for $y,y^\prime\in \cY$, we have $$\left|\nabla_\theta\ell((x,y);\theta)-\nabla_\theta\ell((x,y^\prime);\theta)\right|\le |x|\cdot|y-y^\prime|\leq |y-y^\prime|.$$
Combining these bounds, for $z=(x,y)$ and $z'=(x',y')$,
\[
\begin{aligned}
\left|\nabla_\theta\ell(z;\theta)
-\nabla_\theta\ell(z';\theta)\right|
&\le (1+c_\sigma)|x-x'|+|y-y'| \\
&\le \sqrt{(1+c_\sigma)^2+1}\,\|z-z'\|_2,
\end{aligned}
\]
where the last inequality follows from Cauchy--Schwarz.

A valid joint-smoothness constant is consequently
\[
\beta
=\max\left\{\frac54,\sqrt{(1+c_\sigma)^2+1}\right\}
\approx1.580462,
\qquad
\gamma=1.
\]

{
Next, the mixture weight satisfies $|a'(\theta)|
=0.8\,\sigma(\theta)\{1-\sigma(\theta)\}
\le0.2.$
The uniform and beta components share the same conditional
outcome distribution. Their joint total variation distance
therefore equals the total variation distance between their
covariate distributions, which satisfies
\[
\frac12\int_0^1
\left|30x^4(1-x)-1\right|\,dx<0.41.
\]
Couple the mixture indicators at two deployments using a
common uniform random variable. When they agree, use the
same observation from the selected component. When they
differ, maximally couple the two component distributions,
so the observations differ with probability at most $0.41$.
Since the joint support $[0,1]\times\{0,1\}$ has diameter
$\sqrt2$, this gives
\[
W_1\{\mathcal D(\theta),\mathcal D(\theta')\}
\le0.41\sqrt2\,|a(\theta)-a(\theta')|
\le0.082\sqrt2\,|\theta-\theta'|.
\]
Hence we may take
$\varepsilon=0.082\sqrt2\approx0.115966$.
The RRM contraction bound is therefore
\[
\rho_{\mathrm{RRM}}
=\frac{\varepsilon\beta}{\gamma}
\approx0.18327914<1.
\]
For RGD, the step size $\alpha=0.5$ satisfies
\[
\alpha\le\frac{2}{\beta+\gamma}\approx0.775055,
\quad 
\varepsilon
<
\frac{\gamma}{(\beta+\gamma)(1+1.5\alpha\beta)}
\approx0.177330.
\]
The contraction results of Perdomo et al.\
(2020, Theorems~3.5 and~3.8) therefore apply, with
\[
\rho_{\mathrm{RGD}}
=
1-\alpha\beta
\left\{
\frac{\gamma}{\beta+\gamma}
-\varepsilon(1+1.5\alpha\beta)
\right\}
\approx0.89402797<1.
\]
}
For either population trajectory, with its corresponding $\rho$,
\[
|\theta_t-\theta_{\mathrm{PS}}|
\le \rho^t|\theta_0-\theta_{\mathrm{PS}}|.
\]

Following Corollaries~\ref{col:rrmps} and~\ref{col:rgdps}, we bound
the initial distance using only known model parameters.
Since $X\in[0,1]$, $Y\in\{0,1\}$, and $0\le\sigma(u)\le1$,
\[
\left|\nabla_\theta\ell((x,y);\theta_0)\right|
=
\left|x\{\sigma(\theta_0x)-y\}+\tau\theta_0\right|
\le 1+\tau|\theta_0|.
\]
Consequently, uniformly over $\check\theta\in\Theta$,
\[
\left|
\mathbb E_{\check\theta}
[\nabla_\theta\ell(Z;\theta_0)]
\right|
\le 1+\tau|\theta_0|.
\]
Strong convexity of the population loss under
$\mathcal D(\theta_{\mathrm{PS}})$ and its first-order condition at
$\theta_{\mathrm{PS}}$ therefore give
\[
|\theta_0-\theta_{\mathrm{PS}}|
\le
\frac1\gamma
\left|
\mathbb E_{\theta_{\mathrm{PS}}}
[\nabla_\theta\ell(Z;\theta_0)]
\right|
\le
\frac{1+\tau|\theta_0|}{\gamma}.
\]
Define
\[
\widetilde B
:=
|\theta_0|+\frac{1+\tau|\theta_0|}{\gamma}.
\]
Then both $|\theta_0|$ and $|\theta_{\mathrm{PS}}|$ are bounded by
$\widetilde B$. For the present configuration,
$\theta_0=\tau=\gamma=1$, so $\widetilde B=3$.
Combining this bound with contraction yields
\[
|\theta_t-\theta_{\mathrm{PS}}|
\le
\rho^t\bigl(|\theta_0|+|\theta_{\mathrm{PS}}|\bigr)
\le
2\widetilde B\,\rho^t
=
6\rho^t
=:B_t.
\]
This bound requires no evaluation of the stable point or expectations
under its deployment distribution.
Consequently, whenever an interval contains $\theta_t$, enlarging
each endpoint outward by $B_t$ produces an interval containing
$\theta_{\mathrm{PS}}$.

\textit{Model-specific contraction bounds.}
For the $\kappa=80$ controlled experiment, we bound the
population update derivatives directly.
Let $G(q)$ denote the population RRM update and
$S(q)=\mathbb E_{\mathcal D(q)}[\nabla_\theta\ell(Z;q)]$.
Writing $f_q(x)=1+a(q)\{30x^4(1-x)-1\}$, define
\[
H_q(v)=1+\int_0^1x^2\sigma'(vx)f_q(x)\,dx,
\qquad
\Delta(v)=\int_0^1x\{\sigma(vx)-\sigma(2x-1)\}
\{30x^4(1-x)-1\}\,dx.
\]
Since the ridge coefficient is \(\tau=1\) and the logistic loss is convex, \(H_q(v)\geq1\) uniformly in \(q\) and \(v\). Differentiating the first-order condition for $G$ and
the population gradient $S$ yields
\[
G'(q)=-\frac{a'(q)\Delta(G(q))}{H_q(G(q))},
\qquad
S'(q)=H_q(q)+a'(q)\Delta(q).
\]
Using uniform numerical bounds, we have
\[
\sup_{q\in\mathbb R}|G'(q)|\leq0.632751,
\qquad
0.402321\leq S'(q)\leq1.123810.
\]
Therefore, with $\alpha=0.5$,
\[
\rho_{\mathrm{RRM}}\leq0.632751<1,\ 
\rho_{\mathrm{RGD}}
\leq
\max\{|1-0.5(0.402321)|,\,
      |1-0.5(1.123810)|\}
\leq0.798840<1.
\]
The mean value theorem establishes global contraction.
Together with $\widetilde B=3$, these bounds give
$B_t=6\rho^t$, so the criterion $B_t<10^{-3}$
selects rounds 20 and 39 for RRM and RGD, respectively.

\begin{remark}
For Corollary~\ref{col:rrmps} and \ref{col:rgdps}, the same enlargement argument applies whenever a deterministic
bound
\[
\|\theta_t-\theta_{\mathrm{PS}}\|_2\leq B_0\rho^t
\]
is established, with $B_0\geq0$ and $\rho\in(0,1)$. Specifically,
if, for a fixed $t$,
\[
\liminf_{n\to\infty}
\Pr\{\theta_t\in\widehat R_t\}\geq1-\delta,
\]
then
\[
\liminf_{n\to\infty}
\Pr\{\theta_{\mathrm{PS}}\in
\widehat R_t+B(0,B_0\rho^t)\}\geq1-\delta.
\]
Indeed, whenever $\widehat R_t$ contains $\theta_t$, its enlargement
by $B(0,B_0\rho^t)$ contains $\theta_{\mathrm{PS}}$.
Thus, independently established model-specific contraction bounds
can replace the generic contraction factors in the bias-enlargement
arguments of Corollaries~\ref{col:rrmps} and~\ref{col:rgdps},
provided the conditions ensuring trajectory-region coverage hold.
\end{remark}

\subsubsection{Inference and evaluation.}
Population trajectories, covariance references, and performative
stable points are computed separately for the two synthetic settings
using 512-point Gaussian quadrature. For the mild-feedback setting
with $\kappa=1$, this gives $\theta_{\mathrm{PS}}\approx0.053190$.
We construct nominal $95\%$ confidence intervals for the
population iterate $\theta_t$ using the recursive covariance
estimator. For inference on $\theta_{\mathrm{PS}}$, each
interval is expanded by
\[
B_t=2\widetilde B\,\rho^t,
\qquad
\widetilde B=3,
\]
where $\rho$ is the contraction factor of the corresponding
update method. We use the support-based bound
$\widetilde B=3$ and the deterministic reporting rule
\[
t_{\mathrm{report}}
=\min\{t\geq1:B_t<10^{-3}\}.
\]
For the mild-feedback setting ($\kappa=1$), the reporting rule
selects round 6 for RRM and round 78 for RGD, using contraction
bounds $\rho_{\mathrm{RRM}}\approx0.18327914$ and
$\rho_{\mathrm{RGD}}\approx0.89402797$. Both methods are simulated
for 78 rounds. For the strong-feedback setting ($\kappa=80$), the same reporting
rule selects round 20 for RRM and round 39 for RGD, using the
model-specific contraction bounds
$\rho_{\mathrm{RRM}}\approx0.632751$ and
$\rho_{\mathrm{RGD}}\approx0.798840$. Both methods are simulated
for 40 rounds.

Within each setting and method, the reporting round is fixed
across all Monte Carlo repetitions.

\textit{One-step ERM baseline.}
We include a performativity-agnostic baseline that fits the
predictor once using $n$ independent observations
$Z_1,\ldots,Z_n\sim\mathcal D(\theta_0)$ at the fixed initial
deployment:
\[
\widehat\theta_{1}
=
\arg\min_\theta\frac1n\sum_{i=1}^n\ell(Z_i;\theta).
\]
This estimator coincides with the first RRM update and targets
$\theta_1=\arg\min_\theta
\mathbb E_{\mathcal D(\theta_0)}[\ell(Z;\theta)]$.
We construct its confidence region using the ordinary
sandwich covariance estimator. Writing
$g_i=\nabla_\theta\ell(Z_i;\widehat\theta_{1})$
and $\bar g=n^{-1}\sum_{i=1}^n g_i$, define
\[
\widehat H_1
=
\frac1n\sum_{i=1}^n
\nabla^2\ell(Z_i;\widehat\theta_{1}),
\qquad
\widehat\Omega_1
=\hat{\text{Cov}}[\nabla\ell(Z;\hat{\theta_{1}})],\qquad 
\widehat V_1
=
\widehat H_1^{-1}
\widehat\Omega_1
\widehat H_1^{-\top}.
\]
We evaluate this same region
against both $\theta_1$ and $\theta_{\mathrm{PS}}$ without
bias enlargement to account for performativity agnostic in estimation. Its nominal coverage concerns $\theta_1$;
coverage of $\theta_{\mathrm{PS}}$ measures the consequence
of ignoring subsequent performative feedback.

\textit{Per-round M-estimator baseline.}
We also consider a baseline that uses the same RRM point estimates
and update samples but accounts only for current-round sampling
uncertainty. At round $t$, observations $Z_{t-1,1},\ldots,Z_{t-1,n}$
are drawn independently from $\mathcal D(\widehat\theta_{t-1})$
conditional on the preceding deployment. Define
\[
\widehat H_t
=
\frac1n\sum_{i=1}^n
\nabla^2\ell(Z_{t-1,i};\widehat\theta_t),
\qquad
\widehat\Omega_t
=
\widehat{\operatorname{Cov}}
\bigl[\nabla\ell(Z_{t-1,i};\widehat\theta_t)\bigr].
\]
The baseline replaces the recursive covariance estimator with
the current-round sandwich estimator
\[
\widehat V_t^{\mathrm{M}}
=
\widehat H_t^{-1}\widehat\Omega_t\widehat H_t^{-\top},
\]
which estimates $\Sigma_{\theta_{t-1}}(\theta_t)$.
This is equivalent to setting
$\widehat{\nabla G}_{t-1}=0$ in the covariance recursion,
thereby omitting the propagated uncertainty term
\[
\nabla G(\theta_{t-1})V_{t-1}
\nabla G(\theta_{t-1})^\top.
\]
Thus, repeated retraining still accounts for the changing deployment
distribution, but covariance estimation treats the preceding
deployment as fixed. We construct nominal confidence regions using
$\widehat V_t^{\mathrm{M}}/n$ and evaluate their coverage of
$\theta_t$. For inference on the performative stable point,
we retain the same deterministic bias enlargement as our RRM
method. This comparison isolates the effect of accounting for
uncertainty propagated across deployments.

\subsubsection{Prediction-Powered Inference}
\begin{figure}[htbp]
\centering
\includegraphics[width=\linewidth]{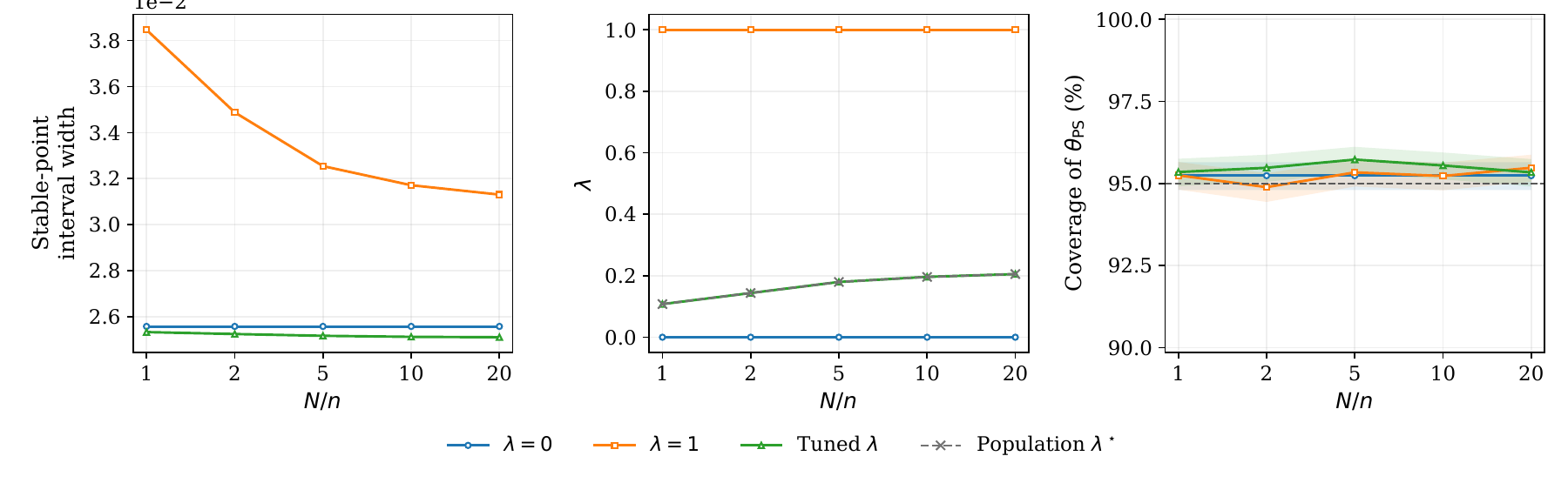}
\caption{Prediction-powered inference in the pure-synthetic
experiment.
From left to right: mean stable-point interval width, mean
selected $\widehat{\lambda}$, and stable-point coverage as
$N/n$ varies. The gray dashed curve shows the population-optimal
$\lambda^\star$. The horizontal dashed line marks nominal
$95\%$ coverage; shaded bands are pointwise exact $95\%$
binomial Monte Carlo confidence intervals.}
\label{fig:pure-appendix-ppi}
\end{figure}

We evaluate the benefits of prediction-powered inference method introduced in Appendix~\ref{sec:ppi} using
the fixed pseudo-labeler $f(x)=\mathbf{1}(x\ge 0.6)$ and varying
the unlabeled-to-labeled sample ratio $N/n$ over
$\{1,2,5,10,20\}$. Table~\ref{tab:pure-ppi-lambda} reports the
mean and RMSE of the selected weight, showing that its mean
closely tracks the population-optimal $\lambda^\star$.
Figure~\ref{fig:pure-appendix-ppi} complements these results
by comparing stable-point interval widths, selected weights,
and coverage across the sample-size ratios.

\begin{table}[htbp]
\centering
\small
\caption{Population-optimal and empirically selected PPI weights
in the pure-synthetic experiment, with $n=2000$ and fixed round
$t={6}$. Means and RMSEs use 10,000 Monte Carlo
repetitions. For each $N/n$, $\lambda^\star$ is the common
reference evaluated at the deterministic population deployment
for round {6}.}
\label{tab:pure-ppi-lambda}
\begin{tabular}{rrrr}
\toprule
$N/n$
& \shortstack{Population $\lambda^\star$}
& \shortstack{Mean $\widehat{\lambda}$}
& \shortstack{RMSE of $\widehat{\lambda}$} \\
\midrule
1  & {0.107863}
   & {0.107762}
   & {0.011998} \\
2  & {0.143818}
   & {0.143778}
   & {0.015899} \\
5  & {0.179772}
   & {0.179700}
   & {0.019544} \\
10 & {0.196115}
   & {0.196447}
   & {0.021505} \\
20 & {0.205454}
   & {0.205390}
   & {0.022777} \\
\bottomrule
\end{tabular}
\end{table}

\subsubsection{Score matching diagnostics}\label{app:score matching}

To examine how score estimation affects inference through the
uncertainty propagation term, we conduct a score-matching ablation
in the strong-feedback synthetic setting of
Section~\ref{subsubsec:expscore}, with activation function
$a(\theta)=0.8\sigma\bigl(80(\theta-0.05)\bigr)$. We vary the
score-model specification and auxiliary sample size to assess
their effects on score-estimation error and confidence-region
coverage. Within each algorithm, all score specifications use
the same update samples and parameter trajectories, isolating
the effect of score estimation on covariance estimation and
the resulting inference.

We test the following methods for score function estimation:
\begin{enumerate}
    \item Basis: an oracle-informed linear family
    $s_{b,c}(x,\theta)=b+c\,s_\theta(x)$, where
    $s_\theta(x)=\partial_\theta\log p_\theta(x)$ is the exact
    score. This family contains the true score and includes
    an intercept. Both coefficients are fitted without
    penalization at each deployment, retaining the
    policy-derivative term in the score-matching objective.

    \item Misspecified basis: a polynomial basis of degree 12 in $x$, refitted at each deployment.

    \item Exact: the analytically evaluated exact score, with no score-estimation error. The covariance estimator
    still uses empirical moments and therefore remains subject to sampling variability.

    \item Zero: the zero-valued constant function as a
    trivial estimator of the score function.
\end{enumerate}

We fix the finite-difference step at $\eta=0.01$ and vary the
auxiliary sample size over $m\in\{10^3,10^4,10^5\}$.
Each fit uses $m$ independent observations from each of the
base and perturbed distributions, totaling $2m$ observations.
All score methods use the same sampled parameter trajectories,
so their comparison isolates the effect of score estimation
on the confidence intervals.
The empirical experiment uses $n=10{,}000$ labeled observations
per deployment and 10,000 Monte Carlo repetitions. Trajectory coverage uses nominal $95\%$ Wald intervals,
whereas performative stable point coverage uses the same
intervals enlarged by $B_t$ at each endpoint. Score-estimation RMSE is evaluated at the deployment parameter used for the reported update. 
Results are evaluated at fixed round $t=20$ for RRM and
$t=39$ for RGD, selected by
\[
t_{\mathrm{report}}
=\min\{t\geq1:B_t<10^{-3}\},
\qquad B_t=6\rho^t.
\]
Here we use the model-specific global contraction bounds
$\rho_{\mathrm{RRM}}\approx0.632751$ and
$\rho_{\mathrm{RGD}}\approx0.798840$ obtained as in previous section. 
The corresponding bias bounds are approximately
$6.35\times10^{-4}$ and $9.42\times10^{-4}$, respectively.

\begin{figure}[htbp]
\centering
\includegraphics[width=\linewidth]{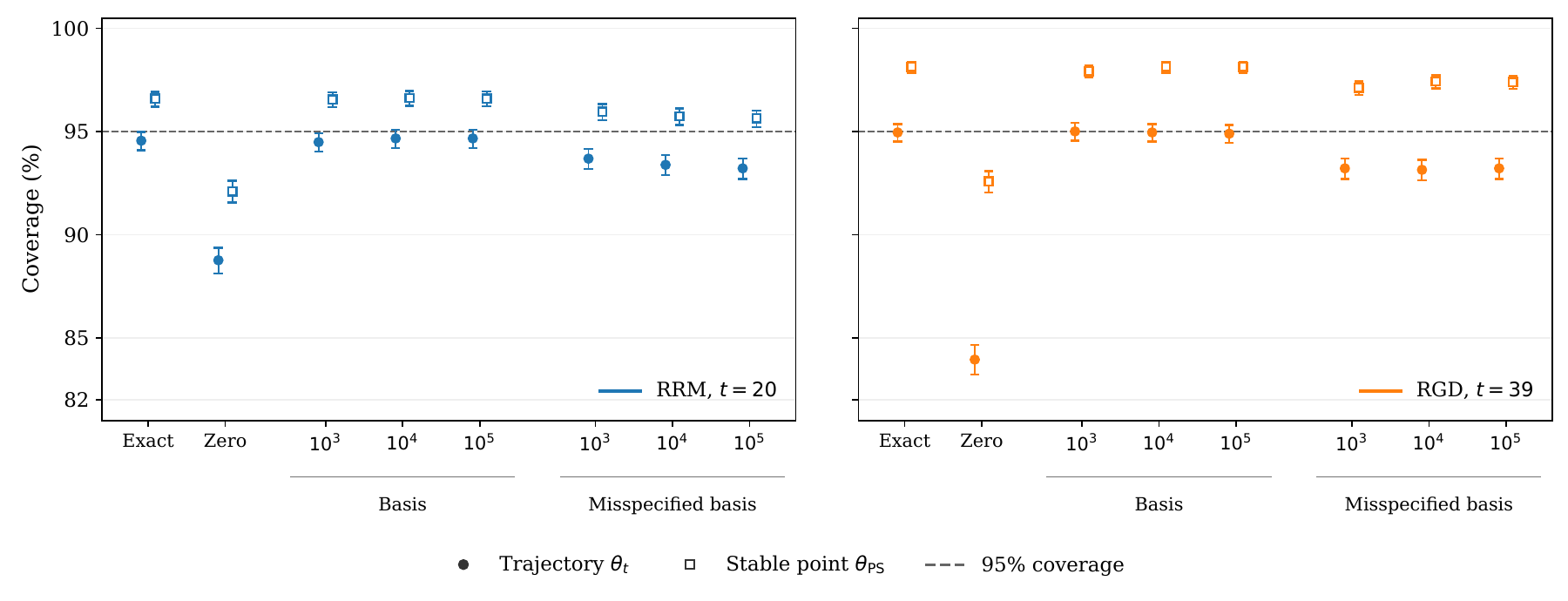}
\caption{Coverage under different score estimators.
RRM and RGD are evaluated at fixed rounds 20 and 39,
respectively. Filled circles show trajectory coverage using
ordinary Wald intervals; open squares show stable-point
coverage after bias enlargement. Error bars are pointwise
$95\%$ Wilson Monte Carlo intervals. The dashed line marks
nominal $95\%$ coverage.}
\label{fig:pure-score-diagnostic}
\end{figure}

Tables~\ref{tab:score-extreme-rrm-rmse}
and~\ref{tab:score-extreme-rgd-rmse} report score-estimation
RMSE and coverage of the population iterate $\theta_t$ and
performative stable point $\theta_{\mathrm{PS}}$ for RRM
and RGD, respectively.
Figure~\ref{fig:pure-score-diagnostic} compares coverage
across the same methods and auxiliary sample sizes.
Increasing $m$ substantially reduces score-estimation RMSE
for both fitted families. The correctly specified, oracle-informed basis yields trajectory coverage close to the exact-score method, whereas the misspecified polynomial basis retains significant trajectory undercoverage.
After bias enlargement, both fitted families attain stable-point coverage above the nominal $95\%$ level, while the zero-score approximation remains below nominal coverage for both algorithms. Both fitted families improve coverage substantially over the zero-score approximation and the score estimation error substantially decreases as the sample number $m$ increases, although coverage need not increase monotonically with $m$. These results demonstrate that perturbation-based score
estimation can substantially improve coverage over the
trivial zero-score approximation.

\begin{table}[htbp]
\centering
\small
\caption{Score estimation and coverage for RRM in the $\kappa=80$ strong feedback experiment,
with $n=10{,}000$, $\eta=0.01$, and 10,000 Monte Carlo repetitions, evaluated at $t=20$.
Each fitted score uses $m$ independent auxiliary observations per distribution, totaling $2m$.
Score RMSE is the square root of the average integrated squared error across repetitions.
Trajectory coverage uses nominal $95\%$ Wald intervals; stable-point intervals
are enlarged at each endpoint by $B_t=6\rho^t$, with the round selected by $B_t<10^{-3}$.}
\label{tab:score-extreme-rrm-rmse}
\begin{tabular}{lrrrr}
\toprule
Score method & $m$ & Score RMSE
& \shortstack{Coverage of\\$\theta_t$ (\%)}
& \shortstack{Coverage of\\$\theta_{\mathrm{PS}}$ (\%)} \\
\midrule
Exact & -- & 0.000 & 94.56 & 96.59 \\
Zero & -- & 14.268 & 88.77 & 92.11 \\
\midrule
Basis & 1,000 & 3.568 & 94.49 & 96.56 \\
Basis & 10,000 & 1.122 & 94.67 & 96.63 \\
Basis & 100,000 & 0.359 & 94.67 & 96.60 \\
\midrule
Misspecified basis & 1,000 & 16.657 & 93.69 & 95.96 \\
Misspecified basis & 10,000 & 5.019 & 93.39 & 95.74 \\
Misspecified basis & 100,000 & 2.442 & 93.22 & 95.64 \\
\bottomrule
\end{tabular}
\end{table}

\begin{table}[htbp]
\centering
\small
\caption{Score estimation and coverage for RGD in the $\kappa=80$ strong feedback experiment,
with $n=10{,}000$, $\eta=0.01$, and 10,000 Monte Carlo repetitions, evaluated at $t=39$.
Each fitted score uses $m$ independent auxiliary observations per distribution, totaling $2m$.
Score RMSE is the square root of the average integrated squared error across repetitions. Trajectory coverage uses nominal $95\%$ Wald intervals; stable-point intervals are enlarged at each endpoint by $B_t=6\rho^t$, with the round selected by $B_t<10^{-3}$.}
\label{tab:score-extreme-rgd-rmse}
\begin{tabular}{lrrrr}
\toprule
Score method & $m$ & Score RMSE
& \shortstack{Coverage of\\$\theta_t$ (\%)}
& \shortstack{Coverage of\\$\theta_{\mathrm{PS}}$ (\%)} \\
\midrule
Exact & -- & 0.000 & 94.96 & 98.13 \\
Zero & -- & 14.342 & 83.96 & 92.59 \\
\midrule
Basis & 1,000 & 3.518 & 95.01 & 97.92 \\
Basis & 10,000 & 1.106 & 94.96 & 98.14 \\
Basis & 100,000 & 0.346 & 94.90 & 98.14 \\
\midrule
Misspecified basis & 1,000 & 16.714 & 93.22 & 97.13 \\
Misspecified basis & 10,000 & 5.042 & 93.15 & 97.43 \\
Misspecified basis & 100,000 & 2.475 & 93.22 & 97.40 \\
\bottomrule
\end{tabular}
\end{table}

\subsection{Semi-Synthetic Experiment}\label{app:semiexp}

In this subsection, we provide implementation details for semi-synthetic experiment (Section~\ref{sec:semiexp}).

\subsubsection{Dataset and deployment population.}
The Learning Mindset dataset \citep{athey2019estimating} contains
10,391 observations recording student characteristics, intervention
assignments, and academic achievement. We allocate 2,000 observations
to simulator training and 500 to validation, with splits stratified
by school and intervention assignment. The binary achievement
threshold is the median achievement score in the simulator-training
sample. The remaining 7,891 student profiles form a construction set
for generating deployment contexts.

We generate 100,000 synthetic contexts by independently sampling
construction profiles with replacement and perturbing their five
continuous school-level covariates. Specifically, the perturbation
is $0.05LU$, where $LL^\top$ is the empirical covariance matrix of
these covariates in the construction set and $U$ is uniformly
distributed on the five-dimensional ball of radius $\sqrt{7}$,
so that $\operatorname{Cov}(U)=I_5$. Perturbed values are reflected
into their respective construction-set ranges, while categorical
and ordinal characteristics retain their sampled values.
Perturbations are applied before the fixed encoding and PCA
transformations. These 100,000 generated contexts form a single fixed deployment pool. Empirical RRM/RGD updates sample uniformly with replacement from this pool, while population reference trajectories and
stable points are computed using the same pool.

\paragraph{Simulator.}
Student characteristics are encoded into 26 covariates, comprising
six standardized numerical variables and 20 categorical indicator
variables. Encoding parameters are fitted using the
simulator-training sample. The outcome simulator maps these
covariates through a shared tanh layer with 24 hidden units and
two treatment-specific logistic output heads. Training minimizes
binary log loss for the observed intervention assignments using
Adam with learning rate $0.006$, batch size $256$, and an
$L_2$ weight-penalty coefficient of $2\times10^{-4}$.

Training runs for at most 1,000 epochs, with early stopping after
80 consecutive epochs without a validation log-loss improvement
exceeding $10^{-5}$. The retained checkpoint is selected using
validation loss and remains fixed throughout the deployment
experiment. In the fitted simulator used here, training terminates
after 120 epochs and the retained checkpoint comes from epoch 40.

\paragraph{Feature representation.}
For the deployed predictor, we retain three principal components
fitted to the encoded training covariates and whiten them using
their training standard deviations. Let
$\phi_0(x)\in\mathbb R^3$ denote these whitened coordinates.
We define
\[
\phi(x) = \frac{s_\phi}{c} \big(1,\ \phi_0(x)\big)^{\top},
\qquad
c
=
\max_{i\in\mathcal I_{\mathrm{train}}}
\left\|
\big(
1, \ 
\phi_0(x_i)\big)^{\top}
\right\|_2,
\]
where $\mathcal I_{\mathrm{train}}$ indexes the simulator-training data,
$c\approx4.67$, and the feature multiplier is
$s_\phi\approx3.54$. The resulting predictor has four parameters,
including the intercept. The retained components explain
approximately $61.688\%$ of the variance in the encoded training
covariates. The encoding, PCA transformation, whitening scales,
and feature normalization remain fixed across deployment rounds
and repetitions.

\subsubsection{Deployment and experimental protocol.}
We initialize $\theta_0=\mathbf 0$, use regularization $\tau=1$,
and set the RGD step size to $\alpha=0.8$. Intervention assignment uses $\delta=0.1$ and $\kappa=6$. At each deployment, student profiles are sampled independently and uniformly with replacement from the fixed deployment pool of 100,000 augmented contexts. Intervention assignments and binary outcomes are then generated
according to the deployment law described in the main text,
using the frozen outcome simulator.

We conduct 10,000 independent Monte Carlo repetitions for each
method and labeled sample size $n=2000$. Fresh observations
are generated at every deployment. The simulator and feature transformation remain fixed across repetitions.

\paragraph{Score estimation.}
In this setup, we can directly evaluate the analytic score at deployment parameter
$q$:
\[
s_q(x,w)
=
\frac{w-e_q(x)}{e_q(x)(1-e_q(x))}
\nabla_q e_q(x).
\]
This expression depends on the intervention policy and does not
require outcome labels.

\subsubsection{Inference and evaluation.}
Population trajectories, covariance references, and the performative stable point are computed by averaging uniformly
over the fixed pool of 100,000 generated contexts and summing over the possible intervention--outcome states under the frozen simulator. Confidence regions are constructed from observations sampled from this same pool, and coverage is evaluated against its population iterates and performative stable point.

We construct nominal $95\%$ Wald confidence ellipsoids for the
four-dimensional population iterate using the feasible covariance
estimators in Corollaries~\ref{col:rrmfeasible}
and~\ref{col:rgdfeasible}, with a $\chi^2_4$ cutoff.
For stable-point inference, we enlarge each ellipsoid by a
Euclidean ball, following Corollaries~\ref{col:rrmps}
and~\ref{col:rgdps}.

\paragraph{Contraction bounds.}
We remark here that as in Appendix~\ref{app:pureexp}, explicitly differentiating the conditional mean
and the population updates gives sharper contraction bounds for the
semi-synthetic model. For the fixed covariate distribution, write
\[
Q=\bE[\phi(X)\phi(X)^\top],
\qquad
L_\phi=\lambda_{\max}(Q),
\qquad
c=\frac{(1-2\delta)\kappa}{4}.
\]
Let $p(x,w)=\Pr(Y=1\mid X=x,W=w)$ be the conditional outcome
probability produced by the fixed simulator. Under deployment $u$,
the intervention probability is
\[
e_u(x)
=\delta+(1-2\delta)\sigma(-\kappa u^\top\phi(x)).
\]
Define the conditional outcome mean under this deployment by
\[
m_u(x)
:=
\bE_{\cD(u)}[Y\mid X=x]
=
p(x,0)+e_u(x)\{p(x,1)-p(x,0)\}.
\]
Differentiating then  gives
\[
\nabla_u m_u(x)
=
-(1-2\delta)\kappa\{p(x,1)-p(x,0)\}
\sigma'(-\kappa u^\top\phi(x))\phi(x).
\] 
Differentiating the first-order condition defining $G$ and the
population gradient $S$ gives
\[
\begin{aligned}
\nabla G(u)
&=
H_u(G(u))^{-1}
\bE[\phi(X)\nabla_u m_u(X)^\top],\\
\nabla S(u)
&=
H_u(u)-\bE[\phi(X)\nabla_u m_u(X)^\top].
\end{aligned}
\]
Because $|p(x,1)-p(x,0)|\le1$ and $0\le\sigma'\le1/4$,
the matrix $\bE[\phi(X)\nabla_u m_u(X)^\top]$ is symmetric
with spectral norm at most $cL_\phi$, while
$\tau I_d\preceq H_u(v)\preceq(\tau+L_\phi/4)I_d$.
Consequently,
\[
\begin{aligned}
\|\nabla G(u)\|_2
&\le \frac{cL_\phi}{\tau}, \|I_d-\alpha\nabla S(u)\|_2
\le
\max\left\{
|1-\alpha(\tau-cL_\phi)|,\,
\left|1-\alpha\{\tau+(1/4+c)L_\phi\}\right|
\right\}.
\end{aligned}
\]
These bounds hold uniformly over $u\in\mathbb R^d$ and give
global contraction factors for RRM and RGD, respectively, whenever
\[
\tau>cL_\phi,
\qquad
0<\alpha<\frac{2}{\tau+(1/4+c)L_\phi}.
\]
For the fixed deployment pool,
$L_\phi\approx0.63$,
$\rho_{\mathrm{RRM}}\approx0.76$, and
$\rho_{\mathrm{RGD}}\approx0.81$.
With $\theta_0=\mathbf 0$, the bias radius used for each method is $B_t=\frac{\sqrt{L_\phi}}{\tau}\rho^t,$
where $\rho$ is the corresponding contraction bound. We use the deterministic reporting rule
\[
t_{\mathrm{report}}
=\min\{t\geq1:B_t<10^{-3}\}.
\]
This selects fixed round 24 for RRM and fixed round 31 for RGD. The same reporting round is used for every repetition within each method. Each simulation saves 100 rounds; the reported results use these fixed rounds rather than empirical stopping rounds. The trajectory-region diameter is
\[
2\sqrt{
\frac{\chi^2_{4,0.95}\lambda_{\max}(\widehat V_t)}{n}
},
\]
where $\hat{V}_{t}$ is the empirical variance estimated using Theorems~\ref{thm:cltrrm} and \ref{thm:cltrgd}, and the stable-point-region diameter adds $2B_t$.

\subsection{LLM Routing}\label{app:llmexp}
This section provides implementation details about
the LLM routing experiment in 
Section~\ref{sec:llmexp}.

\subsubsection{Dataset and response-bank construction.}
We assemble 8,792 questions from GSM8K and 10,962 questions
from the training and validation splits of CommonsenseQA.
Of these, 500 questions are reserved for calibration of
the feature transformation. The remaining 19,254 questions,
comprising 8,572 GSM8K questions and 10,682 CommonsenseQA
questions, form the fixed deployment pool.
Empirical updates and population reference calculations
use this same deployment pool.

{
For each question, the response bank contains ten entries
under each of two modes of the frozen Qwen2.5-1.5B model.
The fast mode uses direct answer generation on both datasets.
For GSM8K, the slow mode uses reasoning followed by a final
answer, retaining all ten draws per mode from the existing
bank without filtering by correctness.
Inspired by self-consistency decoding
\citep{wang2022self}, we construct the
CommonsenseQA slow mode by plurality voting over five
independently sampled direct answers. 
Ties in this vote are broken uniformly at random using a
fixed seed, without consulting reference answers.
Thus, ten slow entries require fifty component generations
per CommonsenseQA question.
The token cost of each voted entry is the sum of the token
counts of its five component generations.
\par}

Generation uses temperature $0.6$ and top-$p=0.95$, with
maximum lengths of 8,192 reasoning tokens and 64
final-answer tokens.
Final answers are constrained to the appropriate format:
numerical answers for GSM8K and option letters for
CommonsenseQA.
Reference answers and worked solutions are excluded from
the prompts.
Incorrect and duplicate responses are retained with their
sampling multiplicities.

The bank contains {395,080 mode-level entries:
10,000 for calibration from 500 questions and 385,080 for empirical deployment. A voted slow answer counts as one bank entry.}
The model and response bank remain fixed throughout the
routing experiment.

\paragraph{Question representation.}
We represent each question by the model's 1,536-dimensional
final-layer hidden state at the last token of a question-only
prompt, including the answer choices for CommonsenseQA.
{Before PCA, we concatenate two unit-valued one-hot
dataset indicators with this embedding, giving a
1,538-dimensional input.}
We fit PCA using only the 500 calibration questions and retain
four components, which explain {$74.852\%$} of the
calibration variance.
Each component is standardized using its calibration mean
and standard deviation and clipped to $[-3,3]$.
{We additionally radially clip the resulting
four-dimensional vector to have Euclidean norm at most
$\sqrt{19}$.}
Writing $\phi_0(x)\in\mathbb R^4$ for these clipped coordinates,
we define
\[
\phi(x)
=
c_\phi(1,\phi_0(x)^\top)^\top,
\qquad
{
c_\phi=\sqrt{\frac{45.125}{20}}\approx1.50.
}
\]
Thus, the router parameter belongs to $\Theta=\mathbb R^5$.
The representation satisfies
{$\|\phi(x)\|_2^2\leq45.125$}.
{Dataset identity enters the router through the
joint PCA representation and is also retained for stratified
reporting.}
The fitted transformation remains fixed across deployment
rounds, sample sizes, and repetitions.

\subsubsection{Deployment and experimental protocol.}
We initialize $\theta_0=\mathbf0$, use regularization $\tau=1$,
and set the RGD step size to $\alpha=0.8$.
The logistic predictor and routing probabilities are
\[
h_\theta(x)=\sigma(\theta^\top\phi(x)),
\qquad
\pi_\theta(1\mid x)
=
(1-\xi)h_\theta(x)+\frac{\xi}{2},
\qquad
\xi=0.2,
\]
where mode $1$ denotes
{the dataset-specific slow mode: reasoning followed
by a final answer for GSM8K, and five-answer voting for
CommonsenseQA}.
The exploration mixture governs candidate generation,
while the regularized logistic loss uses $h_\theta$.

At each deployment, we sample $n$ questions independently
and uniformly with replacement.
For each question, we independently sample five modes from
$\pi_\theta(\cdot\mid x)$ and then sample one response
uniformly from the ten stored entries for each
selected mode.
Response sampling is also with replacement. For the reported experiment, utility is correctness
minus $0.1$ times generated tokens divided by $1000$:
\[
u(x,r)
=
\mathbf 1\{r\text{ is correct}\}
-
0.1\,\frac{\operatorname{tokens}(r)}{1000}.
\] 
The candidate with the highest utility wins, with ties resolved
in favor of the first candidate.
Its mode supplies the binary supervision label.

Let $J^*$ denote the winning candidate, and write $U_j=u(X,R_j)$. Define the highest observed utility within each represented mode: $M_a=\max_{j:A_j=a}U_j,\ a\in\{0,1\}.$
We weight the logistic loss by the observed utility gap:
\[
w(Z)=
\begin{cases}
|M_1-M_0|/C,&\text{if both modes appear},\\
0,&\text{otherwise}.
\end{cases}
\qquad
C=1+\frac{0.1\times8256}{1000}=1.8256.
\]
Here, $8256$ is the maximum total token count for a GSM8K slow-mode response. It also bounds the token count of every other response-bank entry, including CommonsenseQA voting entries, which use at most $5\times 64 = 320$ tokens.
Consequently, utilities lie in $[-0.8256,1]$, so
$|M_1-M_0|\leq1.8256=C$ and $0\leq w(Z)\leq1$. 
The observation-level loss is
\[
\ell(Z;\theta)
=
w(Z)\left[
\log\left(1+e^{\theta^\top\phi(X)}\right)
-
A_{J^*}\theta^\top\phi(X)
\right]
+
\frac{\tau}{2}\|\theta\|_2^2.
\]
Candidate sets containing only one mode, and mixed-mode
candidate sets whose best fast and slow utilities are
equal, have zero weight. 
All $n$ observations remain in the empirical-average
denominator; we do not divide by the sum of the weights.

Fresh question--candidate observations are sampled at every
deployment.
We report \(n=10000\) labeled question-level observations per deployment and 10,000 independent Monte Carlo repetitions. Each observation contains five candidates. Simulations retain 198 rounds for both RRM and RGD.

\paragraph{Distribution score and covariance estimation.}
For the joint observation
$$Z=(X,A_1,R_1,\ldots,A_N, R_N),$$ question sampling and the
mode-specific response distributions do not depend on
$\theta$.
Consequently, the distribution score is available
analytically:
$$ s_q(Z)=
\frac{
\left\{\sum_{j=1}^{N}A_j-N\pi_q(1\mid X)\right\}
(1-\xi)h_q(X)\{1-h_q(X)\}
}{
\pi_q(1\mid X)\{1-\pi_q(1\mid X)\}
}\phi(X).$$
{The results reported here use this analytic score,
so no score fitting or auxiliary perturbation batches
are required.}
Scores are evaluated at the deployment parameter that generated the batch, while loss gradients and Hessians are evaluated at the updated parameter for RRM and at the same deployment parameter for RGD.

\paragraph{Population references.}
All population calculations are conditional on the fixed
merged pool of 19,254 questions, response bank, and feature
transformation. We consider best-of-$N$ selection, with
$N=5$ in the reported experiment. For a deployed parameter $q$, define
\[
h_q(x)=\sigma(q^\top\phi(x)),
\qquad
b_q(x)=\pi_q(1\mid x)
=\frac{\xi}{2}+(1-\xi)h_q(x).
\]
The probability of drawing exactly $j$ slow-mode candidates is
\[
v_j(h_q(x))
=
\binom Nj b_q(x)^j\{1-b_q(x)\}^{N-j},
\qquad j=0,\ldots,N.
\]
Write $\mathbb E_j[\cdot\mid X]$ for expectation over candidate
randomness conditional on exactly $j$ slow candidates and
$N-j$ fast candidates. Responses are sampled independently
and uniformly with replacement from the corresponding
frozen banks. This conditional law does not depend on $q$. Since single-mode draws have zero weight, the population
loss gradient at a candidate parameter $\theta$ is
\[
\mathcal S(q,\theta)
=
\tau\theta+
\mathbb E_X\!\left[
\sum_{j=1}^{N-1}v_j(h_q(X))
\mathbb E_j\!\left[
w(h_\theta(X)-A_{J^\ast})\mid X
\right]\phi(X)
\right].
\]
Here $A_{J^\ast}$ is the winning mode, and $w$ is the
normalized absolute utility gap between the best slow and fast candidates. The conditional expectations are
computed exactly from the frozen response banks using
the distributions of the within-mode maximum utilities.
Thus, the population RRM map $G(q)$ solves
$\mathcal S(q,G(q))=0$, while the population RGD map is
$q-\alpha\mathcal S(q,q)$.

We compute population trajectories, covariance references, and the  performative stable point by averaging uniformly over the merged question pool and integrating over candidate randomness. These reference calculations are performed after empirical sampling.

\subsubsection{Inference and evaluation.}
We construct nominal $95\%$ joint confidence ellipsoids
for the {five-dimensional} population iterate using
the covariance recursions in Theorems~\ref{thm:cltrrm}
and~\ref{thm:cltrgd}.
For inference on the performative stable point, each
ellipsoid is enlarged by a Euclidean ball of radius $B_t$.
Coverage is evaluated using membership in this
ellipsoid-plus-ball region.
Its Euclidean diameter is twice the sum of the largest
ellipsoid semi-axis and $B_t$.

\textbf{Contraction bounds.}
As in the semi-synthetic experiment, we obtain contraction bounds
specific to the routing model by differentiating the population
RRM and RGD updates under the exact probabilistic model.
Conditional on $X$, the number of slow candidates follows a binomial
distribution with parameters $N$ and
$\xi/2+(1-\xi)\sigma(u^\top\phi(X))$.
Conditional on this count, the response-bank distribution is
independent of the deployed parameter.
Using these count probabilities, $0\le w\le1$, and the convention
$w=0$ for single-mode draws, we bound the update Jacobians uniformly
over $u\in\bR^d$. The resulting scalar derivative bounds are verified
using Bernstein polynomial bounds on $[0,1]$.

For $N=5$, $\xi=0.2$, $\tau=1$, $\alpha=0.8$, and
$L_\phi=\lambda_{\max}(\bE[\phi(X)\phi(X)^\top])\le2.643$,
the resulting contraction factors are
\[
\rho_{\mathrm{RRM}}\approx0.851901,
\qquad
\rho_{\mathrm{RGD}}\approx0.956687.
\]
These constants depend only on the candidate count, exploration,
weight bound, feature second-moment bound, and algorithm parameters;
they require neither the stable point nor the response banks'
correctness rates or utility-gap moments.

{

For \(n=10000\), the deterministic reporting criterion \(B_t<10^{-3}\) selects round 55 for RRM and round 198 for RGD, where we have \[
B_t=6.297670\,\rho^t.
\] These rounds are shared by all Monte Carlo repetitions. Coverage is pointwise in the deployment round and conditional on the frozen question pools, response bank, and feature transformation.

\subsection{Jacobian estimation diagnostics.}\label{app:jacobian}
For each setting, we evaluate the RRM Jacobian estimator at its
fixed reporting round $T$. In repetition $r$, the estimator
$\widehat{\nabla G}_{T-1}^{(r)}$ is compared with the population
derivative
$\nabla G(\widehat\theta_{T-1}^{(r)})$ at the same realized
deployment. Population derivatives are computed using numerical
quadrature in the synthetic settings and by averaging over the
fixed deployment pools and integrating over the remaining
randomness in the semi-synthetic and LLM settings.
These calculations are used only for evaluation.

With $R=10{,}000$ repetitions, we report mean population Jacobian norm $\frac1R\sum_{r=1}^{R}
\left\|\nabla G(\widehat\theta_{T-1}^{(r)})\right\|_2,$
and Jacobian RMSE $$\left[
\frac1R\sum_{r=1}^{R}
\left\|
\widehat{\nabla G}_{T-1}^{(r)}
-
\nabla G(\widehat\theta_{T-1}^{(r)})
\right\|_2^2
\right]^{1/2},$$
where $\|\cdot\|_2$ denotes the spectral norm. Thus, the RMSE
measures matrix estimation error rather than the difference
between the estimated and population norms. For the zero-Jacobian baseline, we replace $\widehat{\nabla G}_{T-1}^{(r)}$ by zero in the same calculation. Its RMSEs are approximately
$0.0079$, $0.5737$, $0.1312$, and $0.0661$ for the mild synthetic,
strong synthetic, semi-synthetic, and LLM settings, respectively.
The coverage column in Table~\ref{tab:feedback-jacobian} reports
the per-round M-estimator's coverage of the performative stable
point, retaining the same deterministic bias enlargement as RRM.